\pdfoutput=1
\documentclass[10pt,journal,final]{IEEEtran}

\ifCLASSINFOpdf

\usepackage[pdftex]{graphicx}
\usepackage{color}
\usepackage{multirow}
\usepackage{booktabs}
\else
\usepackage[utf8]{inputenc}

\fi
\usepackage{amsfonts,amssymb}
\usepackage{amsmath}
\usepackage{algorithm}
\usepackage{algorithmicx}

\ifCLASSOPTIONcompsoc
\usepackage[caption=false,font=normalsize,labelfont=sf,textfont=sf]{subfig}
\else
\usepackage[caption=false,font=footnotesize]{subfig}
\fi
\usepackage{cite}
\usepackage{dblfloatfix}
\begin{document}
	%
	
	\title{Semantic Attention and Scale Complementary Network for Instance Segmentation in Remote Sensing Images}
	
	%
	%
	%
	
	\author{Tianyang~Zhang,
		Xiangrong~Zhang,
		Peng~Zhu,
		Xu~Tang,
		Chen~Li,
		Licheng~Jiao,
		and~Huiyu~Zhou}

	\maketitle
	
	\begin{abstract}
		In this paper, we focus on the  challenging multi-category instance segmentation problem in remote sensing images (RSIs), which  aims  at  predicting  the  categories of all instances and localizing them with pixel-level masks. Although many landmark frameworks have demonstrated promising performance in instance  segmentation,  the complexity in the background and scale variability instances still remain challenging for instance segmentation of RSIs. To address the above problems, we propose an end-to-end multi-category instance segmentation model, namely Semantic Attention and Scale Complementary Network, which mainly consists of a Semantic Attention (SEA) module and a Scale Complementary Mask Branch (SCMB). The SEA module contains a simple fully convolutional semantic segmentation branch with extra supervision to strengthen the activation of interest instances on the feature map and reduce the background noise’s interference. To handle the under-segmentation of geospatial instances with large varying scales, we design the SCMB that extends the original single mask branch to trident mask branches and introduces complementary mask supervision at different scales to sufficiently leverage the multi-scale information. We conduct comprehensive experiments to evaluate the effectiveness of our proposed method on the iSAID dataset and the NWPU Instance Segmentation dataset and achieve promising performance.
	\end{abstract}
	
	

	%
	\IEEEpeerreviewmaketitle
	\section{Introduction}
	%
	%
	%
	%
	\IEEEPARstart{T}{hanks} to the rapid development in remote sensing technology, RSIs have become easily available and the understanding of RSIs has become a popular topic. Recently, many researchers commit to the scene classification \cite{SceneClassification1,SceneClassification2,SceneClassification3,SceneClassification5,SceneClassification4} and object detection \cite{RSIObjectDetection1,RSIObjectDetection2,RSIObjectDetection3,RSIObjectDetection4,hyperli,RSIObjectDetection6} in RSIs and achieve outstanding performance. In this paper, we concentrate on a new and challenging problem of instance segmentation in RSIs.

	Instance segmentation aims to classify the categories and predict the pixel-level results of each instance.  Contrary to the bounding-box annotation in object detection, instance segmentation delineates the boundary of each instance and results in a more accurate location. Benefiting from more accurate pixel-level information for each instance,  instance segmentation has great development potential in land planning, urban monitoring, and military reconnaissance.
	
\begin{figure}[t]
		\centering
		\subfloat[Ground Truth]{
			\begin{minipage}[t]{0.32\linewidth}
				\centering
				\includegraphics[width=1.1in, height=1.1in ]{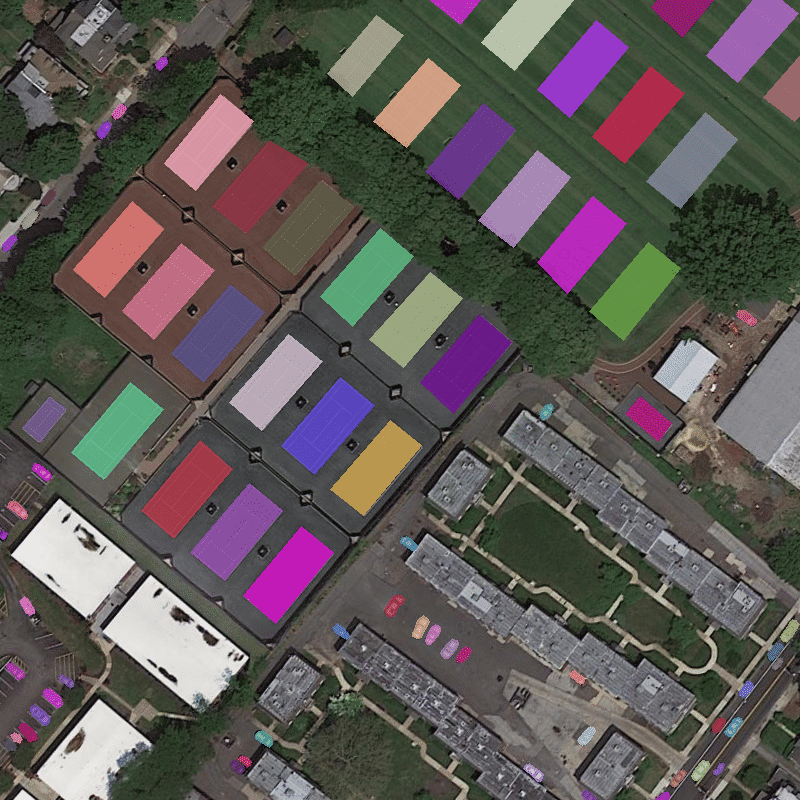}\\
				\vspace{0.2cm}
				\centering
				\includegraphics[width=1.1in, height=1.1in]{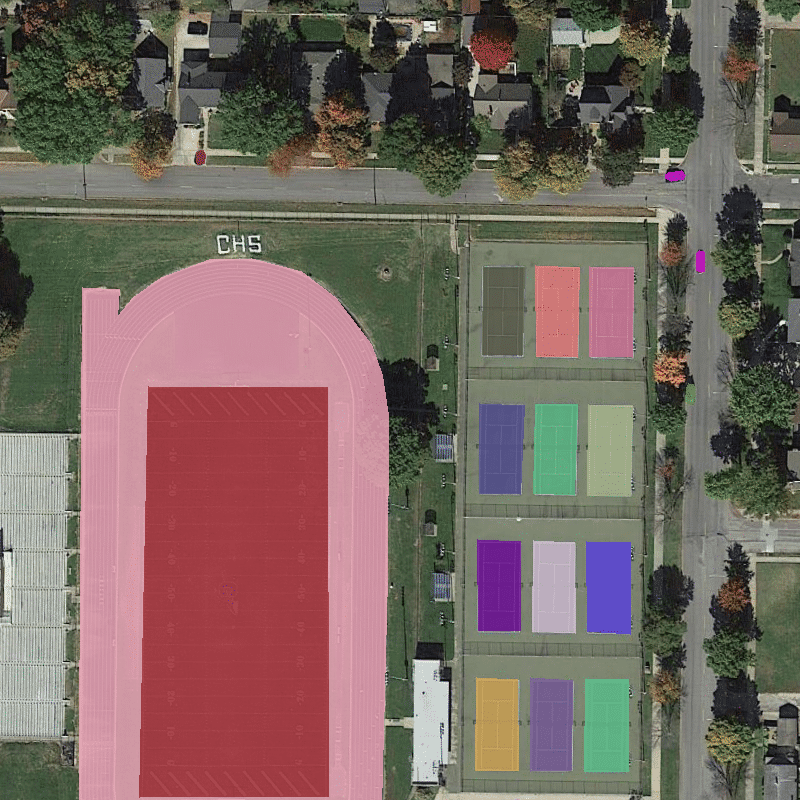}\\
			\end{minipage}%
		}%
		\subfloat[PANet]{
			\begin{minipage}[t]{0.32\linewidth}
				\centering
				\includegraphics[width=1.1in, height=1.1in]{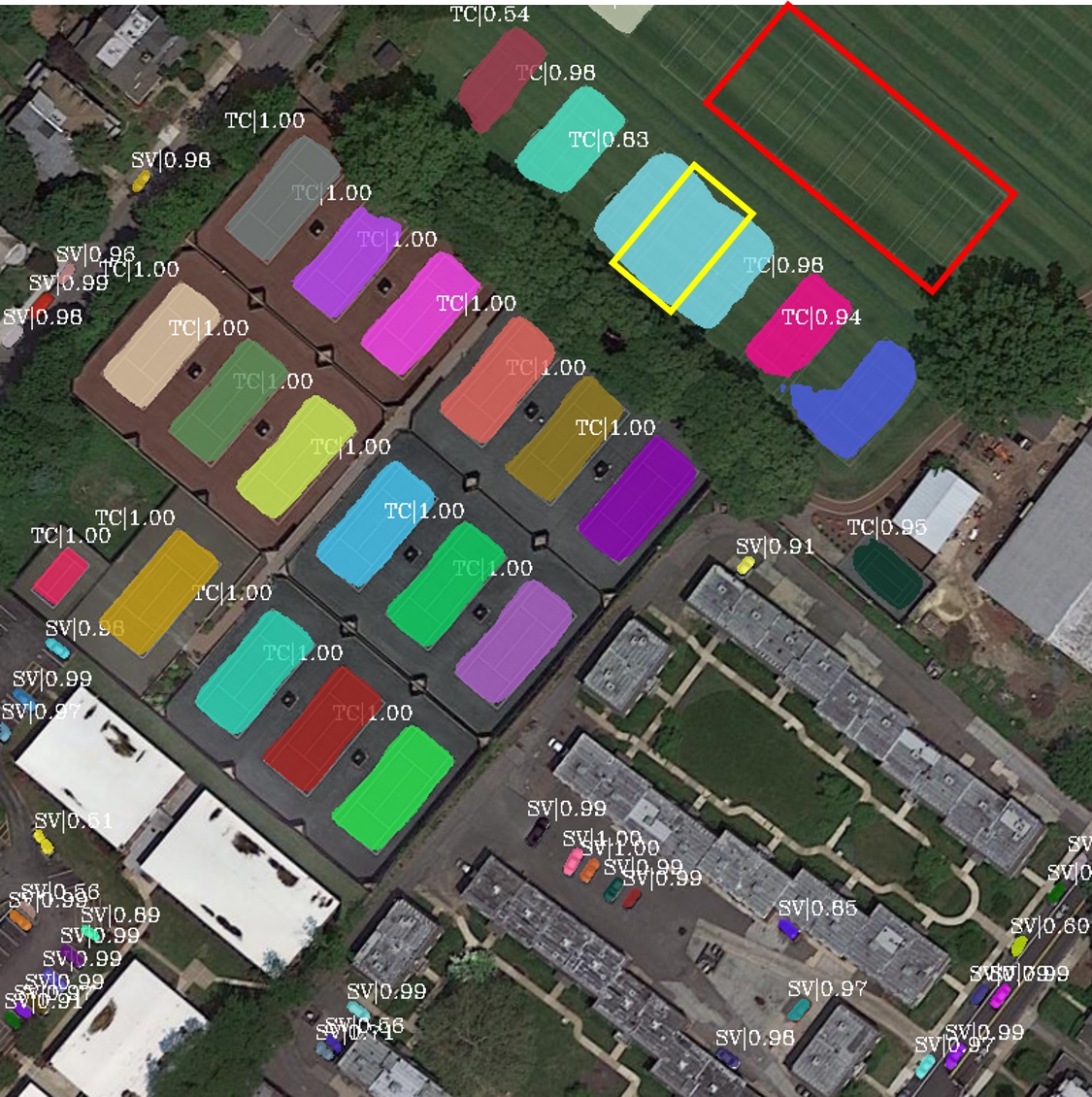}\\
				\vspace{0.2cm}
				\includegraphics[width=1.1in, height=1.1in]{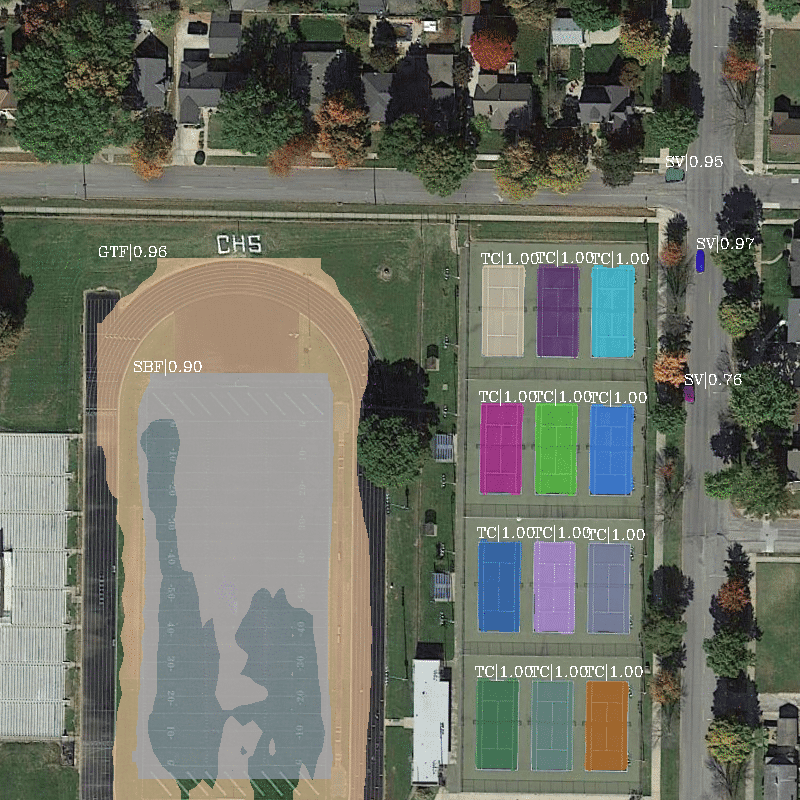}\\
			\end{minipage}%
		}%
		\subfloat[Ours]{
			\begin{minipage}[t]{0.32\linewidth}
				\centering
				\includegraphics[width=1.1in, height=1.1in]{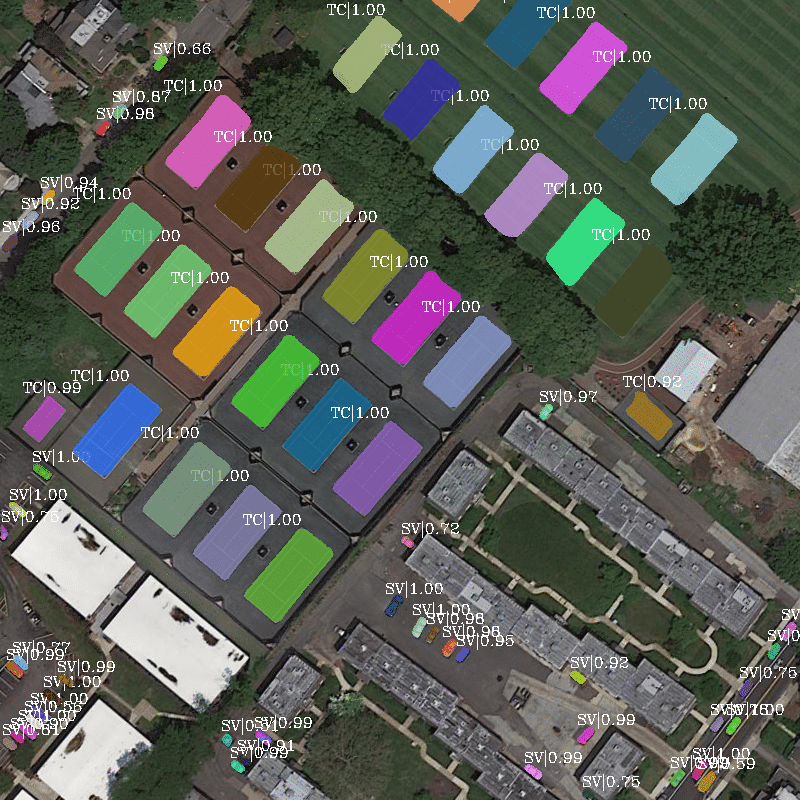}\\
				\vspace{0.2cm}
				\centering
				\includegraphics[width=1.1in, height=1.1in]{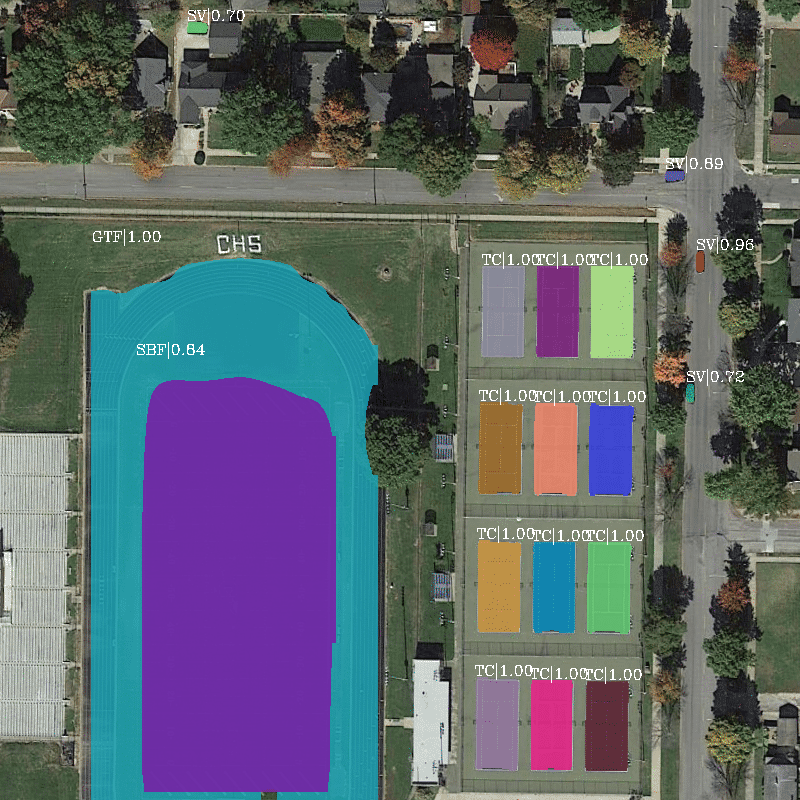}\\
			\end{minipage}%
		}%
		\caption{Problems of instance segmentation network designed for a natural scene in RSIs. The prediction of PANet may contain many false detections and under-segmentation. While our proposed network can better address these problems. The false prediction results and the miss prediction results are indicated by yellow and red rectangles, respectively. The bounding boxes are removed for simplicity.}
		\label{fig:data_show}
	\end{figure}

	In the past two decades, with the development of convolutional neural networks (CNN) \cite{ImageNet,Resnet}, many instance segmentation architectures \cite{FCIS, MaskRCNN,PANet,Mask-score,HTC,D2Det,CascadeR-CNN,DeepWatershed, PixelwiseInstanceSegmentation, SGN, InstanceCut, DiscriminativeLossFunction, SemanticInstanceSegmentationviaDeepMetricLearning, RecurrentPixelEmbedding, AssociativeEmbedding} have been proposed and achieved outstanding performance in the natural scene. However, few researchers \cite{SLC,DBLP:journals/tgrs/MouZ18,X-LineNet,Building,PreciseMask-RCNN,HQ-ISNet,GCP} focus on the instance segmentation in RSIs and the available methods just apply the instance segmentation models designed for the natural images to the RSIs, without taking into account the characteristics of RSIs such as the complex background and diversely scaled instances. Specifically, RSIs typically contain a highly complex background area that may interfere with the region of interest. As shown in the first row of Fig. \ref{fig:data_show}, tennis courts have a color similar to that of the surrounding grassland. When directly apply the off-the-shelf PANet \cite{PANet} on the RSIs, it has difficulties in separating the adjacent tennis courts or even causes miss detection. Besides, the geospatial instances can largely vary in scale, which leads to under-segmentation with the original single mask branch in \cite{PANet}. For example, the boundary of the ground track field is incomplete (see the second row of Fig. \ref{fig:data_show}).

	To alleviate the problem of complex background and large scale-variations of instances, we introduce the Semantic Attention (SEA) module and the Scale Complementary Mask Branch (SCMB) and design an end-to-end multi-category instance segmentation network for RSIs. For the SEA module, a new supervised semantic segmentation branch is proposed to strengthen the activation of the foreground instances and reduce the effect of the background noise. For the SCMB, a multi-scale structure is exploited to capture the complementary information at different scales to get more accurate segmentation results. We evaluate the proposed method on two public remote sensing datasets. Compared with the  other state-of-art approaches, our method achieves superior  performance.
	
	\begin{figure*}[t]
		\centering
		\includegraphics[scale=0.135]{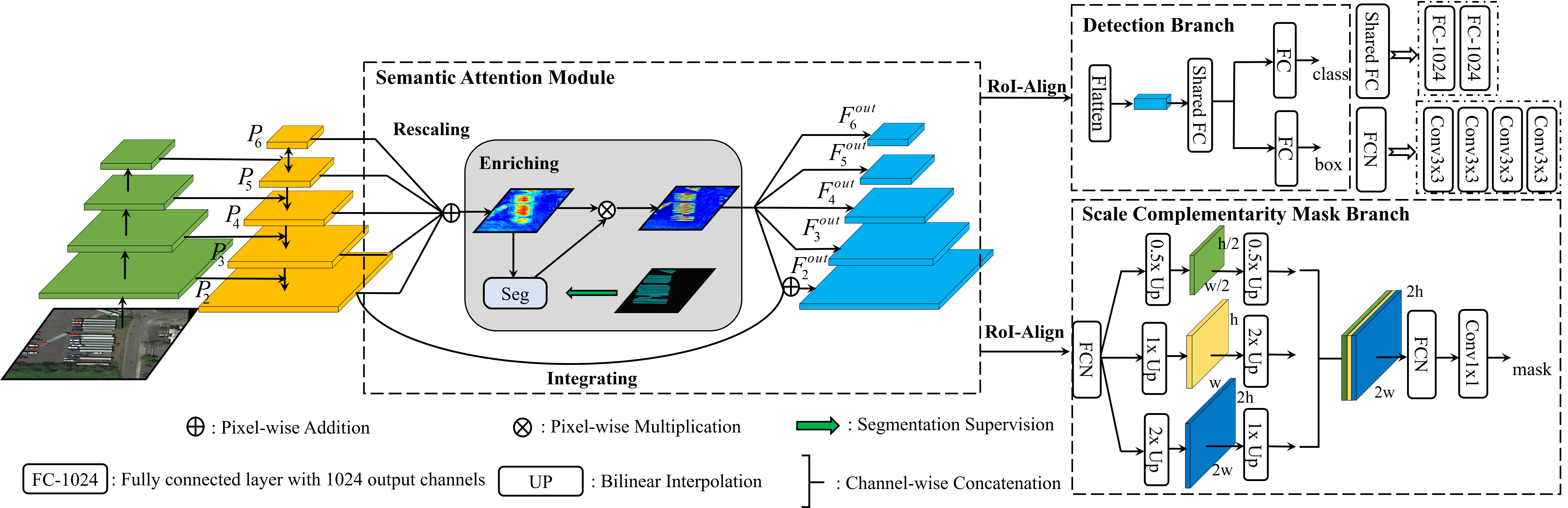}
		\caption{Overview of our instance segmentation network. It is based on the Mask-RCNN/PANet and adds the proposed Semantic Attention (SEA) module and Scale Complementary Mask Branch (SCMB). The SEA module consists of rescaling, enriching and integrating and introduces extra segmentation branch (details in Fig. \ref{fig:semantic segmentation branch}). For simplicity, we only represent the integrating operation at the $P_2$ level in the SEA module. The SCBM is composed
		of the trident mask branch, the scale complementarity guidance module and the feature fusion module (details in Fig. \ref{fig. SCMB}).}
		\label{fig:overall framework}
	\end{figure*}	
	Our contributions can be summarized as follows:
	
	1) We propose the Semantic Attention (SEA) module with semantic segmentation supervision and introduce it in the Feature Pyramid Network (FPN) to reduce the complex background interference on the feature maps. With the help of the SEA module, the network focuses on the instances’ regions and suppresses backgrounds.
	
	2) We extend the original single-scale mask branch into the Scale Complementary Mask Branch (SCMB) to deal with the under-segmentation problem caused by multiple scales of geospatial instances. The SCMB not only introduces scale complementary supervision to supervise the proposed trident mask branch but also fuses multi-scale feature maps to integrate information over multiple scales.
	
	3) The best performance is achieved in two challenging remote sensing instance segmentation datasets against the other state-of-the-art methods. The ablation studies show the effectiveness of each proposed module.
	
	The remainder of this paper is organized as follows. In Section II, we briefly introduce related work in instance segmentation methods on both natural scene and remote sensing community, semantic attention, and scale complementarity. In Section III, we describe our proposed method in detail. We report and discuss the experiments in Section IV. Finally, Section V concludes this paper.
	
	\section{RELATED WORK}
	\subsection{Instance Segmentation}
	Instance segmentation is mainly divided into proposal-based
	and proposal-free methods. Proposal-based methods are based on object detection frameworks. These methods first obtain the instances' proposals in the image through an object detector, then perform segmentation on each proposal to obtain its mask. Li \textit{et al.} \cite{FCIS} predicted position-sensitive inside/outside score maps and simultaneously rendered the instance mask and category with these score maps. He \textit{et al.} \cite{MaskRCNN} modified the Faster-RCNN \cite{Faster-RCNN} with a simple fully convolutional mask branch, which runs in parallel to the detection branch, to predict the mask of the proposals. Driven by the excellent performance of Mask-RCNN \cite{MaskRCNN}, the literature such as \cite{PANet, Mask-score, HTC}  have explored various extensions to Mask-RCNN. PANet \cite{PANet} adds a fully connected layer in the mask branch for accurate segmentation results. In \cite{MaskRCNN} \cite{PANet}, the classification confidence from the detection branch is used to measure the mask quality (i.e. the IoU between the instance mask and its ground truth). However, Huang \textit{et al.} \cite{Mask-score} found the mask quality cannot be well correlated with the classification confidence and presented the Mask-IoU block to learn the mask quality of predictions. Hybrid Task Cascade \cite{HTC} adopts a cascaded approach \cite{CascadeR-CNN}, where the mask features of the preceding stage are fed into the next stage for further improvements.
	
	Proposal-free methods are mainly built upon segmentation and aim at clustering pixel-level semantic class labels into different instances. Many researches \cite{DeepWatershed, PixelwiseInstanceSegmentation, SGN, InstanceCut} further transform semantic segmentation results into instances. Bai and Urtasun \cite{DeepWatershed} used the direction network and the watershed transform network to learn the energy map for watershed transform. InstanceCut \cite{InstanceCut} first predicts the semantic segmentation result with a semantic segmentation network and then uses an instance-awareness edge detector to obtain the instance segmentation results. Besides, several methods \cite{DiscriminativeLossFunction, SemanticInstanceSegmentationviaDeepMetricLearning, RecurrentPixelEmbedding, AssociativeEmbedding} map pixels into the embedding space for instance segmentation. Brabandere \textit{et al.} \cite{DiscriminativeLossFunction} introduced a new discriminative loss function that guides the network to pull the pixels that belong to the same instance while pushing away the pixels of different instances. Fathi \textit{et al.} \cite{SemanticInstanceSegmentationviaDeepMetricLearning} used deep metric learning to determine the similarity of the embedding points.
	
	 Despite the flourishing development of instance segmentation in the nature scene, there are only few studies \cite{SLC,DBLP:journals/tgrs/MouZ18,X-LineNet,Building,PreciseMask-RCNN,HQ-ISNet,GCP} in RSIs. Feng \textit{et al.} \cite{SLC} introduced the sequence local context module to address the confusion between densely arranged ships. Mou and Zhu \cite{DBLP:journals/tgrs/MouZ18} abandoned the detector-based method and decomposed the vehicle instance segmentation task into vehicle semantic segmentation and semantic boundary detection. HQ-ISNet \cite{HQ-ISNet} introduces the HR-FPN to maintain high-resolution feature maps in the network and designs a tiny network to refine the original mask branch. Liu \textit{et al.} \cite{GCP} embedded a global context parallel attention module into the anchor-free instance segmentation framework to capture the global information. Different from the methods \cite{SLC, DBLP:journals/tgrs/MouZ18, X-LineNet,Building} that only focused on a single category (e.g. ship, vehicle, aircraft, building, etc.), our proposed approach takes full account of the complex background and large scale-variance of instances in RSIs, and verifies the effectiveness of our network on a more challenging multi-category instance segmentation dataset.

	\subsection{Attention Mechanism}
	Recently, a number of works \cite{SENet, DANet, LibraR-CNN, ScarfNet, DES} have studied in the attention mechanism to facilitate different computer vision tasks. SENet \cite{SENet} designs an efficient Squeeze-and-Excitation (SE) block to adaptively re-weight channel-wise feature responses and achieves superior performance for image classification. Libra-RCNN \cite{LibraR-CNN} fuses the 5-level feature maps from FPN \cite{FPN} and uses a Gaussian Non-Local \cite{Non-Local} attention to obtain the balance semantic features. ScarfNet \cite{ScarfNet} generates features with strong semantic attention for each pyramid scale by bidirectional long short term memory (biLSTM) \cite{LSTM} and channel-wise attention. DES \cite{DES}, built upon SSD \cite{SSD}, adds an extra semantic attention branch supervised with weak segmentation ground-truth for semantic enrichment. In the remote sensing community, Zhang \textit{et al.} \cite{hyperli} designed the Channel and Spatial Attention Module to highlight the important features and suppress inessential features, which improves the performance in SAR ship detection. Yang \textit{et al.} \cite{SCRDet} proposed a Multi-Dimensional Attention Network to strengthen  the response of the region of interest. In contrast to \cite{LibraR-CNN, ScarfNet, SENet, DANet,hyperli}, our proposed method adds accurate supervision to guide the learning of the attention mechanism. Besides, our attention mechanism has a simple structure compared to the biLSTM in ScarfNet\cite{ScarfNet} and the atrous convolution in DES \cite{DES}. 
	
	\subsection{Scale Complementarity}
	The scale variation across instances is one of the most challenging problems in both natural scene images and RSIs. To alleviate this problem, many works explore the complementary information between the low-level and high-level features of CNN. In the natural scene, SSD \cite{SSD} sets different scales of default boxes in multiple layers and outputs the combination detection results of each layer.  FPN \cite{FPN} uses a top-down approach and horizontal connections to generate five-level features, and assigns each proportion of the proposal to the corresponding level. Considering the superior performance of FPN, many researchers \cite{HRNet,PANet,EfficientDet,CARAFE-FPN} have made further improvements to it. DeepLab \cite{DeepLab} proposes the atrous spatial pyramid pooling (ASPP) module to capture more scale information and make more accurate segmentation results. \cite{Track} proposes an online scale adaptive tracking approach by constructing a scale pyramid based on multi-layer convolutional features. In remote sensing fields, Azimi \textit{et al.} \cite{ICN} combined the image pyramid and the feature pyramid with the same resolution to detect the diverse scale geospatial objects. Deng \textit{et al.} \cite{RSIObjectDetection5} designed a multi-scale object proposal network (MS-OPN) with different receptive fields to generate different scales of proposals. Zhang \textit{et al.} \cite{Ship20} designed a lightweight scale share feature pyramid (SSFP) module to achieve high-speed and high-accurate multiscale SAR ship detection. The above method achieves complementary information at different scales at image-level \cite{ICN} or feature-level \cite{FPN,SSD,DeepLab,Track,Ship20}. Our method introduces label-level multi-scale information to improve the scale-invariant ability of the network.

	\section{METHODOLOGY}
	Our proposed network can be regarded as an extension of Mask-RCNN/PANet and the overall framework is illustrated in Fig. \ref{fig:overall framework}. First, we use CNN and FPN/PA-FPN\cite{PANet} to generate multi-scale feature maps of the given image. Then, we employ the SEA module to output the multi-scale feature maps with meaningful semantic information. Finally, the candidate proposals generated by the region proposal network (RPN)\cite{Faster-RCNN} and multi-scale semantic meaningful feature maps are sent to the detection branch and the proposed SCMB for the detection and segmentation. In the following section, we describe the details of the SEA module and SCMB.
	\subsection{Semantic Attention Module}
\begin{figure}[t]
		\centering
		\subfloat[(a)]{
			\begin{minipage}[b]{0.32\linewidth}
				\centering
				\includegraphics[width=1.1in, height=1.1in]{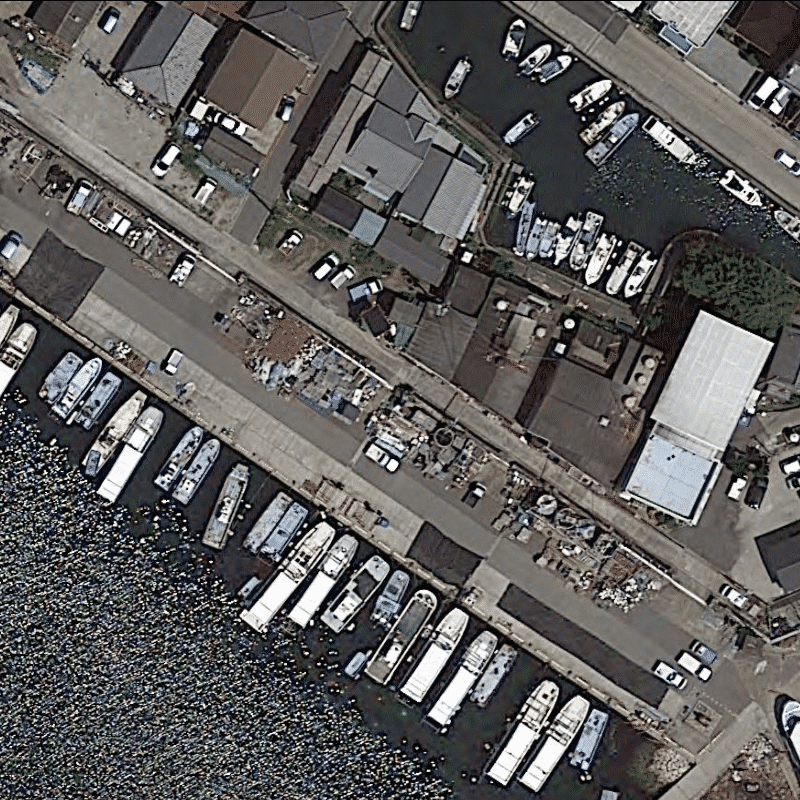}\\
				\vspace{0.2cm}
				\centering
				\includegraphics[width=1.1in, height=1.1in]{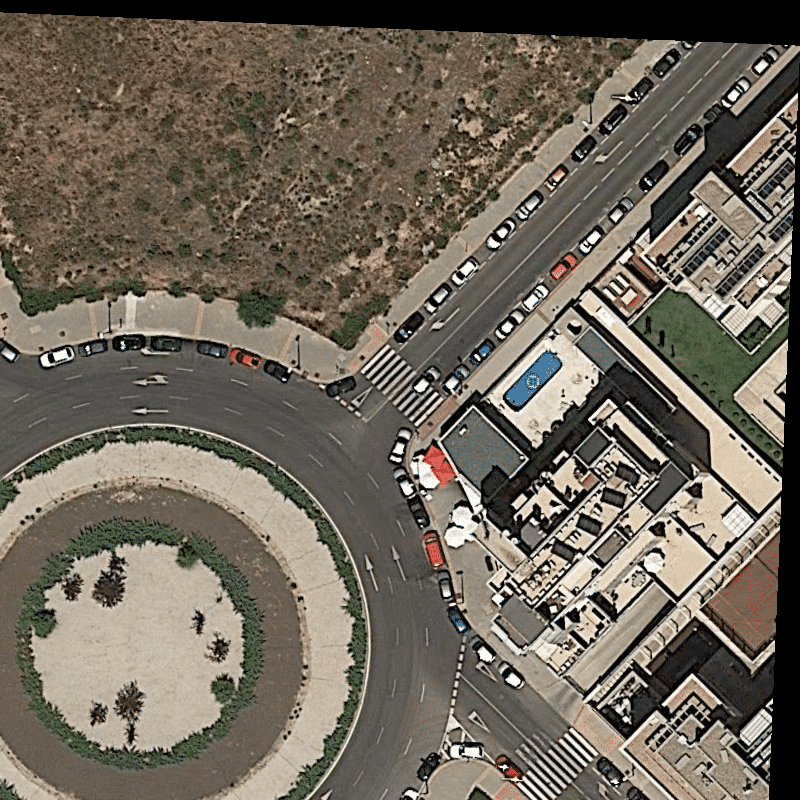}		
			\end{minipage}%
		}%
		\subfloat[(b)]{
			\begin{minipage}[b]{0.32\linewidth}
				\centering
				\includegraphics[width=1.1in, height=1.1in]{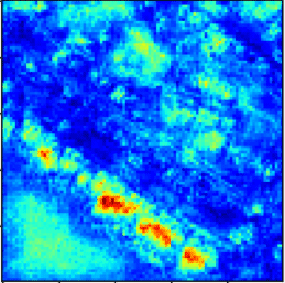}\\
				\vspace{0.2cm}
				\centering
				\includegraphics[width=1.1in, height=1.1in]{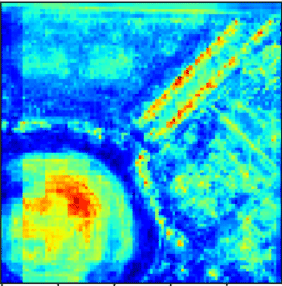}
			\end{minipage}%
		}%
		\subfloat[(c)]{
			\begin{minipage}[b]{0.32\linewidth}
				\centering
				\includegraphics[width=1.1in,height=1.1in]{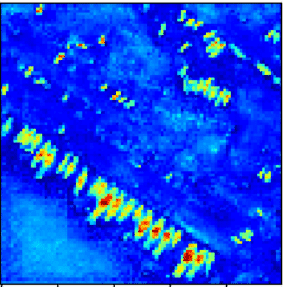}\\
				\vspace{0.2cm}
				\centering
				\includegraphics[width=1.1in,height=1.1in]{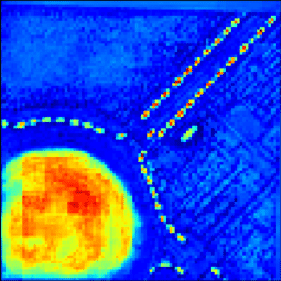}			
			\end{minipage}%
		}%
		\caption{Visualization of the feature map. (a) Input image. (b) Feature map of original FPN. (c) Feature map of the FPN with proposed SEA module}
		\label{fig. show feature map}
	\end{figure}

	As shown in Fig. \ref{fig. show feature map}(b), the feature maps obtained by FPN contain complex background information, which may result in false predictions. Thus, we propose the SEA module that introduces the semantic segmentation supervision to enhance the activation of instances and reduce the responses of noises. Many semantic attention modules \cite{ScarfNet,LibraR-CNN,DES} have been published to enrich the semantic information of feature maps. Different from these methods, the SEA module adds semantic segmentation supervision and has a simple and straightforward architecture.
	\begin{figure*}[t]
		\centering
		\includegraphics[scale=0.15]{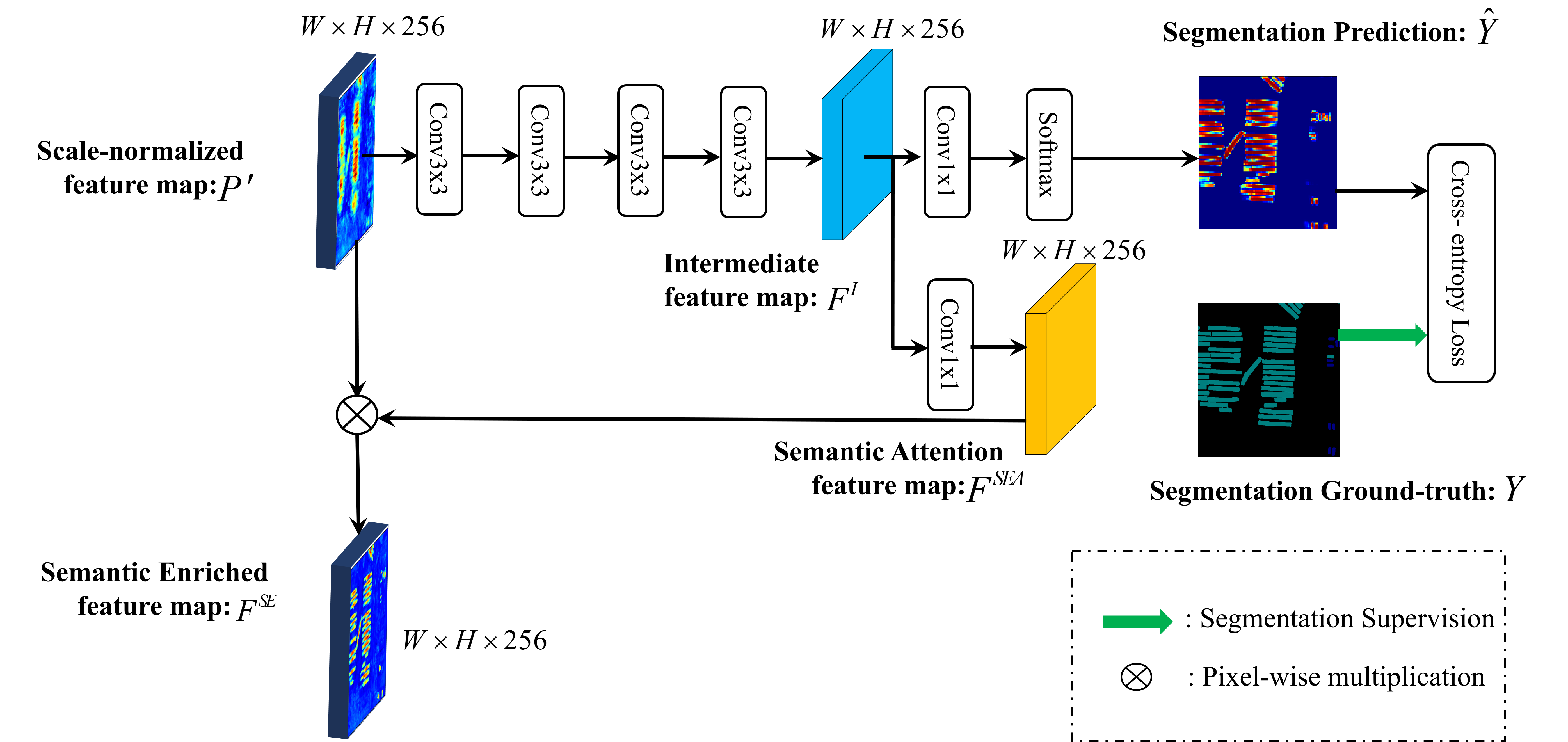}
		\caption{The detailed architecture of the segmentation branch in the SEA module. The segmentation branch takes the scale-normalized feature map $P^{\prime}$ as the input, and a series of convolutional layers is adopted to obtain the intermediate feature map $F^{I}$. Then, $F^{I}$ is broadcast into two streams to generate the semantic attention feature map $F^{SEA}$ and segmentation prediction $\hat{Y}$. Finally, the element-wise multiplication is applied between $P^{\prime}$ and $F^{SEA}$ to generate the semantic enriched feature map $F^{SE}$.}
		\label{fig:semantic segmentation branch}
	\end{figure*}
	\begin{figure*}[t]
		\centering
		\includegraphics[scale=0.13]{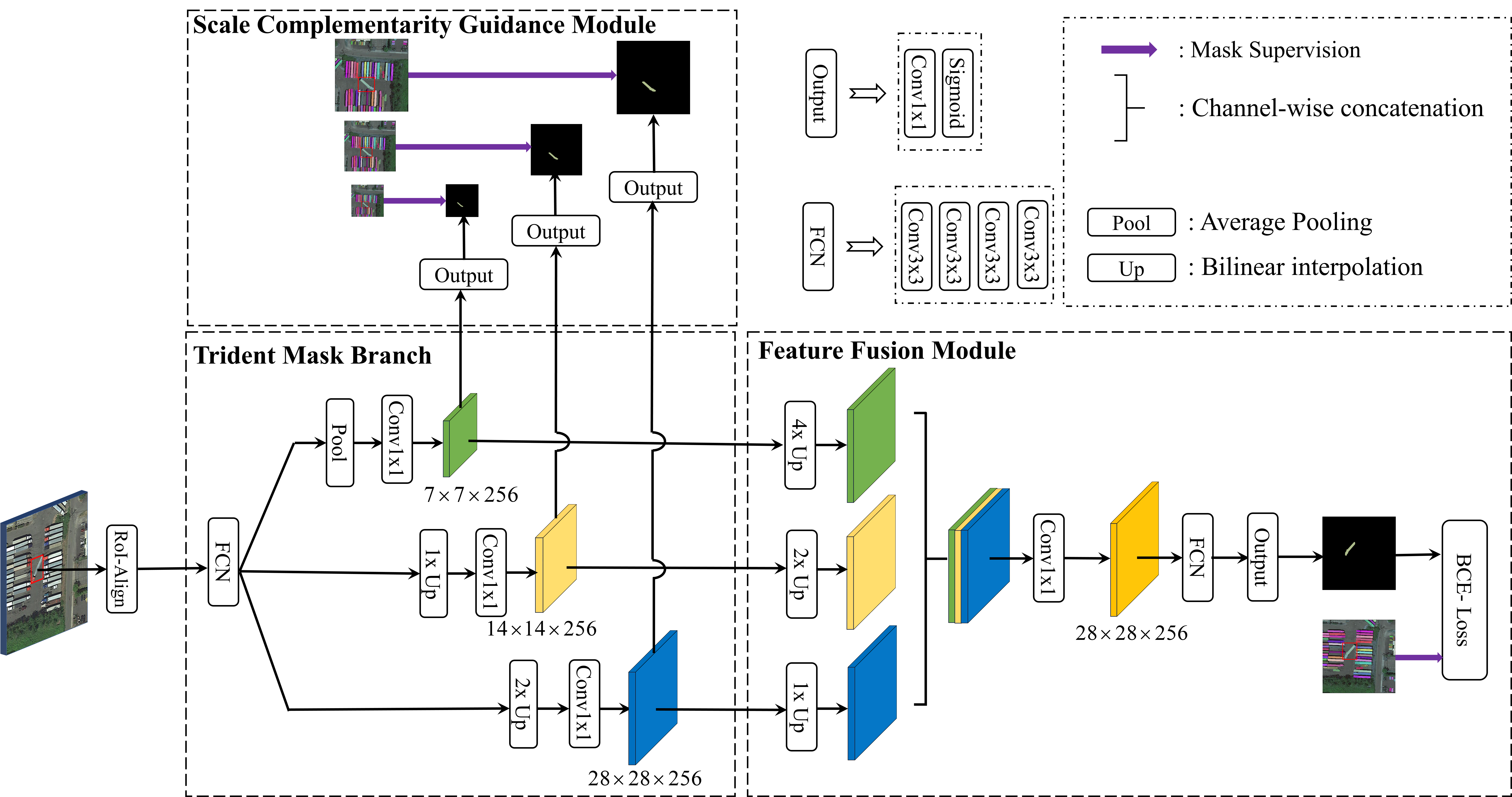}
		\caption{Illustration of Scale Complementary Mask Branch. The feature maps produced by RoI-Align layer are resized to the other two scales using the average pooling layer and the bilinear interpolation. After that, the three scale feature maps are fed to Scale Complementary Guidance (SCG) Module and Feature Fusion (FF) module. The SCG module embeds the corresponding scale mask supervision to improve the discrimination of each scale feature map. The FF module fuses three scale feature maps for precision segmentation. For simplicity, we do not show the Binary Cross-Entropy (BCE) Loss of the three guidance paths in the SCG Module.}
		\label{fig. SCMB}
	\end{figure*}
	
	For the semantic segmentation’s ground truth, we generate it using a simple transformation strategy: Given an image, if a pixel belongs to an instance, we assign the class label of this instance to the pixel, otherwise,  we set the pixel to the background class. An example of the semantic segmentation's ground truth is shown in Fig. \ref{fig. prediction}(b).
	
	The pipeline of the SEA module is shown in Fig. \ref{fig:overall framework}. There are mainly three steps, including rescaling, enriching and integrating.
	\subsubsection{Rescaling}
	Following the definition of FPN, we use \{$P_2$, $P_3$, $P_4$, $P_5$, $P_6$\} to define the 5-level output feature maps with different strides of \{4, 8, 16, 32, 64\} pixels corresponding to the original image. To enrich the semantic information of the above 5-level multi-scale feature maps, we first resize these feature maps to a uniform scale, i.e. the corresponding scale of $P_3$ (Ablation study illustrates why we choose $P_3$). Here, we use bilinear interpolation and average pooling layer to generate resized feature maps \{${P_2}^{\prime}$, ${P_3}^{\prime}$, ${P_4}^{\prime}$, ${P_5}^{\prime}$, ${P_6}^{\prime}$\}. Then, we can obtain the scale-normalized feature map by:
	\begin{equation}
	P^{\prime}=\frac{1}{5} \sum_{i=2}^{6} P_{i}^{\prime}
	\label{Eq.1}
	\end{equation}
	\subsubsection{Enriching}
	 In contrast to early researches \cite{LibraR-CNN} \cite{ScarfNet} that implemented the semantic enrichment in an unsupervised way, we consider embedding the semantic segmentation supervision is an intuitive approach to enrich the semantic information. We design a tiny fully convolutional semantic segmentation branch, including four convolutional layers with a $3 \times 3 $ kernel and two convolutional layers with a $1 \times 1 $ kernel. We set the filter number to 256 for the four $3 \times 3 $ convolutional layers. For the two $1 \times 1 $ convolutional layers, one is set to $C+1$ for prediction where $C$ is the number of the classes and the other is 256 for semantic attention.
	
	As shown in Fig. \ref{fig:semantic segmentation branch}, the proposed semantic branch takes the scale-normalized feature map $P^{\prime}$ as input, and the four $3 \times 3 $ convolutional layers first extract the scale-normalized feature map $P^{\prime}$ to get the intermediate feature map $F^{I}$. Then, $F^{I}$ is attached to two streams, called prediction and attention streams.

	In the attention stream, we append a $1 \times 1 $ convolutional layer to obtain the semantic attention feature map $F^{SEA}$, and then multiply $F^{SEA}$ with the original scale-normalized feature map $P^{\prime}$ to generate the semantic enriched feature map $F^{SE}$. Thus, the  generation of the semantic enriched feature map $F^{SE}$ is as follows:
	\begin{equation}
	F^{I} = \textit{{Extraction}}\left(P^{\prime}; \theta_{E}\right)
	\end{equation}
	\begin{equation}
	F^{SEA} = Attention_{-}Stream\left(F^{I}; W_{A}\right)
	\end{equation}
	\begin{equation}
	F^{SE}=P^{\prime} \odot F^{SEA}
	\end{equation}
	where $\textit{{Extraction}}\left(*; \theta_{E} \right)$ represents the four $3 \times 3 $ convolutional layers with parameter $\theta_{E}$.  $Attention_{-}Stream\left(*; W_{A}\right)$ is the convolutional layer in the attention stream and  $W_{A}$ denotes the weights of the convolutional layer.
	
	The prediction stream contains a $1 \times 1 $ convolutional layer with $C+1$ output channels and a softmax layer aiming to produce semantic segmentation prediction $\hat{Y} \in {H \times W}$:
	\begin{equation}
	\hat{y}_{i j}^{c}=\frac{\exp \left(f_{i j}^{c}\right)}
	{\sum_{k=0}^{C} \exp \left(f_{i j}^{c}\right)}\\
	\end{equation}
	where $ f = Conv_{1 \times 1}\left(I; W_{1 \times 1} \right)$ and $\hat{y}_{i j}^{c}$ measures the probability that the pixel in $i$th row and $j$th column belongs to the category $c$. We define the loss function $L_{segmentation}$ as:
	\begin{equation}
	L_{segmentation}=-\frac{1}{H \cdot W}\sum_{\hat{y_{i j}} \in \hat{Y} \atop y_{i j} \in Y} \sum_{k=0}^{C} y_{i j}^{k} \log \left(\hat{y}_{i j}^{k}\right)
	\end{equation}
	where $Y$ denote the ground truth of semantic segmentation.
	
	\subsubsection{Integrating}
	After the enriching step, the semantic enriched feature map is resized to different scales corresponding to \{$P_2$, $P_3$, $P_4$, $P_5$, $P_6$\}, and we denote these generated feature maps as \{$F^{SE}_2$, $F^{SE}_3$, $F^{SE}_4$, $F^{SE}_5$, $F^{SE}_6$\}. Similar to \cite{LibraR-CNN} \cite{FPN}, we deploy the skip connection to integrate feature maps $F^{SE}_{i}$ and the original feature maps $P_{i}$, which can sufficiently leverage original information and enrich semantic information. The integrated operation can be represented as:
	\begin{equation}
	F^{out}_{i}=F^{SE}_{i}+P_{i}
	\end{equation}
	
	With the above three steps, the output multi-scale feature maps with meaningful semantic information \{$F^{out}_2$, $F^{out}_3$, $F^{out}_4$, $F^{out}_5$, $F^{out}_6$\} can be used for the following RPN and RCNN modules. Significantly, our proposed SEA module can be well embedded in FPN to effectively identify the instance regions on feature maps, and it can be easily applied to other computer vision tasks.
	
	\subsection{Scale Complementary Mask Branch}
	Because the scale variations of the instances in RSIs are generally larger than that of natural scene images, the original single scale mask branch \cite{MaskRCNN} \cite{PANet} may lead to under-segmentation, as shown in the second row of Fig. \ref{fig:data_show}. Inspired by the studies \cite{Triedentnet} \cite{PSPNet} that fuse the multi-scale information to remedy the weakness of single-scale network, we introduce the SCMB to alleviate the under-segmentation problem. Specifically, we replace the single-scale mask branch with a trident mask branch and generate scale complementary mask supervision for the corresponding branch. Besides, a feature fusion module is designed to facilitate the combination of different scale features.
	
	The detailed architecture of the SCMB is shown in Fig. \ref{fig. SCMB}, including Trident Mask Branch (TMB), Scale Complementary Guidance (SCG) module and Feature Fusion (FF) module.
	\subsubsection{Trident Mask Branch}
	Our trident mask branch is an extension of the original one \cite{MaskRCNN}. In \cite{MaskRCNN}, given the Region of Interest (RoI) feature map, the mask branch employs a tiny fully convolutional network (FCN) with parameter $\theta_{1}$ and a deconvolutional layer with the upsampling ratio of 2 to predict a binary pixel-wise mask for each class independently. The binary prediction is presented as follows:
	\begin{equation}
	Pred = Sigmoid\left(Deconv\left(
	\operatorname{FCN}\left(RoI;\theta_{1} \right)\right); 2\right)
	\end{equation} 
	where 
	\begin{equation}
	Sigmoid(x) = \frac{1}{1+e^{-x}}
	\end{equation}
	
	Considering the absence of multi-scale information in the original single-scale mask branch, we transform it into the trident form, as shown in the TMB of Fig. \ref{fig. SCMB}. Following  \cite{MaskRCNN}, the TMB first applies a tiny FCN to extract the feature map of each RoI. Different from \cite{MaskRCNN}, we use bilinear interpolation and average pooling layer to upsample ($28 \times 28$) and downsample ($7 \times 7$) the feature map and keep the scale of the original feature map ($14 \times 14$), resulting in three different scales of feature maps \{$F_1$, $F_2$, $F_3$\}. To reduce the computational overhead, we adopt a $1 \times 1 $ convolutional layer to shrink output channels to half for each scale of the feature map. The program of our Trident Mask Branch is as follows:
	\begin{equation}
	F_{1} = Conv_{1 \times 1}\left(Avg_{-}Pooling\left(
	\operatorname{FCN}\left(RoI;\theta_{s}\right)\right) ;W_{1}\right)
	\end{equation}
	\begin{equation}
	F_{2} = Conv_{1 \times 1}\left(
	\operatorname{FCN}\left(RoI;\theta_{s}\right) ;W_{2}\right)
	\end{equation} 
	\begin{equation}
	F_{3} = Conv_{1 \times 1}\left(Up\left(
	\operatorname{FCN}\left(RoI;\theta_{s}\right); 2\right) ;W_{3} \right)
	\end{equation} 
	where $Up\left(* ;2 \right)$  and $Avg_{-}Pooling\left(* ;2 \right)$ represent bilinear interpolation and average pooling layer, respectively. $\operatorname{FCN}\left(* ;\theta_{s} \right)$ denotes the weight shared FCN with parameter $\theta_{s} $ and $ Conv_{1 \times 1}\left(* ;W_{i} \right)$ denotes the convolutional layer with parameter $W_{i}$ for the computational reduction in each branch.
	
	\subsubsection{Scale Complementary Guidance Module}
	In order to obtain discriminative feature maps at each scale, we introduce the scale complementary guidance module composed of three guidance paths \{$GP_1$, $GP_2$, $GP_3$\}. In each path, we adopt a $1 \times 1 $ convolutional layer to produce the prediction and embed the corresponding scale mask supervision. The prediction ($Pred_{GP}^{(i)} \in {H_{GP}^{(i)} \times W_{GP}^{(i)}}$ ) of each guidance path is denoted as:
	\begin{equation}
	Pred_{GP}^{(i)}= Sigmoid\left(Conv_{1 \times 1}\left(
	F_{i}; W_{G P_{i}}\right)\right)
	\end{equation} 
	
	In all the three guidance paths, we use the binary cross-entropy which could be defined as:  
	\begin{equation}
	\begin{aligned}
	L_{GP}^{(i)}=-\frac{1}{{H_{GP}^{(i)} \cdot W_{GP}^{(i)}}}\sum_{\hat{y} \in Pred_{GP}^{(i)} \atop y \in S_{GP}^{(i)}}[y \log \hat{y}+(1-y) \log (1-\hat{y})] 
	\end{aligned}
	\label{Eq.BCE}
	\end{equation}
	where $i \in[1,2,3]$ and $S_{GP}^{(i)}$ denotes the mask supervision in the $i$-th guidance path. Thus the total loss of this module could be denoted as:
	\begin{equation}
	L_{SCG}=L_{GP}^{(1)}+L_{GP}^{(2)}+L_{GP}^{(3)}
	\end{equation}
	\subsubsection{Feature Fusion Module}
	The goal of the feature fusion module is to integrate the information at different scale feature maps for precise segmentation. For the three different spatial resolution feature maps generated from TMB, we upsample the two low-resolution feature maps to $28 \times 28$ using bilinear interpolation. Then the three feature maps are merged by channel-wise concatenating. Finally, we append four consecutive convolutional layers consisting of $3 \times 3$ kernel sizes and a $1 \times 1$ convolutional layer to produce the binary prediction for each class. The binary prediction can be denoted as:
	\begin{equation}
	F_{fusion} = Concat\left( 
	Up\left(F_{1}; 4\right), Up\left(F_{2}; 2\right), F_{3}
	\right) 
	\end{equation} 
	\begin{equation}
	Pred = Sigmoid\left(  FE\left( F_{fusion}; \theta_{FE}  \right)  \right)
	\end{equation} 
	where $Up\left(* ;2 \right)$ and $Up\left(* ;4 \right)$ denote the bilinear interpolation.  $FE\left(* ;\theta_{FE} \right)$ represents four consecutive convolutional layers.
	
	We also use the binary cross-entropy to calculate the $L_{FF}$, which has the same form as Eq. (\ref{Eq.BCE}). The loss function of the overall SCMB is denoted as:
	\begin{equation}
	L_{SCMB}=L_{SCG}+L_{FF}
	\end{equation}
	\subsection{Joint Loss Function}
	Our proposed approach is an end-to-end instance segmentation network and the joint loss function $L_{total}$ consists of three parts: $L_{detection}$, $L_{segmentation}$ and $L_{SCMB}$. Thus, the joint loss function could be expressed as:
	\begin{equation}
	L_{\text {total}}=\alpha_{1} L_{detection}+\alpha_{2} L_{segmentation}+\alpha_{3} L_{SCMB}
	\end{equation}
	
	In this paper, we set the loss weights $\alpha_{1}$, $\alpha_{2}$, and $\alpha_{3}$ to 1. The previous work \cite{HybirdLoss,MultiLoss} has demonstrated that a good choice of the loss weights can further improve the performance, which will be our future research.
	
	\section{EXPERIMENTS}
	\subsection{Evaluation Datasets}
	\subsubsection{iSAID}The iSAID \cite{iSAID} dataset is a new open benchmark dataset for multi-categories instance segmentation in RSIs. The dataset consists of 2,806 images with different sizes (from 800 to 13,000 in width) and 655,451 annotated instances. There are 15 common object categories in the dataset, including large vehicle (LV), small vehicle (SV), storage tank (ST), plane (PL), ship (SH), swimming pool (SW), harbor (HA), tennis court (TC), ground track field (GTF), soccer-ball field (SBF), baseball diamond (BD), bridge (BR), basketball court (BC) roundabout (RA) and helicopter (HC). The whole dataset is split into three parts: 1/2 for training, 1/6 for validation and 1/3 for testing. The ground truth of the training set and the validation set are available.
	
	Due to the large spatial resolution of the original images, we crop the original images into $800 \times 800$ patches with a stride set to 200 by the official provided toolkit\footnote{https://github.com/CAPTAIN-WHU/iSAID\_{Devkit}} and acquire 28,249 images for training set, 9,581 for validation set and 19,377 for test set.
	
	\subsubsection{NWPU VHR-10 instance segmentation}The NWPU VHR-10 instance segmentation dataset \cite{PreciseMask-RCNN} is an extension of the remote sensing object detection dataset NWPU VHR-10 \cite{NWPUv1}. This instance segmentation dataset includes 650 and the spatial size of images ranges from $ 533 \times 597$ to $ 1,728 \times 1,028$ pixels. This dataset contains 10 object categories, including airplane, baseball diamond, basketball court, bridge, ground track field, harbor, ship, storage tank, tennis court, and vehicle.
	
	In the experiments, we randomly select 70\% of the image set (i.e. 454 images) as the training set and the rest of the positive set (i.e. 196 images) as the test set.
	
	\begin{table*}[t]  
	\centering
	\setlength{\abovecaptionskip}{0pt}
	\setlength{\belowcaptionskip}{10pt}
	\caption{Effects of SEA Module and SCMB. All Models are Evaluated on iSAID Validation Set}
	\label{table-I}	
	\begin{tabular}{c|c|cc|ccc|ccc|c|c|c}  			
	\toprule   			
	Model&Backebone&SEA&SCMB & $AP^{m}$ & $AP_{50}^{m}$ & $AP_{75}^{m}$ & $AP^{b}$ & $AP_{50}^{b}$ & $AP_{75}^{b}$ & FPS & Params & Model Size \\  			
	\midrule   	
	\multirow{4}*{PANet \cite{PANet}} &\multirow{4}*{ResNet-101}
	&             &            		&38.1 &62.8 &40.5 & 43.9 & 67.0 & 48.3 & \textbf{5.1} & 85.14M & 682.6MB\\  
	&&	\checkmark &  	            &38.6 &63.4 &41.0 & \bf{44.7} & 68.1 & \bf{49.5} & 4.5 & 89.08M & 709.2MB\\ 
	&&	           & \checkmark  	&38.8 &62.9 &41.6 & 44.0 & 67.9 & 48.3 & 4.3 & 87.44M & 701.1MB\\    
	&&	\checkmark & \checkmark 	&\bf{39.5} &\bf{64.1} &\bf{42.1} & 44.6 & \bf{68.5} & 48.4 & 3.6 &91.38M & 727.6MB\\  	
		\bottomrule 
	\end{tabular}	
\end{table*}

\begin{figure*}[!t]
	\subfloat[Ground Truth]{
		\begin{minipage}[t]{0.16\linewidth}
			\centering
			\includegraphics[width=1.1in, height=1.1in ]{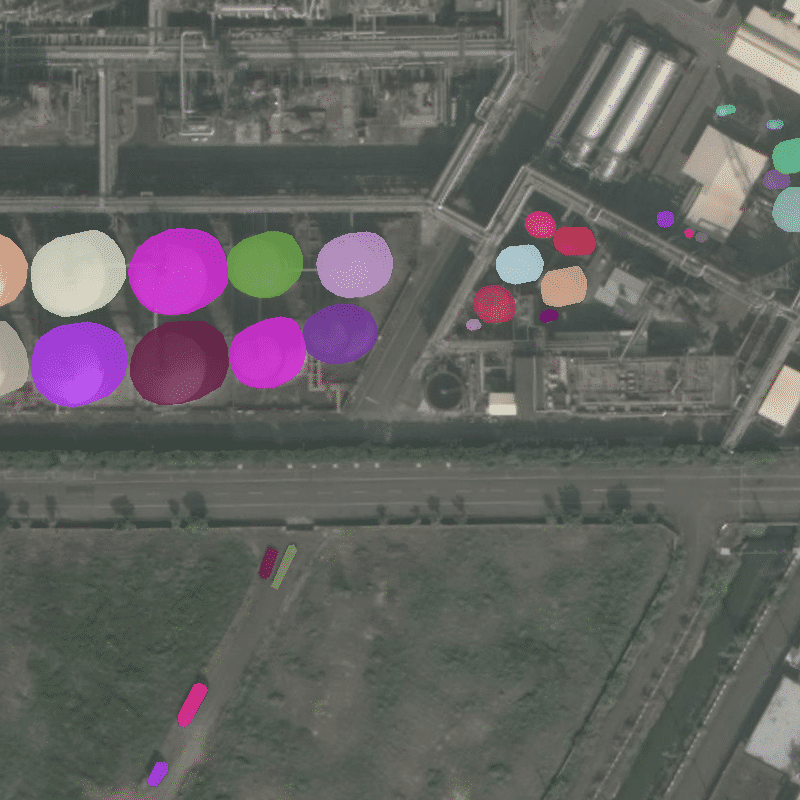}\\
			\vspace{0.2cm}
			\centering
			\includegraphics[width=1.1in, height=1.1in]{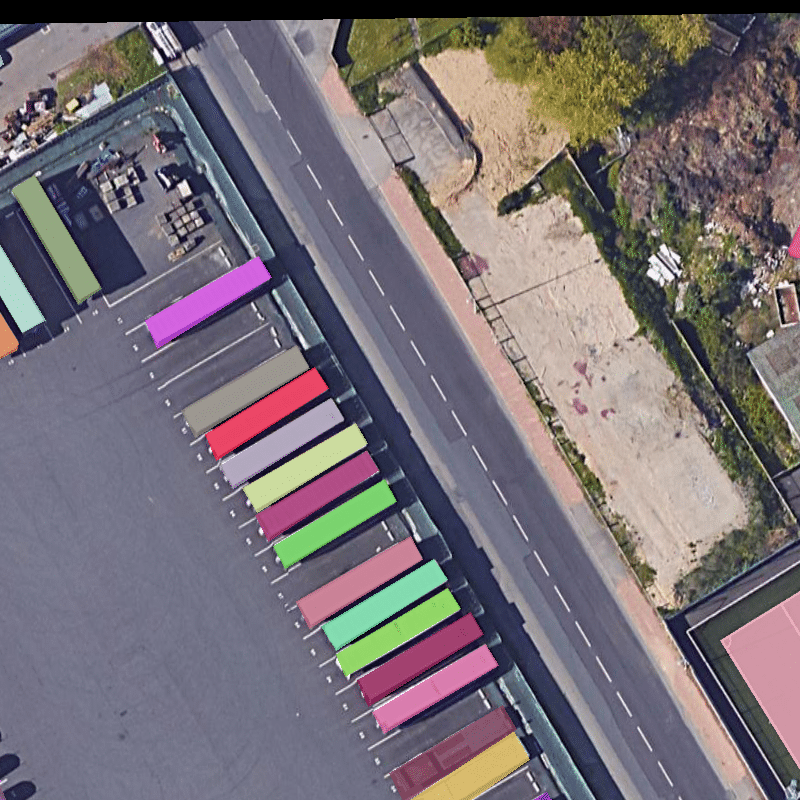}\\
		\end{minipage}%
	}%
	\subfloat[PANet]{
		\begin{minipage}[t]{0.16\linewidth}
			\centering
			\includegraphics[width=1.1in, height=1.1in]{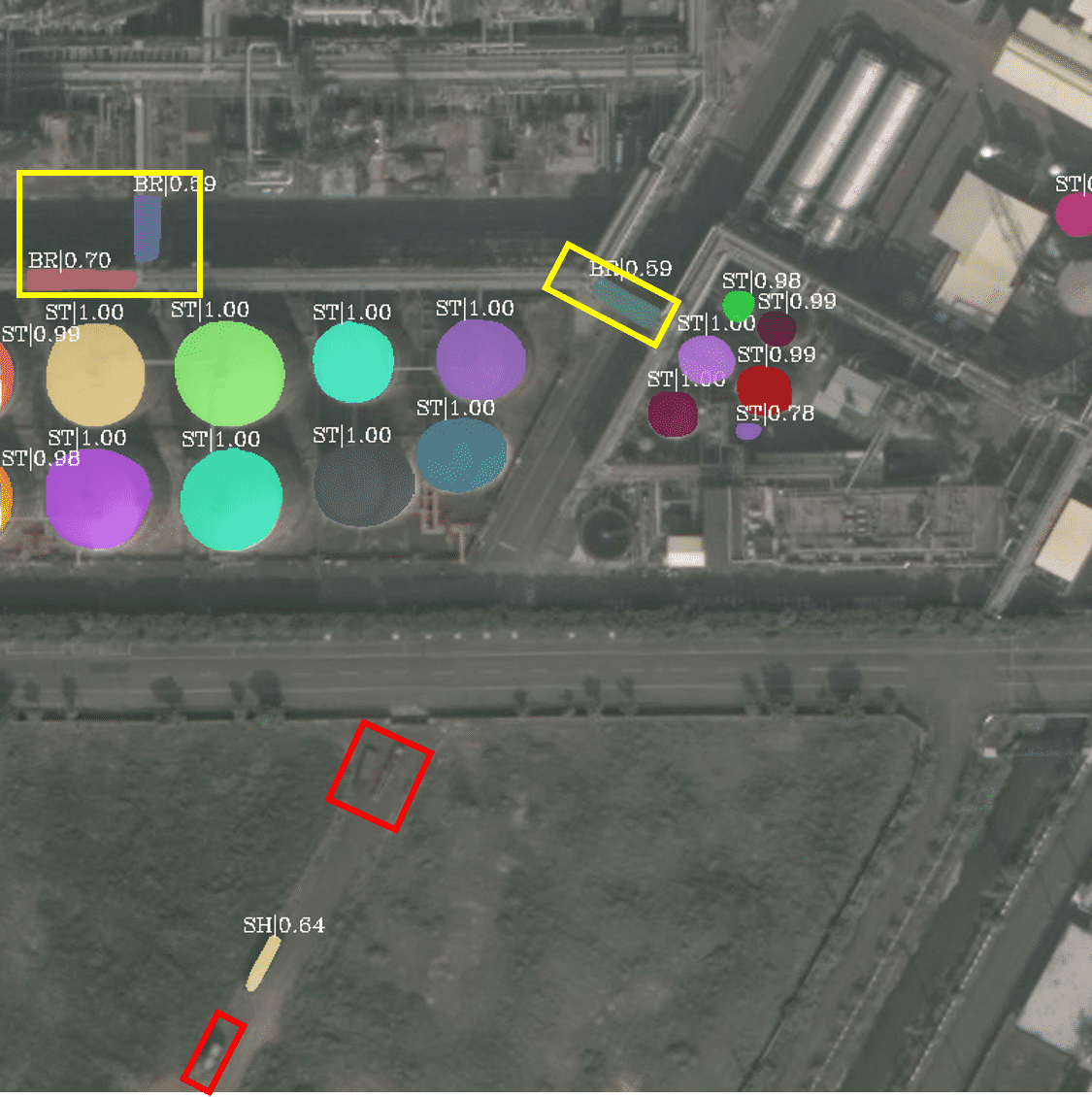}\\
			\vspace{0.2cm}
			\includegraphics[width=1.1in, height=1.1in]{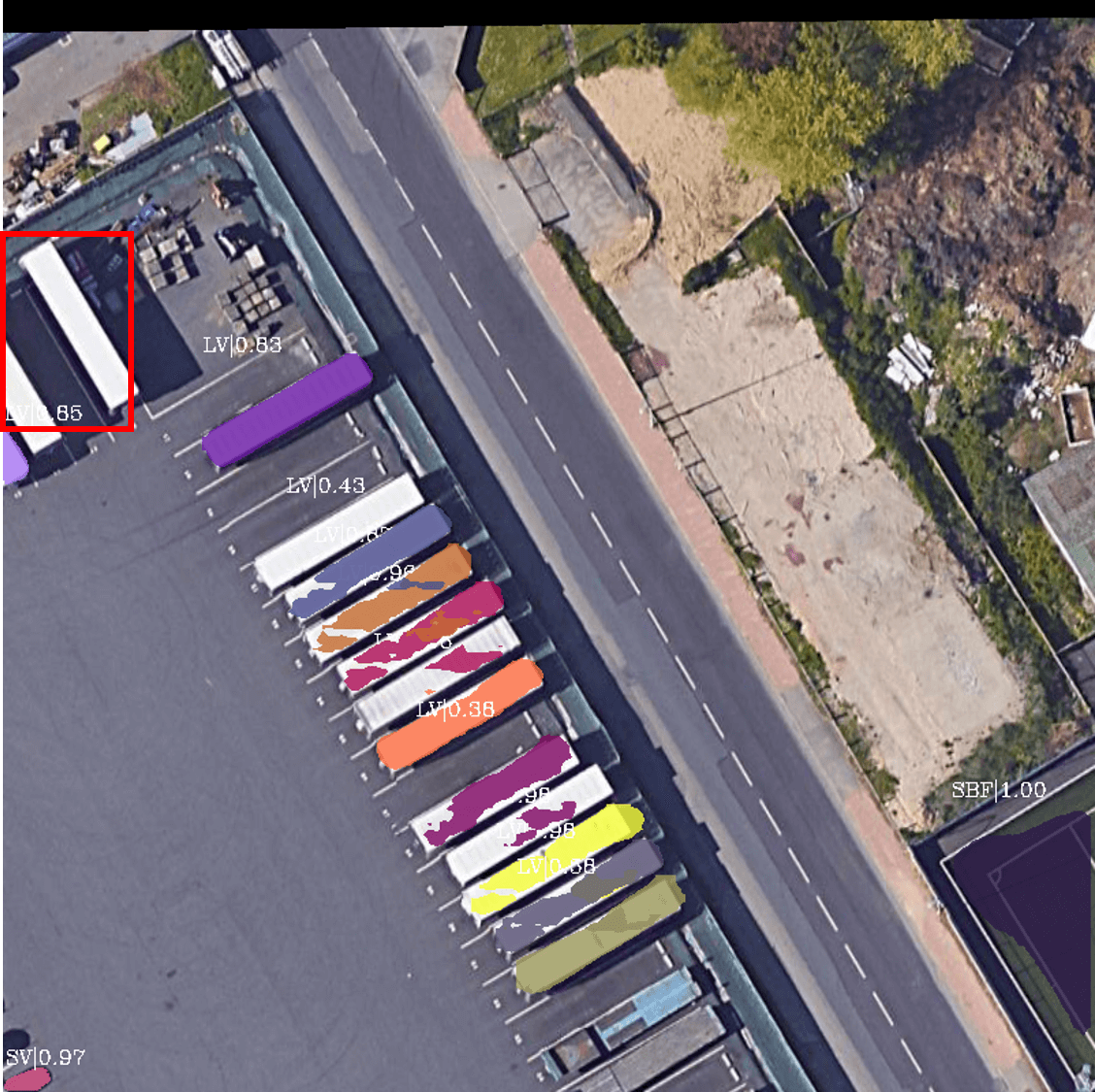}\\
		\end{minipage}%
	}%
	\subfloat[Ours]{
		\begin{minipage}[t]{0.16\linewidth}
			\centering
			\includegraphics[width=1.1in, height=1.1in]{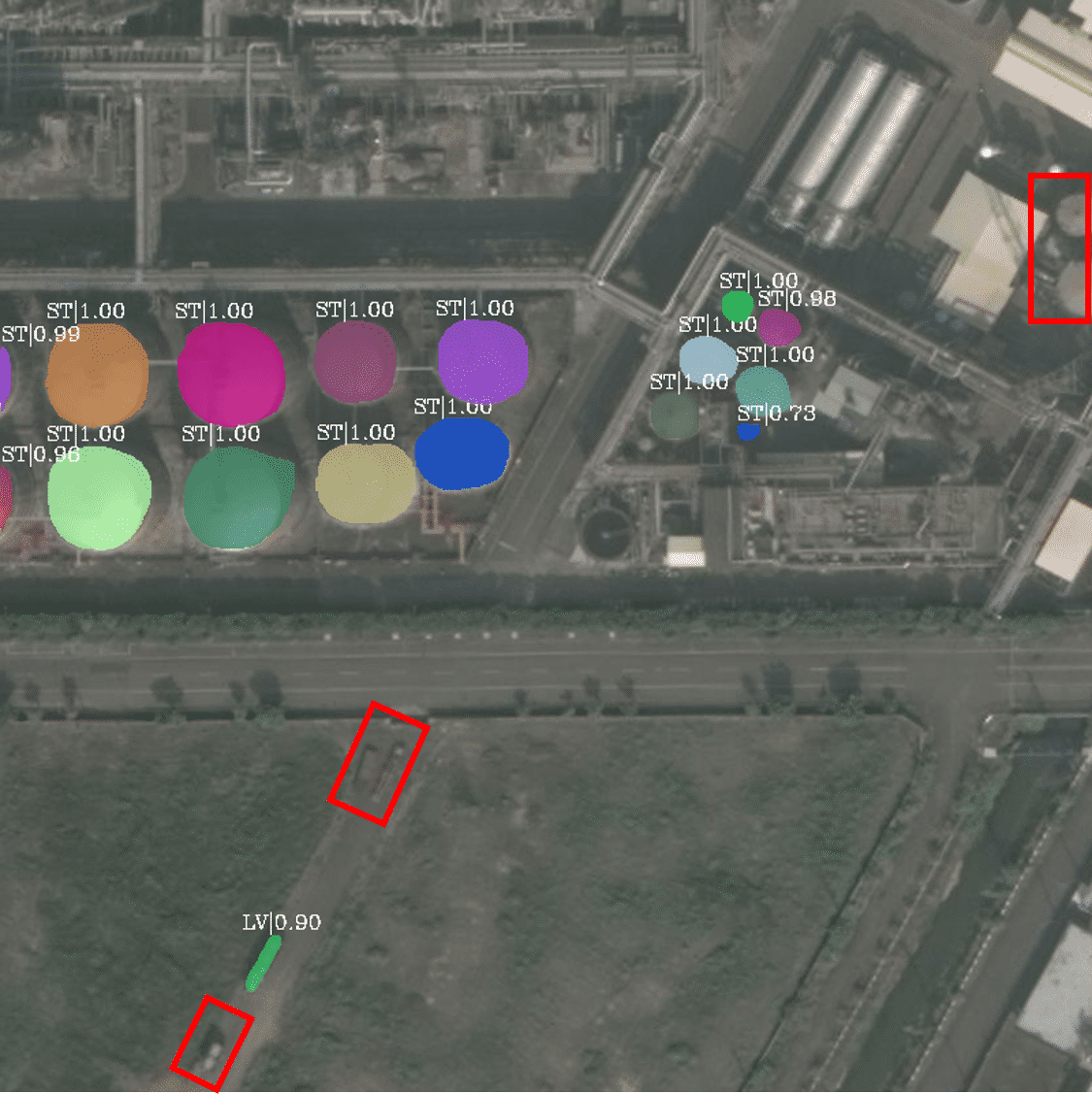}\\
			\vspace{0.2cm}
			\includegraphics[width=1.1in, height=1.1in]{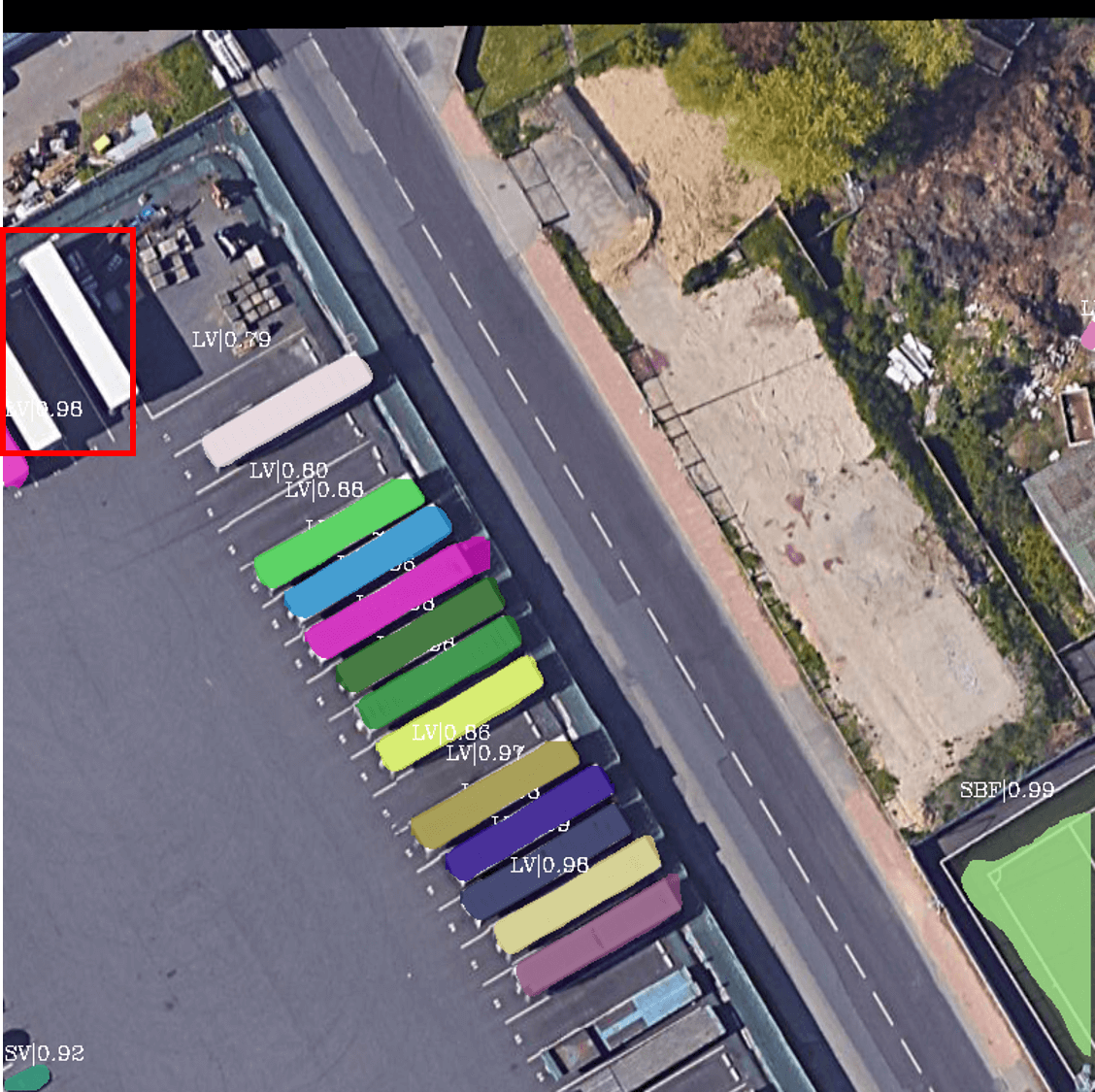}\\
		\end{minipage}%
	}%
	\centering
	\subfloat[Ground Truth]{
		\begin{minipage}[t]{0.16\linewidth}
			\centering
			\includegraphics[width=1.1in, height=1.1in ]{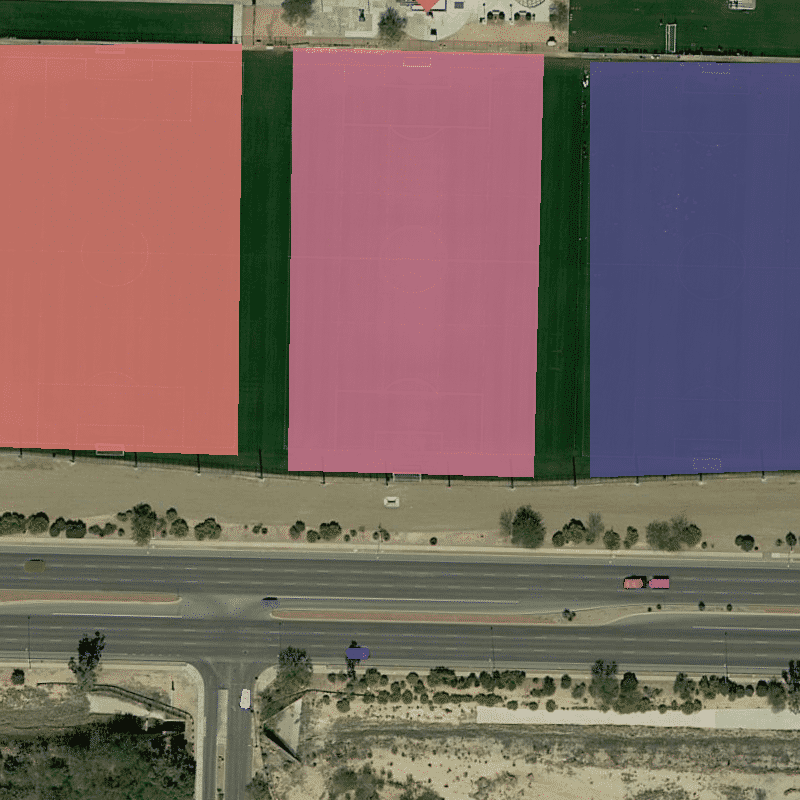}\\
			\vspace{0.2cm}
			\centering
			\includegraphics[width=1.1in, height=1.1in]{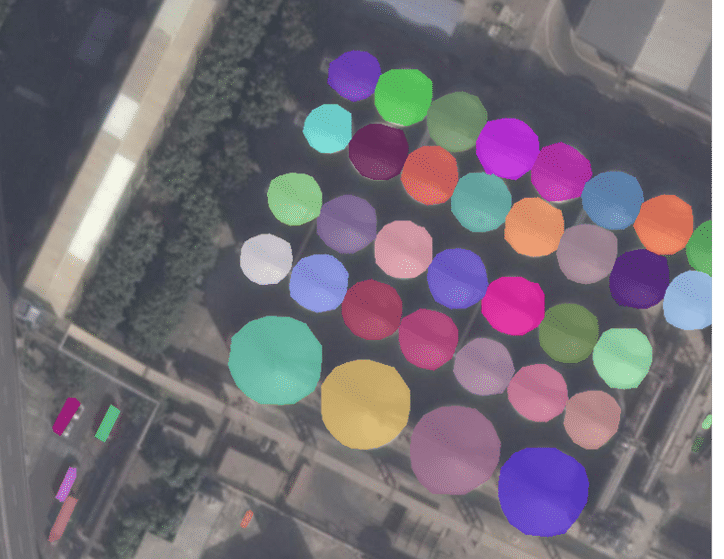}\\
		\end{minipage}%
	}%
	\subfloat[PANet]{
		\begin{minipage}[t]{0.16\linewidth}
			\centering
			\includegraphics[width=1.1in, height=1.1in]{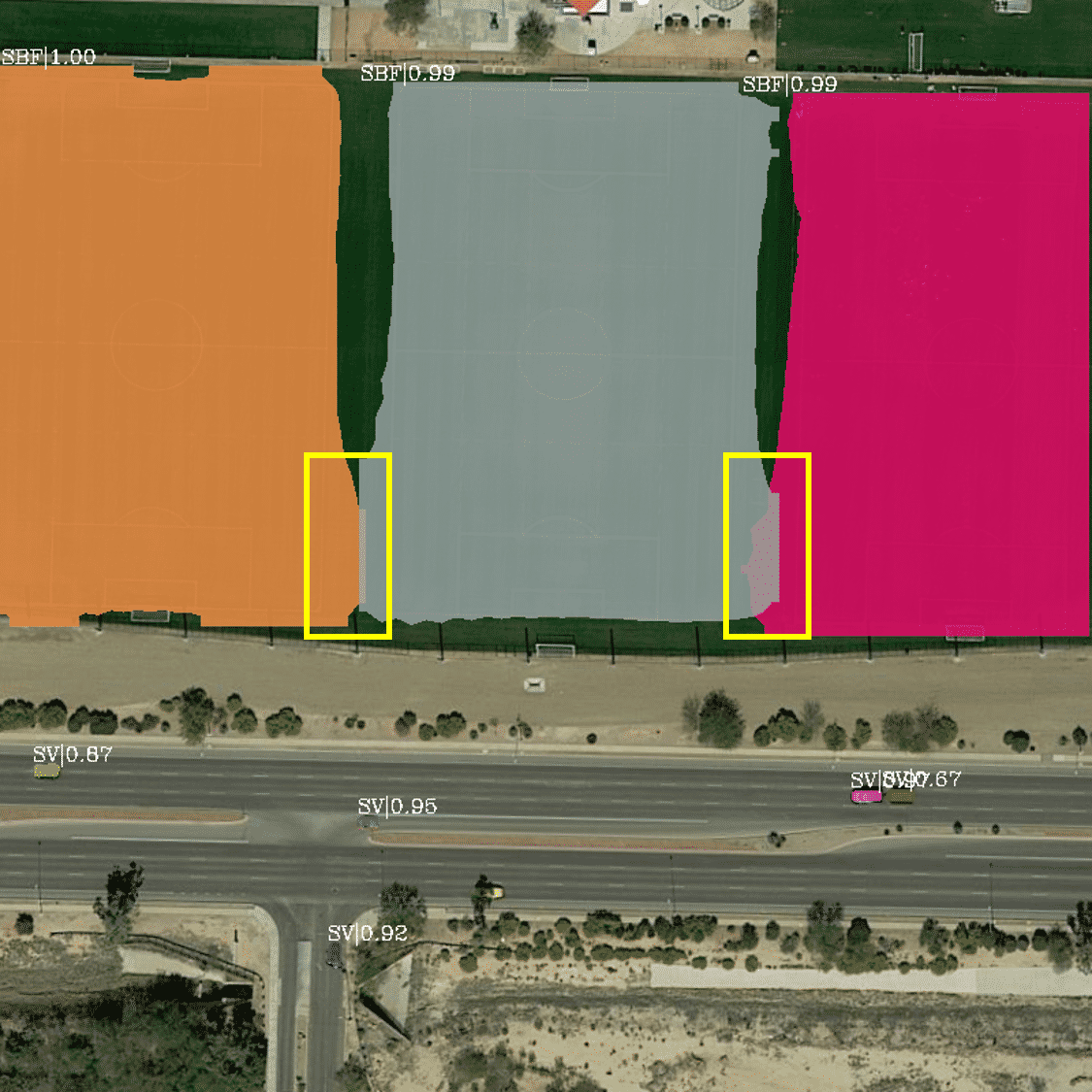}\\
			\vspace{0.2cm}
			\includegraphics[width=1.1in, height=1.1in]{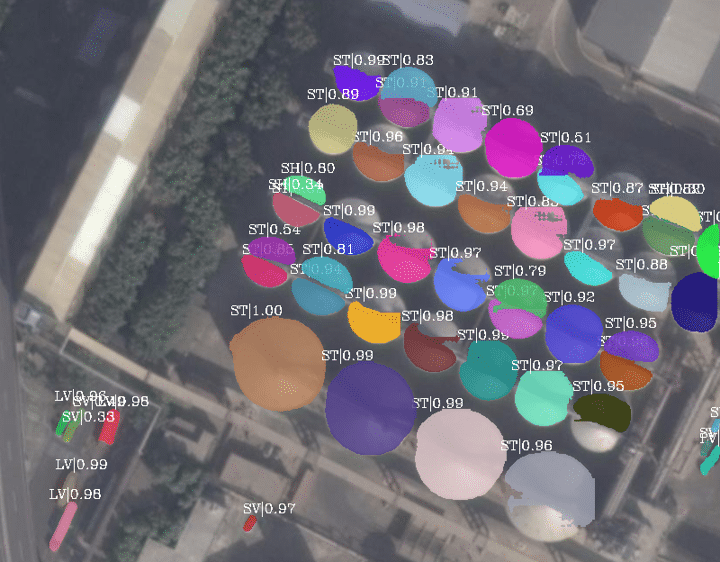}\\
		\end{minipage}%
	}%
	\subfloat[Ours]{
		\begin{minipage}[t]{0.16\linewidth}
			\centering
			\includegraphics[width=1.1in, height=1.1in]{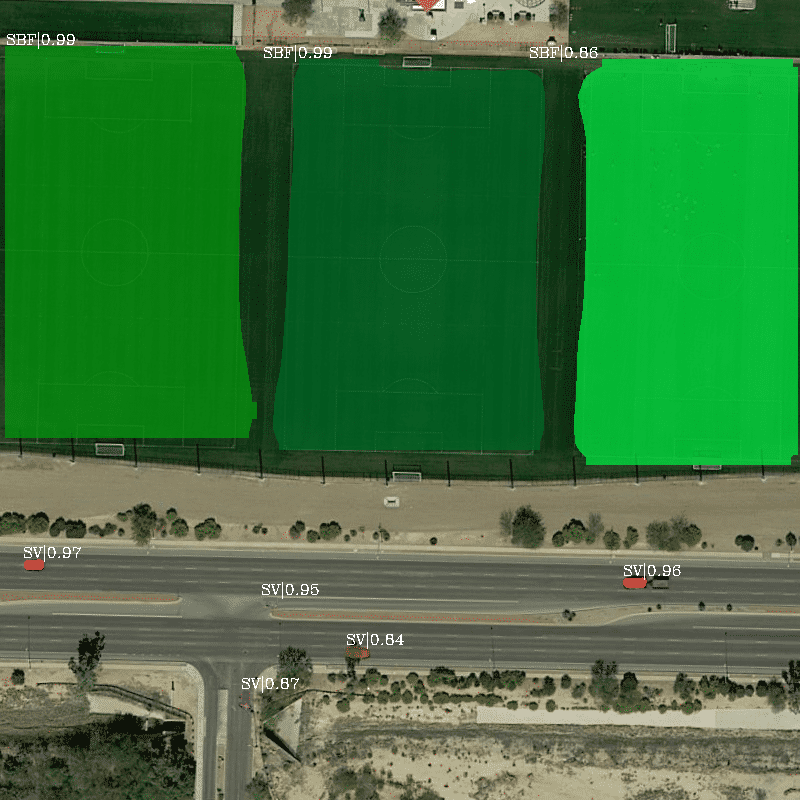}\\
			\vspace{0.2cm}
			\includegraphics[width=1.1in, height=1.1in]{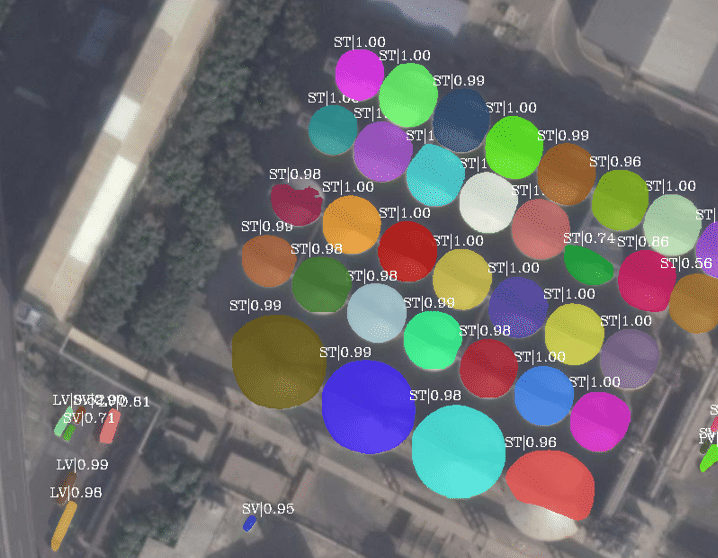}\\
		\end{minipage}%
	}%
	\caption{Comparison results on iSAID dataset. In the first row, we can see our proposed network can reduce the false prediction with complicated background interference. In the second row, we can obtain the complete segmentation results compared with the baseline. The false prediction results and the miss prediction results are indicated by yellow and red rectangles, respectively. The bounding boxes are removed for simplicity.}
	\label{fig. different iSAID}
\end{figure*}

	\subsection{Evaluation Metrics}
	We utilize the COCO evaluation metric \cite{COCO} to evaluate the network performance. 
	
	\subsubsection{COCO Evaluation} The COCO evaluation metric is based on the \textit{average precision} metric. The \textit{average precision} metric computes the average value of \textit{Precision} in the interval of \textit{Recall} from 0 to 1 under a certain IoU threshold, where the \textit{Precision} and the \textit{Recall} calculates the fraction of true positives and the fraction of positives that are correctly predicted. There are mainly 6 metrics of COCO evaluation metric for both object detection and instance segmentation:
	
	$AP$: The $AP$ measures the mean value of 10 \textit{average precision} values under the Intersection over Union (IoU) threshold from 0.5 to 0.95 with intervals of 0.05.
	
	$AP_{50}$ and $AP_{75}$: These two metrics indicate the \textit{average precision} value under the IoU threshold of 0.5 and 0.75 respectively.
	
	$AP_{S}$, $AP_{M}$ and $AP_{L}$: They correspond to the $AP$ value for small, medium and large scale instances.
	
	However, there are generally a large number of instances in RISs \cite{iSAID} and the instance scale distribution is different from natural images. We use the modified COCO evaluation metric \cite{iSAID} to evaluate the performance of our model. In the modified evaluation metric, the number of the detection boxes is set to be 1000 per image (instead of 100 by default) and the area range of large, medium, and small instances are changed, where small instances range from 10 to 144, medium instances range from 144 to 512 pixels and large instances range from 512 and over. In all the experiments, we use $AP^{m}$ and $AP^{b}$ to report the performance of segmentation and detection results.

	\subsection{Implementation Details and Parameter Optimization}
	We conduct all the experiments based on the PyTorch framework. For the network initialization, we use ImageNet pre-trained weights to initialize the backbone (i.e. ResNet-101) and the newly added layers are initialized by a zero-mean normal distribution with a standard deviation of 0.01. We choose the stochastic gradient descent with the momentum of 0.9 and weight decay of 0.0001 to fine-tune the overall network.
	
	For the training phase, we resize the input image with a short side of 800 pixels and train 12 epochs in total, where the learning rate starts from 0.01 and decreased by a factor of 0.1 at the 8th and 11th epoch. We train the network in a mini-batch size of 8 on 4 NVIDIA GeForce GTX 1080Ti with 12 GB GPU memory.  
	
	For the testing phase, the test images are resized to 800 pixels on the short side. The NMS (non-maximal suppression) threshold and the mask binarized threshold are both set as 0.5. Besides,  considering a large number of instances are present in each image, we output the top 1000 results in each image.

	\subsection{Ablation Studies}
	We conduct comprehensive experiments to evaluate the performance of the SEA module and SCMB. All ablation experiments are performed on PANet based on ResNet101 and evaluated on the iSAID validation dataset. In addition, we do not apply any data-augmentation strategies in this section.	
	\subsubsection{Evaluation of Semantic Attention Module}
	The effectiveness of the proposed SEA module can be observed from the visualization of feature maps shown in Fig. \ref{fig. show feature map}. As shown in Fig. \ref{fig. show feature map}, without the SEA module, the boundaries between different instances are blurred, and there is significant interfering noise in the background (i.e. in Fig. \ref{fig. show feature map}(b)). In contrast, Fig. \ref{fig. show feature map}(c) demonstrates clearer boundaries with less noise in the background. Similar results are shown in the first row of Fig. \ref{fig. different iSAID},  where our SS-PANet does not recognize pipes in the factory as bridges and better separates the boundaries of the soccer-ball field. Besides, we can see that with the SEA module, our network obtains performance improvement in all six metrics and increases the $AP^{m}$ by 0.5\% and $AP^{b}$ by 0.8\% compared with the baseline approaches shown in Table \ref{table-I}.
	
\begin{figure}[t]
	\centering
	\includegraphics[scale=0.35]{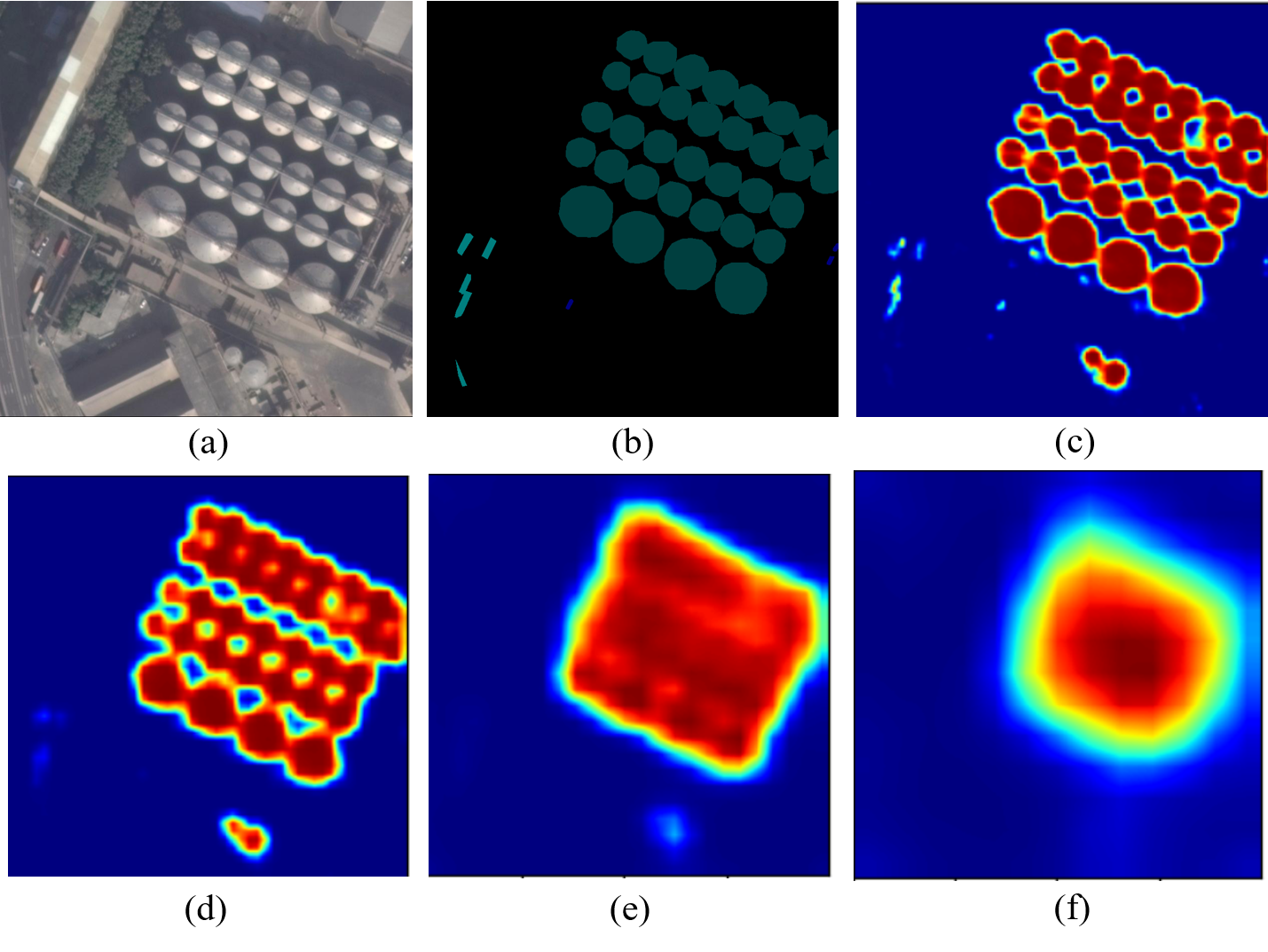}
	\caption{Semantic segmentation prediction with different uniform scale. (a) Input image. (b) Ground truth. (c) Prediction of $P_3$-$scale$. (d) Prediction of $P_4$-$scale$. (e) Prediction of $P_5$-$scale$. (f) Prediction of $P_6$-$scale$.}
	\label{fig. prediction}	
\end{figure}

	We report the other two experiments to further discuss the design of the SEA module. As described in Sec.III, we resize the 5-level output feature maps into a uniform scale and obtain the scale-normalized feature map by Eq. (\ref{Eq.1}). We first study the influence of the different uniform scales. Due to the large scale of $P_2$, once we set the uniform scale as $P_2$, the model will exceed the maximum GPU memory. Thus, we only conduct ablation studies at the scale of $P_3$, $P_4$, $P_5$, $P_6$ and term them as $P_3$-$scale$, $P_4$-$scale$, $P_5$-$scale$, $P_6$-$scale$, respectively. Table \ref{table-II} shows the results and the larger uniform scale, the better result we obtain, similar to the results in \cite{FCN}. For the segmentation performance, $P_3$-$scale$ and $P_4$-$scale$ achieve a gain of about 0.5\% compared to the baseline, and $P_5$-$scale$ has slight improvement, while $P_6$-$scale$ decreases the performance by 0.3\%. A similar trend can be observed in the detection performance. To explore the above phenomenon, we resize the semantic segmentation prediction of each setting to $ 800 \times 800 $ and present them in Fig. \ref{fig. prediction}. We can find that $P_6$-$scale$ has the coarsest segmentation prediction than the others, which will produce false semantic attention and deteriorate network performance.
	\begin{table}[t]  
	\centering
	\setlength{\abovecaptionskip}{0pt}
	\setlength{\belowcaptionskip}{10pt}
	\caption{Ablation Study on the Semantic Attention Module. All Settings are Evaluated on iSAID Validation Set}
	\label{table-II}	
	\setlength{\tabcolsep}{1.6mm}{	
		\begin{tabular}{c|ccc|ccc}  			
			\toprule   			
			Settings & $AP^{m}$ & $AP_{50}^{m}$ & $AP_{75}^{m}$ & $AP^{b}$ & $AP_{50}^{b}$ & $AP_{75}^{b}$\\  			
			\midrule   	
			\midrule 		
			PANet baseline \cite{PANet}   &38.1 &62.8 &40.5 & 43.9 & 67.0 & 48.3\\ 
			\midrule 
			\midrule  
			$P_3$-$scale$    		&\bf{38.6} &63.4 &\bf{41.0} & \bf{44.7} & \bf{68.1} & \bf{49.5}\\    	
			$P_4$-$scale$  			&38.5 &\bf{63.6} &40.8 & 44.4 & 68.1 & 49.0\\	
			$P_5$-$scale$  			&38.2 &63.4 &40.5 & 43.6 & 67.6 & 47.6\\
			$P_6$-$scale$ 			&37.8 &62.9 &40.0 & 41.9 & 66.4 & 45.6\\
			\midrule 
			\midrule 
			MULTIPLY  		&\bf{38.6} & 63.4 &\bf{41.0} & \bf{44.7} & \bf{68.1} & \bf{49.5}\\
			CONCATE  		&38.3 &63.3 &40.8 & 44.5 & 67.4 & 49.2\\
			\bottomrule  	
	\end{tabular}}	
\end{table}

	\begin{table}[!t]  
		\centering
		\setlength{\abovecaptionskip}{0pt}
		\setlength{\belowcaptionskip}{10pt}
		\caption{Ablation Study on Scale Complementary Mask Branch. All Settings are Evaluated on iSAID Validation Set}
		\label{table-III}	
		\begin{tabular}{cc|ccc|ccc}  			
			\toprule   			
			Settings & $AP^{m}$ & $AP_{50}^{m}$ & $AP_{75}^{m}$\\  			
			\midrule   	
			\midrule 		
			PANet baseline \cite{PANet}    		
			&38.6 &63.4 &41.0 \\ 
			\midrule 
			\midrule  
			$7+14$  	   		&38.9 &63.8 &41.7 \\    	
			$14+28$    			&39.3 &64.1 &42.1 \\	
			$7+14+28$    		&\bf{39.5} &\bf{64.1} &\bf{42.1} \\
			\midrule 
			\midrule 
			MULTIPLY  		&39.1 &63.8 &41.6 \\
			CONCATE  		&\bf{39.5} &\bf{64.1} &\bf{42.1} \\	
			\bottomrule  	
		\end{tabular}
	\end{table}

	We also investigate different feature fusion schemes between the semantic attention feature map and the scale-normalized feature map and design the following two feature fusion approaches. First, we employ the element-wise multiplication, represented by "MULTIPLY". Second, we concatenate the corresponding feature maps and append $1 \times 1 $ convolutional to reduce channel dimensions, named by "CONCATE". We set the uniform scale as $P_3$-$scale$ for both two schemes and report the results in Table \ref{table-II}. We can find the improvement in both two feature fusion ways and the element-wise multiplication achieves better than the channel-wise concatenation. We consider the semantic attention feature map has strong activation in the instance region, and the response of the background is almost 0. Thus, the element-wise multiplication is an intuitive way to enhance the instance activation and reduce the background noise.

\begin{table*}[t]  
	\centering
	\setlength{\abovecaptionskip}{0pt}
	\setlength{\belowcaptionskip}{10pt}
	\caption{Overall Performance Comparisons on iSAID Validation Set}
	\label{table-IV}	
	\setlength{\tabcolsep}{1.6mm}{	
		\begin{tabular}{c|cccccc|cccccc}  			
			\toprule   			
			Settings & $AP^{m}$ & $AP_{50}^{m}$ & $AP_{75}^{m}$ & $AP_{s}^{m}$ & $AP_{m}^{m}$ & $AP_{l}^{m}$ & $AP^{b}$ & $AP_{50}^{b}$ & $AP_{75}^{b}$ & $AP_{s}^{b}$ & $AP_{m}^{b}$ & $AP_{l}^{b}$\\  			
			\midrule   			
			Mask-RCNN \cite{MaskRCNN} 	  & 37.4 & 62.0 & 39.4 & 39.7 & 51.5 & 35.7 & 43.1 & 66.2 & 47.4 & 45.8 & 55.9 &51.2\\  
			
			PANet \cite{PANet}	  & 38.1 & 62.8 & 40.5 & 40.5 & 51.9 & \bf{36.7} & 43.9 & 67.0 & 48.3 & 46.5 & 56.7 & \bf{62.3}\\
			\midrule  
			
			SS-Mask-RCNN  & 39.2 & 63.7 & 41.8 & 41.8 & \bf{54.4} & 24.3 & 43.8 & 67.7 & 48.4 & 46.6 & 57.2 & 28.2\\ 
			
			SS-PANet 	  & 39.5 & 64.1 & 42.1 & 41.7 & 53.5 & 35.0 & 44.6 & 68.5 & 48.4 & 47.3 & 58.4  & 55.8\\ 
			
			\midrule  
			SS-Mask-RCNN+ & 39.7 & 64.4 & 42.2 & 42.5 & 53.6 & 26.3  &  45.0 & 68.6 & 49.7 & 47.9 & 56.6  & 37.1\\ 
			
			SS-PANet+ 	  & \bf{40.8} & \bf{65.6} & \bf{43.8} & \bf{43.7} & 54.0 & 32.1 & \bf{46.9} & \bf{70.0} & \bf{52.0} & \bf{49.8} & \bf{57.2} & 44.7\\ 
			
			\bottomrule  	
	\end{tabular}}	
\end{table*}

\begin{table*}[t]  
	\centering
	\setlength{\abovecaptionskip}{0pt}
	\setlength{\belowcaptionskip}{10pt}
	\caption{Class-wise Instance Segmentation Results on iSAID Validation Set}
	\label{table-V}		
	\setlength{\tabcolsep}{2mm}
	\begin{tabular}{c|c c c c c c c c c c c c c c c c}  			
		\toprule   			
		Model		& $AP$ & PL & BD & BR & GTF & SV & LV & SH & TC & BC & ST & SBF & RA & HA & SP & HC \\			
		\midrule
		Mask-RCNN \cite{MaskRCNN} 	& 37.4 & 48.4 & 55.8 & 22.9 & 31.8 & 14.0 & 38.5 & 50.2 & 76.6 & 42.2 & 34.8 & 46.1 & 37.5 & 26.7 & 30.3 & 5.1\\  
		PANet \cite{PANet}  		& 38.1 & 49.2 & 55.9 & 22.8 & 32.4 & 14.1 & 40.6 & 50.4 & 77.9 & 45.5 & 35.2 & 47.1 & 38.7 & 26.9 & 30.5 & 4.9\\ 
		\midrule
		SS-Mask-RCNN & 39.2 & 50.8 & 58.0 & 23.9 & 33.1 & 14.6 & 41.6 & 52.1 & 78.7 & 44.2 & 37.4 & 48.0 & 39.9 & 28.8 & 31.0 & 6.0\\
		SS-PANet	 & 39.5 & 50.9 & 58.8 & 23.5 & 34.4 & 14.7 & 41.8 & 52.0 & 78.8 & 46.8 & 37.2 & 46.8 & 40.4 & 28.3 & 31.2 & \bf{6.9}\\
		\midrule
		SS-Mask-RCNN+ & 39.7 & 51.5 & 59.1 & 24.5 & 34.0 & 15.6 & 42.6 & 52.9 & 79.2 & 45.3 & 37.7 & 46.0 & 40.8 & 28.7	 & 33.0 & 4.4\\
		SS-PANet+	 & \bf{40.8} & \bf{53.1} & \bf{60.3} & \bf{24.8} & \bf{35.6} & \bf{16.1} & \bf{43.7} & \bf{54.0} & \bf{79.8} & \bf{49.3} & \bf{38.7} & \bf{48.0} & \bf{40.8} & \bf{29.9} & \bf{33.4} & 5.3\\
		\bottomrule  	
	\end{tabular}	
\end{table*}
\begin{table*}[t]  
	\centering
	\setlength{\abovecaptionskip}{0pt}
	\setlength{\belowcaptionskip}{10pt}
	\caption{Class-wise Object Detection Results on iSAID Validation Set}
	\label{table-VI}		
	\setlength{\tabcolsep}{2mm}
	\begin{tabular}{c|c c c c c c c c c c c c c c c c}  			
		\toprule   			
		Model		& $AP$ & PL & BD & BR & GTF & SV & LV & SH & TC & BC & ST & SBF & RA & HA & SP & HC \\			
		\midrule
		Mask-RCNN \cite{MaskRCNN} 	& 43.1 & 67.2 & 55.7 & 27.2 & 45.6 & 16.6 & 44.8 & 54.9 & 77.2 & 42.5 & 35.5 & 44.5 & 37.7 & 48.5 & 33.8 & 15.2\\  
		PANet \cite{PANet}		& 43.9 & 67.7 & 56.4 & 26.8 & 47.0 & 16.5 & 45.3 & 54.7 & 78.9 & 44.6 & 35.7 & 45.9 & 39.1 & 49.3 & 34.2 & 16.6\\ 
		\midrule
		SS-Mask-RCNN & 43.8 & 66.6 & 57.4 & 27.5 & 46.2 & 16.9 & 45.8 & 55.4 & 79.2 & 42.4 & 37.4 & 44.9 & 40.5 & 49.2 & 33.5 & 15.0\\
		SS-PANet	 & 44.6 & 68.0 & 58.0 & 28.2 & 49.0 & 16.4 & 46.2 & 55.2 & 78.7 & 43.3 & 36.0 & 46.2 & 40.2 & 49.7 & 34.3 & 19.9\\
		\midrule
		SS-Mask-RCNN+ & 45.0 & 68.6 & 58.4 & 27.8 & 47.1 & 18.3 & 46.9 & 57.0 & 79.7 & 44.3 & 37.4 & 44.5 & 41.6 & 50.5 & 36.5 & 16.2\\
		SS-PANet+ 	& \bf{46.9} & \bf{70.8} & \bf{60.1} & \bf{29.6} & \bf{50.3} & \bf{18.4} &\bf{48.6} & \bf{58.0} & \bf{81.0} & \bf{48.0} & \bf{39.1} &\bf{46.6} & \bf{42.1} & \bf{52.9} & \bf{36.6} & \bf{20.8}\\
		\bottomrule  	
	\end{tabular}	
\end{table*}

\begin{table*}[!t]  
	\centering
	\setlength{\abovecaptionskip}{0pt}
	\setlength{\belowcaptionskip}{10pt}
	\caption{Overall Performance Comparisons on iSAID Test Set}
	\label{table-VII}	
	
	\setlength{\tabcolsep}{1.6mm}{	
		\begin{tabular}{c|cccccc|cccccc}  			
			\toprule   			
			Settings & $AP^{m}$ & $AP_{50}^{m}$ & $AP_{75}^{m}$ & $AP_{s}^{m}$ & $AP_{m}^{m}$ & $AP_{l}^{m}$ & $AP^{b}$ & $AP_{50}^{b}$ & $AP_{75}^{b}$ & $AP_{s}^{b}$ & $AP_{m}^{b}$ & $AP_{l}^{b}$\\  	  			
			\midrule

			Mask-RCNN+ \cite{iSAID} 	  & 33.4 & 56.8 & 34.7 & 35.8 & 46.5 & 23.9 & 37.2 & 60.8 & 40.7 & 39.8 & 43.7 &16.0\\
			
			D2Det \cite{D2Det} & 37.5 & 61.0 & 39.8 & - & - & - & - & - & - & - & - &-\\ 
			
			SS-PANet & 39.3 & 62.5 & 42.5 & 42.4 & 47.8 & 13.8 & 44.5 & 66.2 & 50.1 & 47.8 & 50.2 &16.5\\      
			
			HTC \cite{HTC}& 39.4 & 62.5 & 42.5 & 42.3 & 49.0 & 14.8 & \textbf{46.6} & 66.5 & 52.2 & 49.6 & \textbf{55.7} & 17.4 \\ 
			Cascade-Mask-RCNN \cite{CascadeR-CNN}& 39.4 & 62.5 & 42.5 & 42.3 & 49.0 & 14.8 &\textbf{46.6} & 66.5 & 52.2 & 49.6 & \textbf{55.7} & 17.4 \\ 		
			PANet+ \cite{iSAID}	  & 39.5 & 63.6 & 42.2 & 42.1 & \bf{53.6} & \bf{38.5} & 46.3 & 66.9 & 51.7 & 48.9 & 53.3 & \bf{26.5}\\
			
			SS-PANet+ 	  & \bf{40.6} & \bf{64.1} & \bf{44.0} & \bf{44.0} & 49.8 & 13.8 & \bf{46.6} & \bf{67.7} & \bf{52.4} & \bf{50.0} & 54.0 &17.3\\  
			\bottomrule  	
	\end{tabular}}	
\end{table*} 

	\subsubsection{Evaluation of Scale Complementary Mask Branch}
	By introducing the SCMB, the network increase segmentation performance from 38.6 to 39.5 and remains comparable detection performance as shown in the fourth row of Table \ref{table-I}. We also visualize the comparison segmentation result in the second row of Fig. \ref{fig. different iSAID}. With the SCMB, the network avoids separating the storage tank into two parts and achieves complete segmentation results for large vehicles, compared with the single-scale mask branch\cite{PANet}.
	
	Furthermore, we conduct ablation studies on the setting of SCMB. Based on the original mask branch \cite{MaskRCNN} with only a spatial resolution of $14 \times 14$, we extend it to the following three multi-scale forms. The first two forms introduce only a parallel branch with a spatial resolution of $7 \times 7$ and a spatial resolution of $28 \times 28$, respectively. For the last one, it simultaneously includes the above two parallel branches. The corresponding scale complementary supervision is applied in all three settings. Table \ref{table-III} gives the corresponding results where '$7 + 14$', '$14 + 28$' and '$7 + 14 + 28$' represent the above three settings. We can find that all these settings improve the segmentation performance and the third setting leads to the best performance.
	
	We also consider two fusion operations for the feature fusion module. As shown in Table \ref{table-III}, the channel-wise concatenation achieves a better result. It is noticed that this result is exactly the opposite of the result shown in Table \ref{table-II}. For channel-wise concatenation, it can better fuse information from different scales of feature maps. However, for element-wise multiplication, the weak results from a certain feature map may affect the final fusion results.
	
	It can be seen from the ablation study that the element-wise multiplication has a better result in the SEA module, while channel-wise concatenating performs well in the SCMB. Thus, we follow the above settings in all subsequent experiments.

	\subsubsection{Speed and Complexity Comparison}
	For the inference speed, we report the comparison results in Table \ref{table-I}. First, we extend the baseline with our proposed SEA, and the FPS is reduced by 0.6. This reduction is mainly due to the additional convolutional layer of the proposed SEA module. Then, we extend the baseline with SCMB. The FPS
	drops by 0.8, and we consider that the SCMB changes the original single mask branch to the trident form, which affects the inference speed. Compared with the SEA module, the SCMB needs to operate on each RoI feature, thus it has a worse impact on speed.
	Finally, when using both the proposed SEA module and SCMB, the FPS is reduced from 5.1 to 3.6. Besides, we also report the comparison results of complexity in the last two columns in Table \ref{table-I}.

	\begin{figure*}[htbp]
		\centering
		\subfloat[PL]{
			\begin{minipage}[t]{0.24\linewidth}
				\centering
				\includegraphics[width=1.7in, height=1.7in]{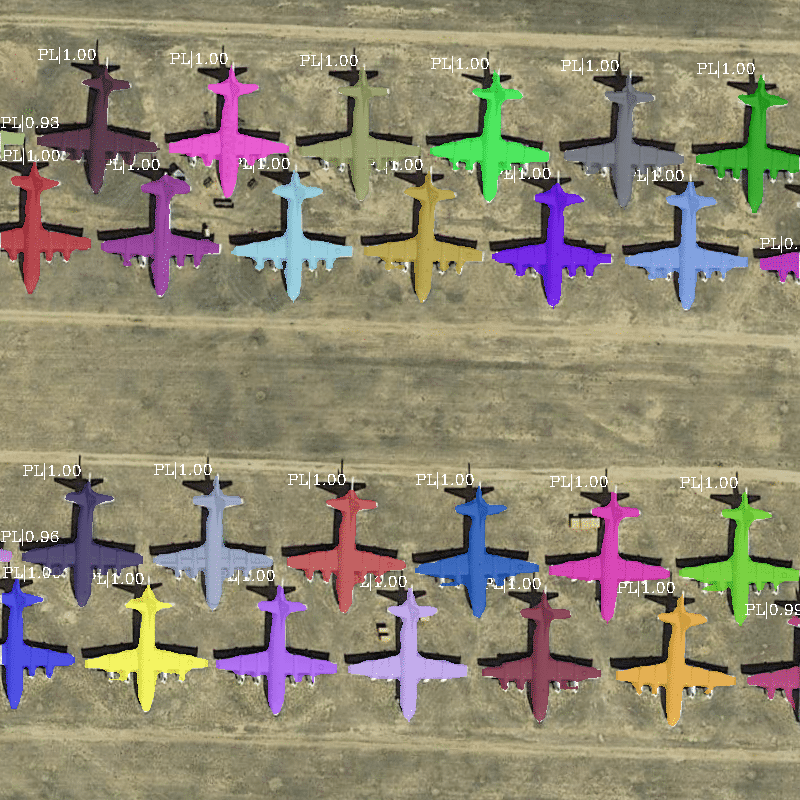}
			\end{minipage}%
		}%
		\subfloat[BD]{
			\begin{minipage}[t]{0.24\linewidth}
				\centering
				\includegraphics[width=1.7in, height=1.7in]{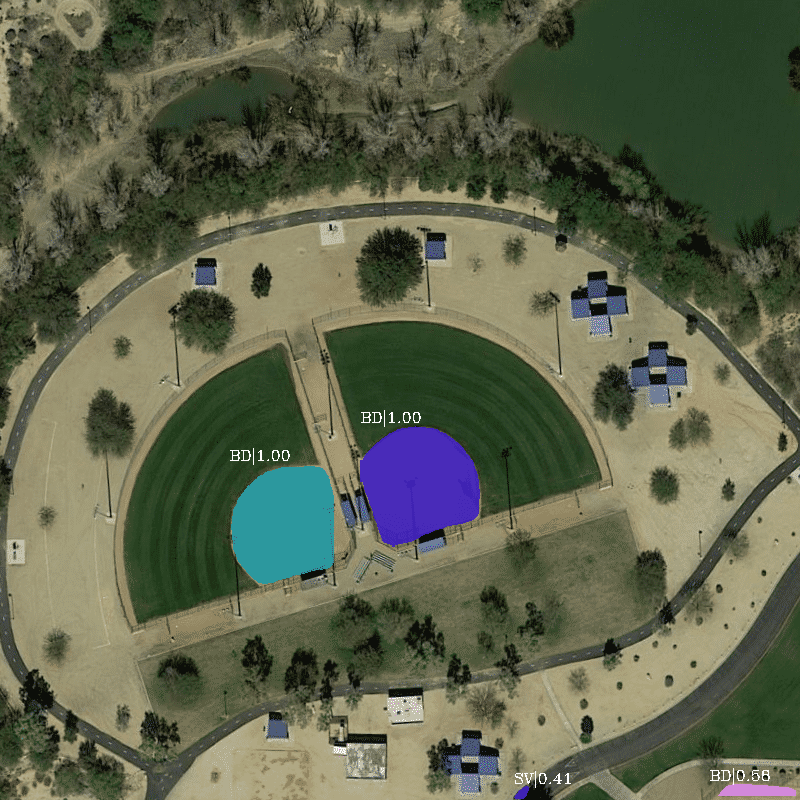}
			\end{minipage}%
		}%
		\subfloat[SP and SV]{
			\begin{minipage}[t]{0.24\linewidth}
				\centering
				\includegraphics[width=1.7in, height=1.7in]{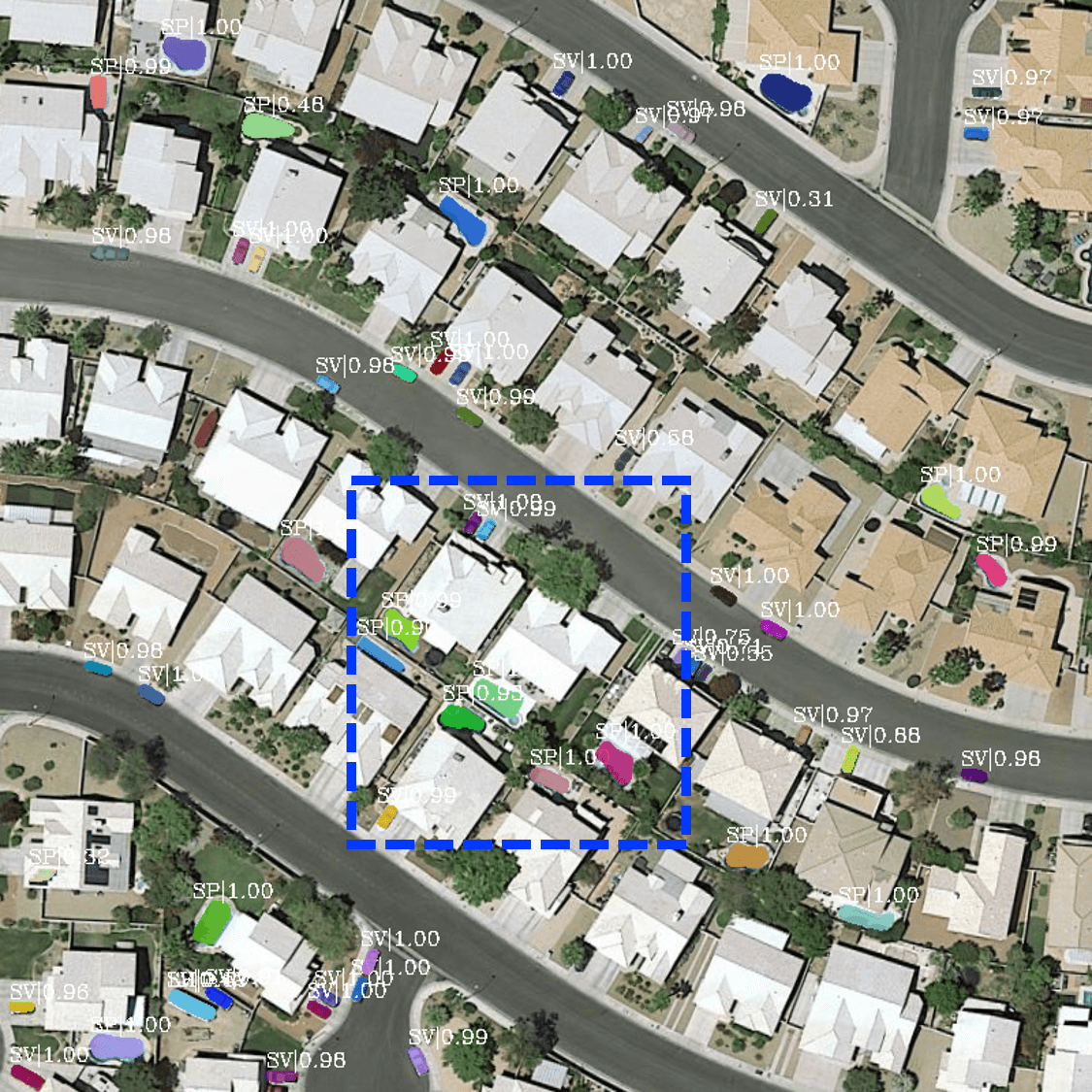}
			\end{minipage}%
		}%
		\subfloat[]{
			\begin{minipage}[t]{0.24\linewidth}
				\centering
				\includegraphics[width=1.7in, height=1.7in]{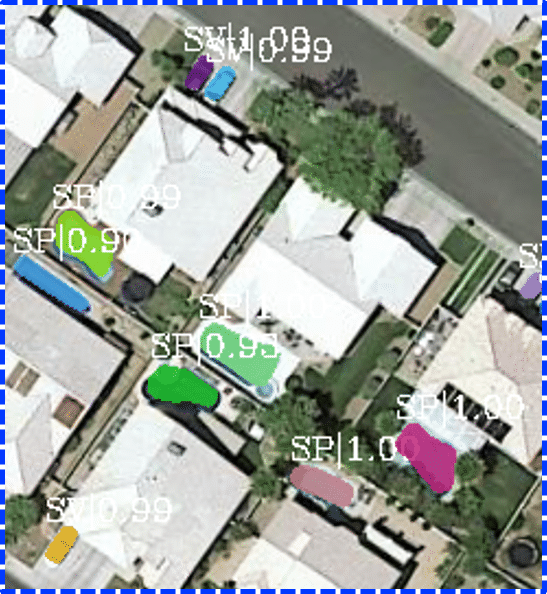}
			\end{minipage}%
		}%
		\hfill
		\centering
		\subfloat[TC and BC]{
			\begin{minipage}[t]{0.24\linewidth}
				\centering
				\includegraphics[width=1.7in, height=1.7in]{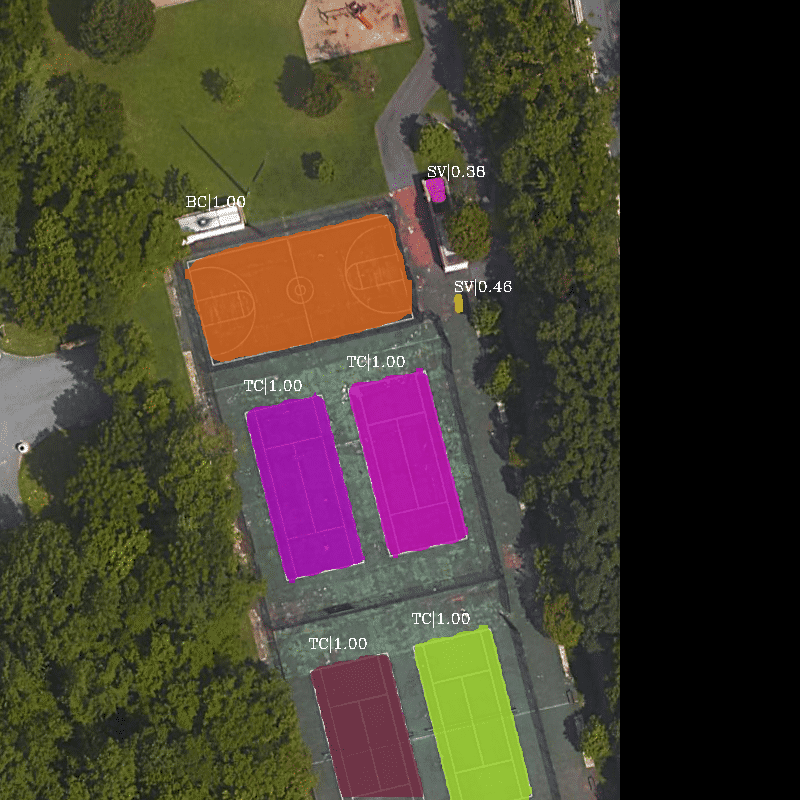}
			\end{minipage}%
		}%
		\subfloat[BR]{
			\begin{minipage}[t]{0.24\linewidth}
				\centering
				\includegraphics[width=1.7in, height=1.7in]{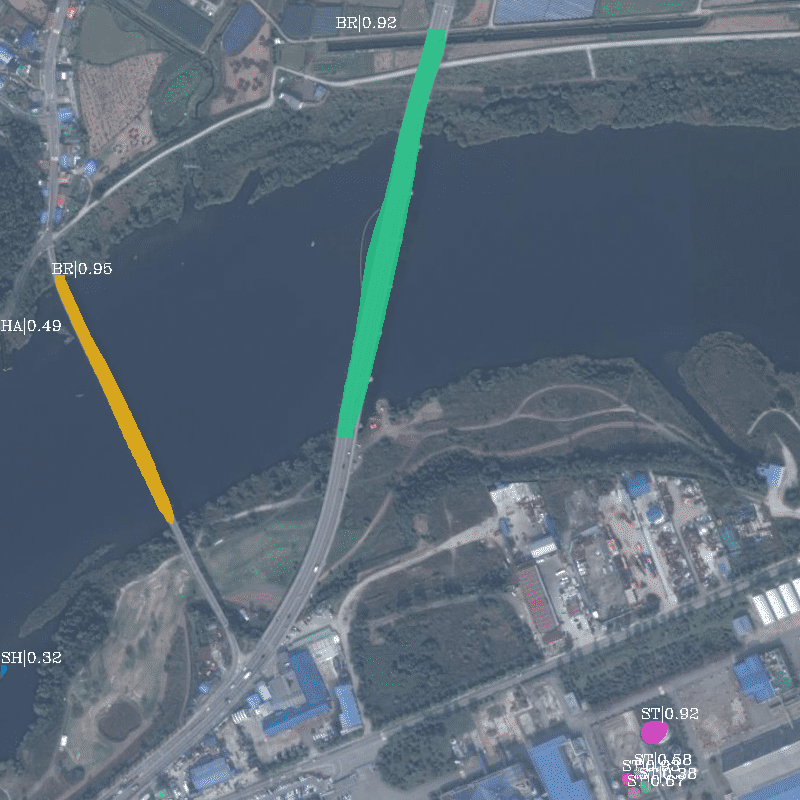}
			\end{minipage}%
		}%
		\subfloat[HA and SH]{
			\begin{minipage}[t]{0.24\linewidth}
				\centering
				\includegraphics[width=1.7in, height=1.7in]{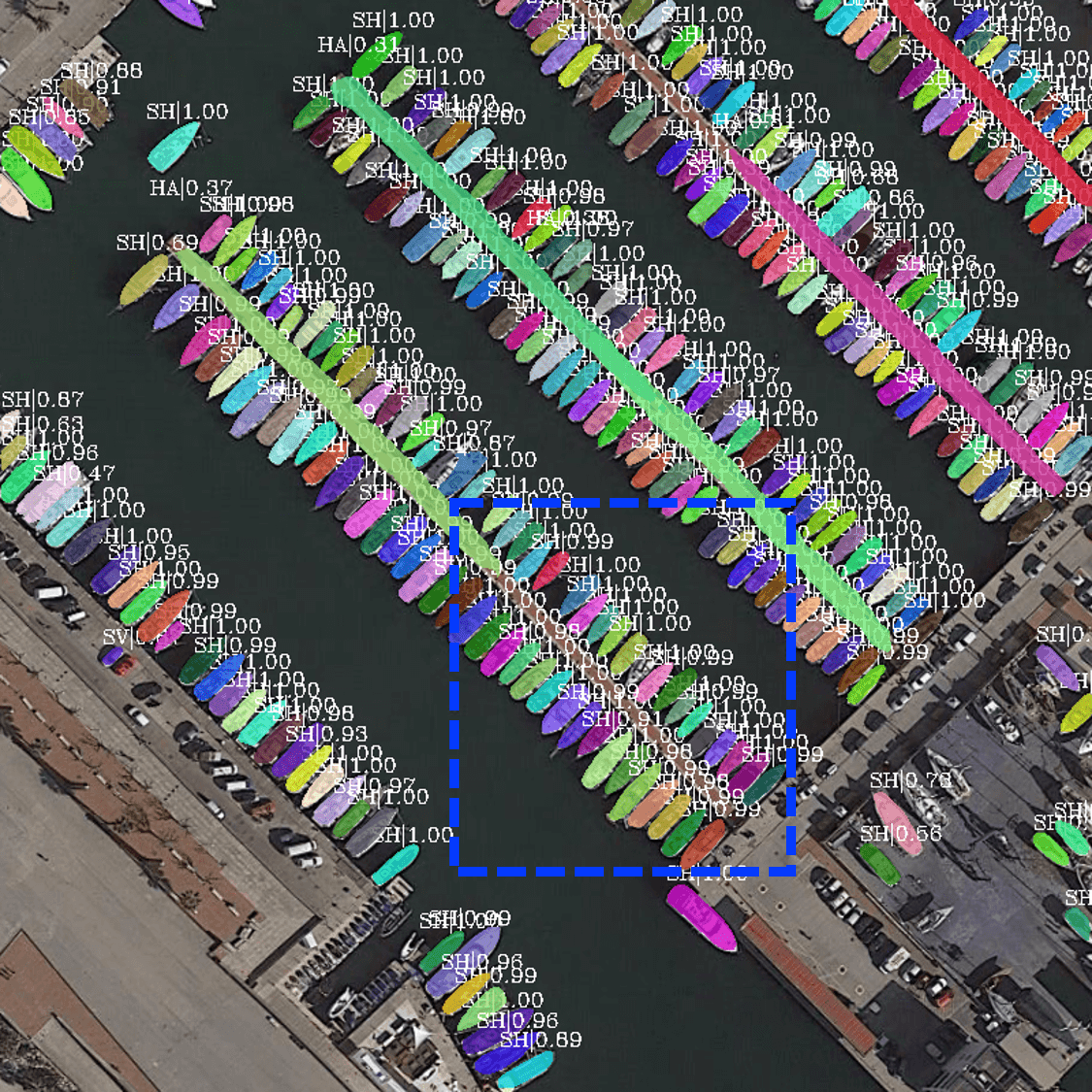}
			\end{minipage}%
		}%
		\subfloat[]{
			\begin{minipage}[t]{0.24\linewidth}
				\centering
				\includegraphics[width=1.7in, height=1.7in]{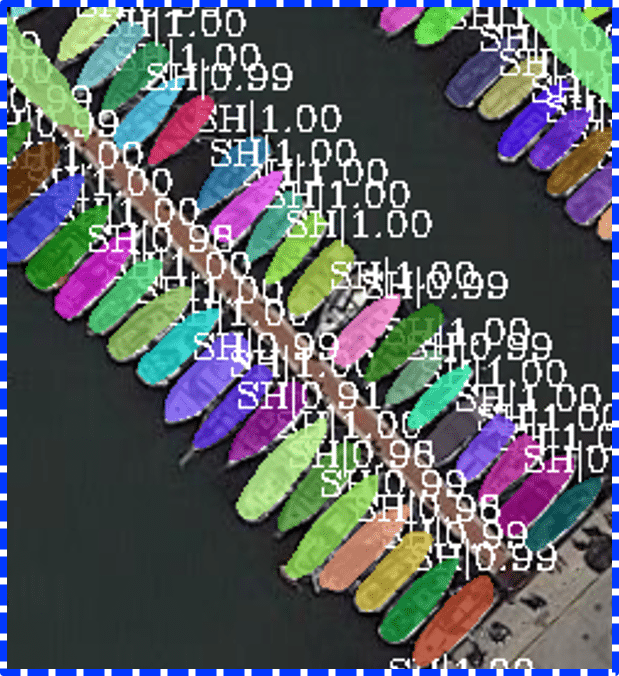}
			\end{minipage}%
		}%
		\hfill
		\centering
		\subfloat[GTF and SBF]{
			\begin{minipage}[t]{0.24\linewidth}
				\centering
				\includegraphics[width=1.7in, height=1.7in]{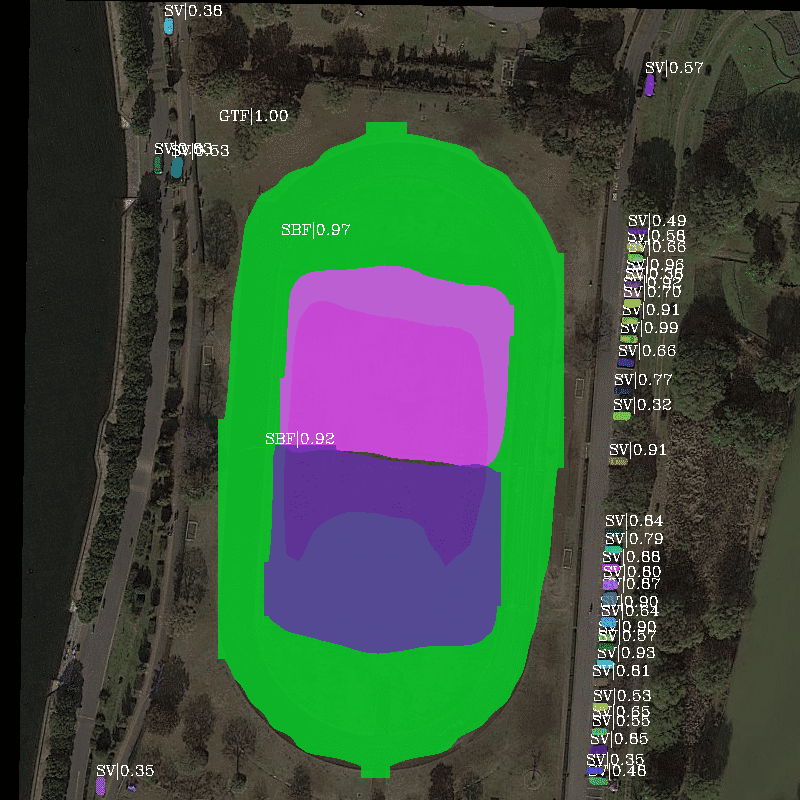}
			\end{minipage}%
		}%
		\subfloat[RA and SV]{
			\begin{minipage}[t]{0.24\linewidth}
				\centering
				\includegraphics[width=1.7in, height=1.7in]{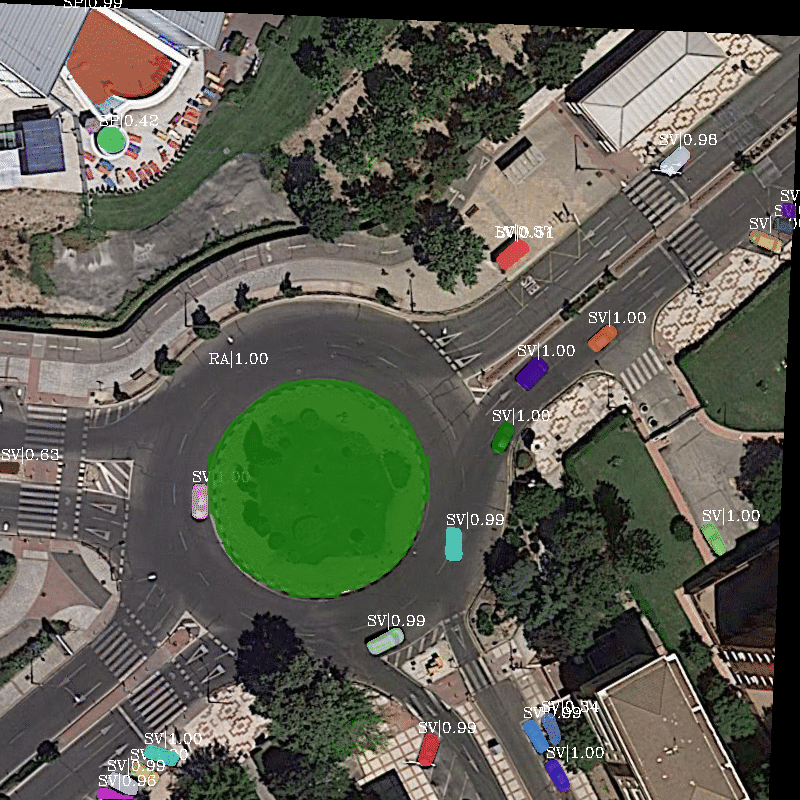}
			\end{minipage}%
		}%
		\subfloat[SV]{
			\begin{minipage}[t]{0.24\linewidth}
				\centering
				\includegraphics[width=1.7in, height=1.7in]{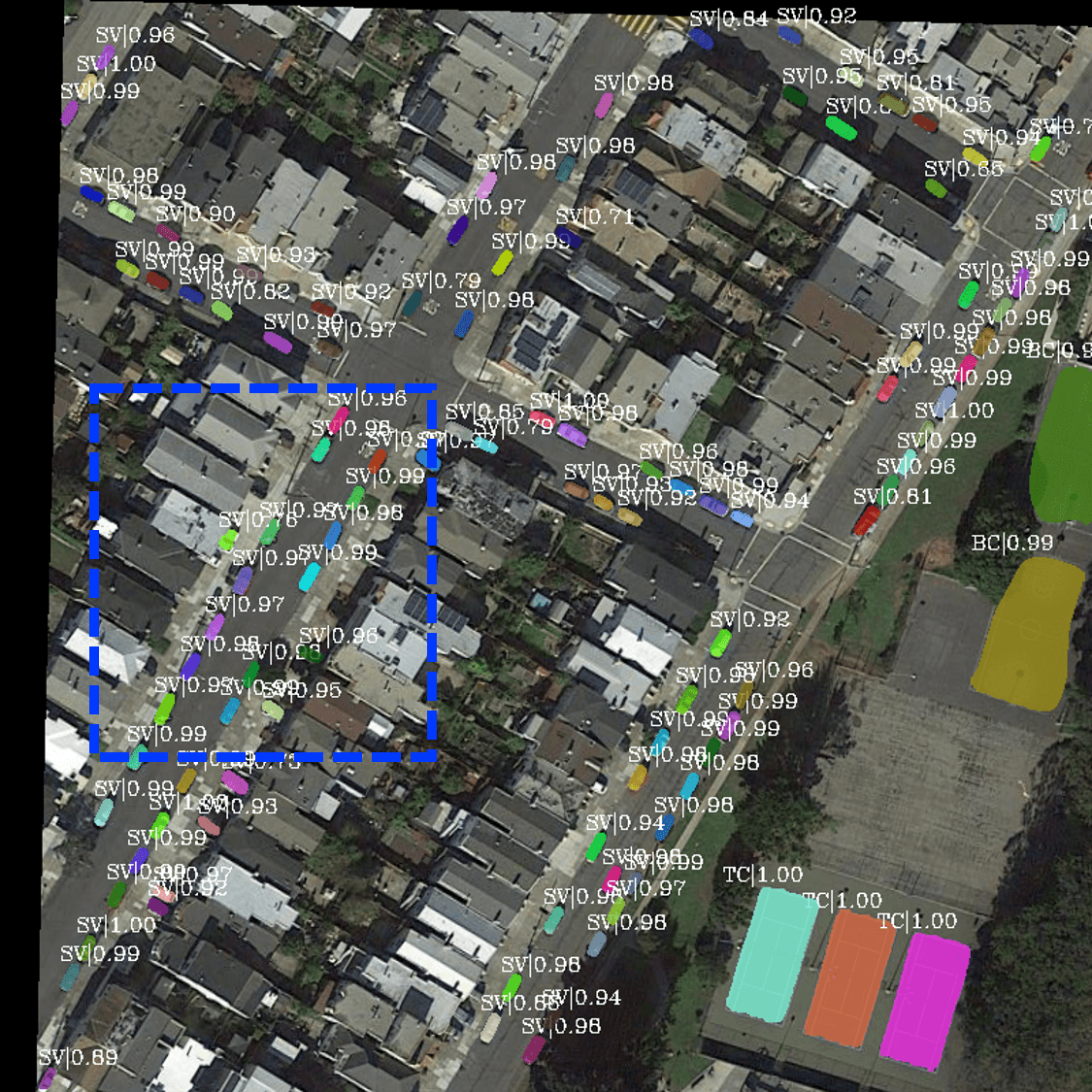}
			\end{minipage}%
		}%
		\subfloat[]{
			\begin{minipage}[t]{0.24\linewidth}
				\centering
				\includegraphics[width=1.7in, height=1.7in]{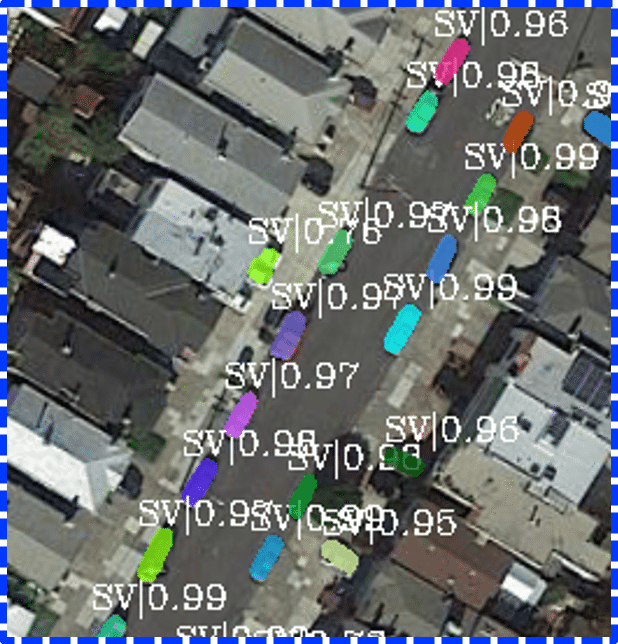}
			\end{minipage}%
		}%
		\hfill
		\centering
		\subfloat[HC]{
			\begin{minipage}[t]{0.24\linewidth}
				\centering
				\includegraphics[width=1.7in, height=1.7in]{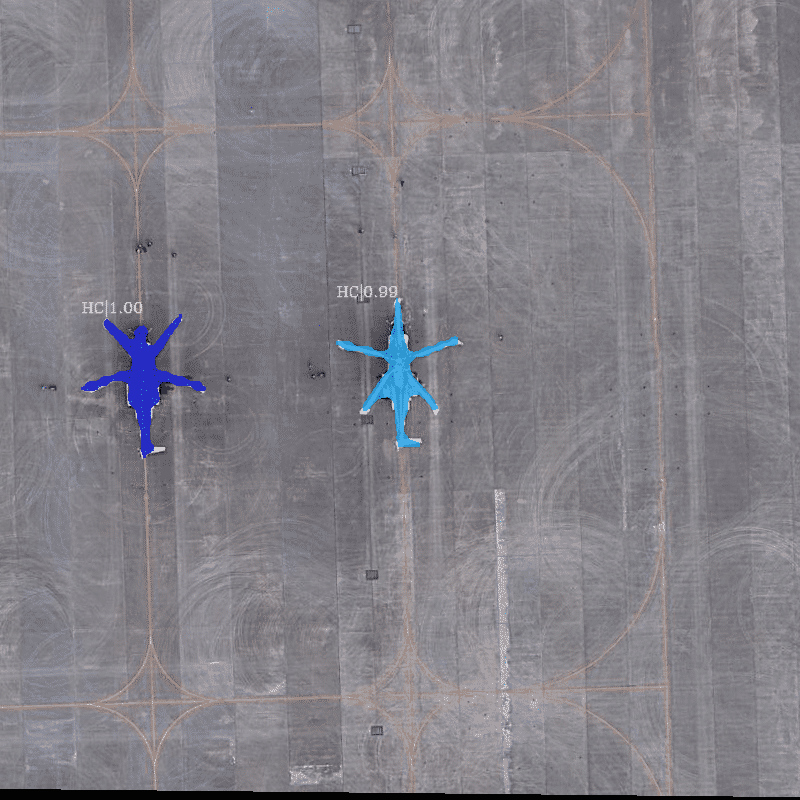}
			\end{minipage}%
		}%
		\subfloat[ST]{
			\begin{minipage}[t]{0.24\linewidth}
				\centering
				\includegraphics[width=1.7in, height=1.7in]{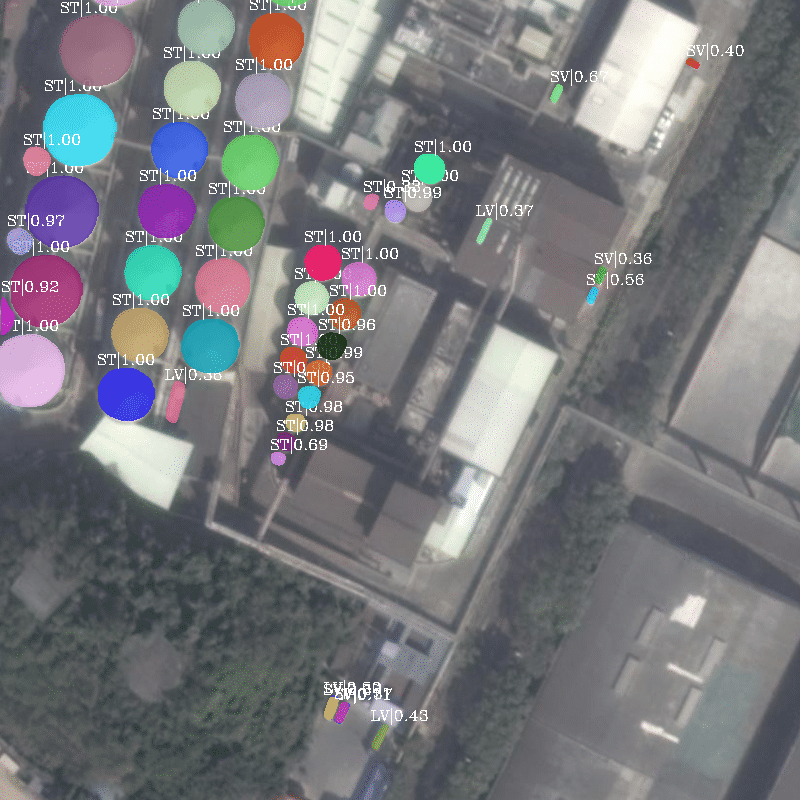}
			\end{minipage}%
		}%
		\subfloat[LV]{
			\begin{minipage}[t]{0.24\linewidth}
				\centering
				\includegraphics[width=1.7in, height=1.7in]{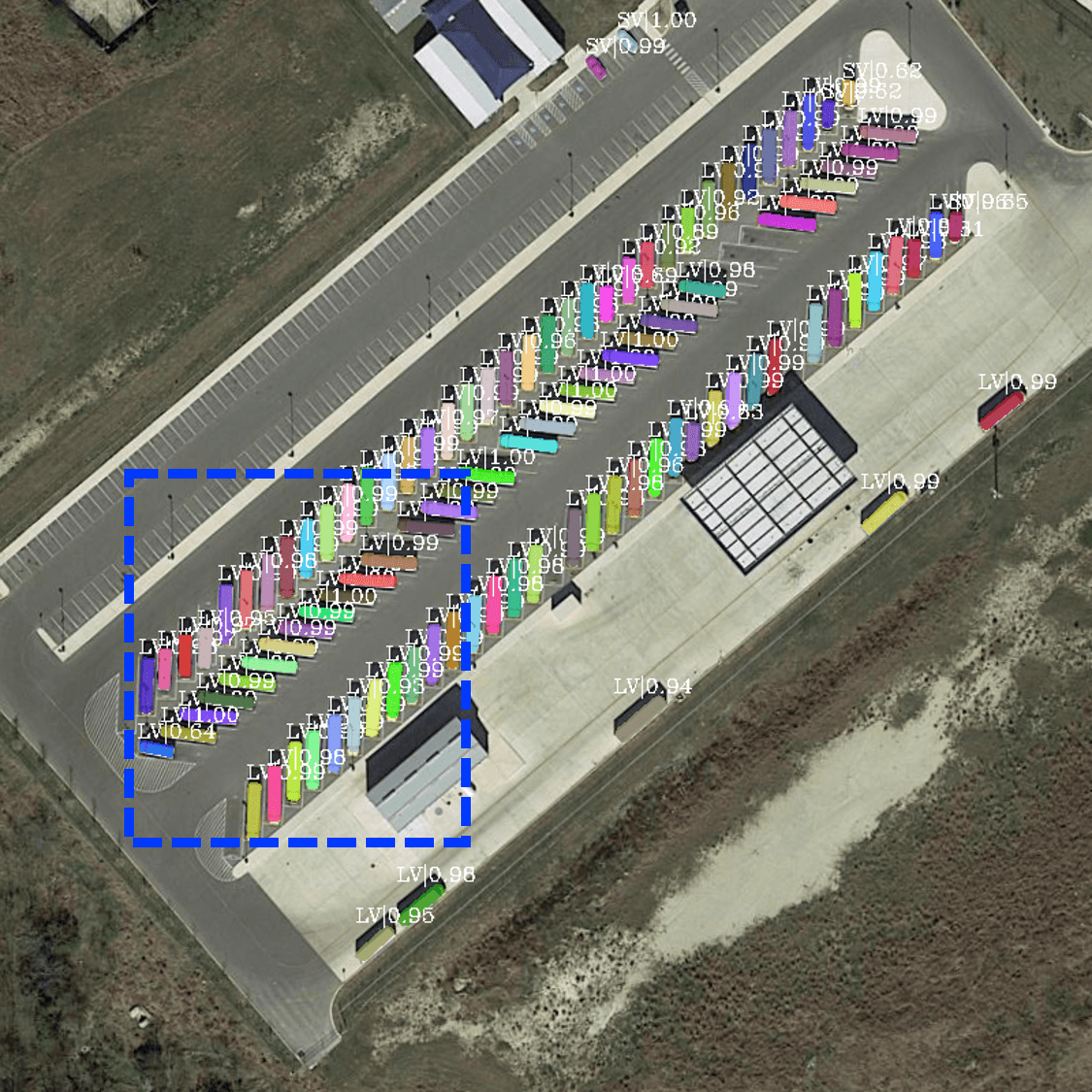}
			\end{minipage}%
		}%
		\subfloat[]{
			\begin{minipage}[t]{0.24\linewidth}
				\centering
				\includegraphics[width=1.7in, height=1.7in]{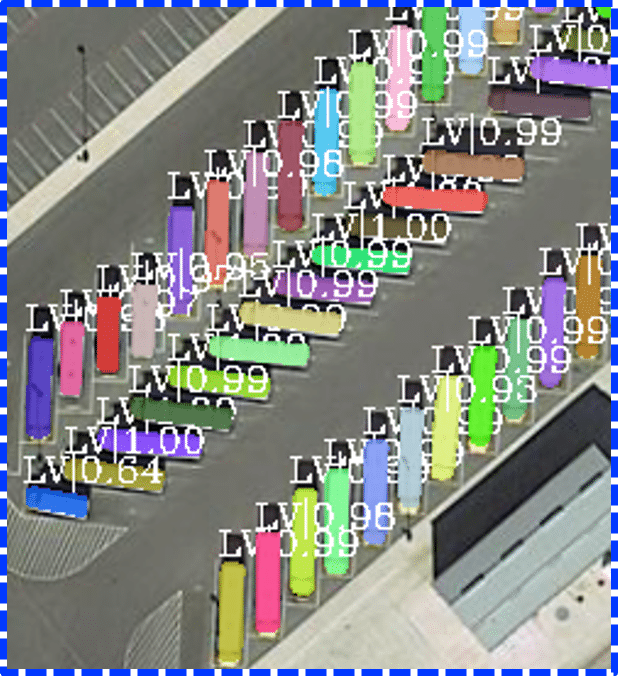}
			\end{minipage}%
		}%
		
		\caption{Performance of proposed network on iSAID dataset. The first to the third column shows the results of each category, and we zoom in the densely arranged results in the last column. The bounding boxes are removed for simplicity.}
		\label{fig. result iSAID}
	\end{figure*}

	\begin{table*}[!t]  
	\centering
	\setlength{\abovecaptionskip}{0pt}
	\setlength{\belowcaptionskip}{10pt}
	\caption{Overall Performance Comparisons on NWPU VHR-10 Instance Segmentation Test Set}
	\label{table-VIII}		
	\setlength{\tabcolsep}{2mm}
	\begin{tabular}{c|cccccc|cccccc}  			
		\toprule   			
		Model & $AP^{m}$ & $AP_{50}^{m}$ & $AP_{75}^{m}$ & $AP_{s}^{m}$ & $AP_{m}^{m}$ & $AP_{l}^{m}$ & $AP^{b}$ & $AP_{50}^{b}$ & $AP_{75}^{b}$ & $AP_{s}^{b}$ & $AP_{m}^{b}$ & $AP_{l}^{b}$\\  			
		\midrule   			
		Mask-RCNN \cite{MaskRCNN}	  & 60.9 & 90.2 & 67.2 & 61.3 & 55.3 & - & 60.6 & 90.3 & 69.9 & 61.1 & 47.3  
		&-\\   
		PANet \cite{PANet}	  & 62.3 & 91.4 & 68.1 & 62.5 & 56.1 & - & 61.7 & 91.1 & 71.2 & 62.0 & 48.6 &-\\ 
		\midrule  
		SS-Mask-RCNN  & 62.3 & 92.0 & 69.4 & 62.2 & 64.7 & - & 63.6 & 92.6 & 72.2 & 63.9 & 56.3	 
		&-\\  
		SS-PANet 	  & 63.6 & 92.4 & 70.2 & 63.4 & 65.3 & - & 64.5 & 92.9 & 74.3 & 64.7 & 58.1 
		&-\\ 
		\midrule 
		SS-Mask-RCNN+ & 65.1 & 93.8 & 74.0 & 65.0 & 65.3 & - & 65.2 & 93.6 & 76.0 & 65.3 & 57.2 &-\\  
		SS-PANet+ 	  & \bf{66.1} & \bf{94.5} & \bf{74.6} & \bf{65.9} & \bf{66.4} & - & \bf{65.9} & \bf{94.2} & \bf{76.7} & \bf{66.0} & \bf{58.8} &-\\  
		\bottomrule  	
	\end{tabular}	
\end{table*}

\begin{table*}[!h]  
	\centering
	\setlength{\abovecaptionskip}{0pt}
	\setlength{\belowcaptionskip}{10pt}
	\caption{Class-wise Instance Segmentation Results on NWPU VHR-10 Instance Segmentation Test Set}
	\label{table-IX}		
	\setlength{\tabcolsep}{1.8mm}
	\begin{tabular}{c|c c c c c c c c c c c}  			
		\toprule   			
		Model		& $AP$ & Airplane & Ship & Storagek & Baseball  & Tennis & Basketball & Ground track  & Harbor & Bridge & Vehicle \\		
		&  	  & 		 &  	& tank 	   & diamond   & court  &court 		 &field 
		& 		& 		 & \\	
		\midrule
		Mask-RCNN \cite{MaskRCNN}	 	& 60.9 & 37.7 & 51.6 & 79.3 & 84.2 & 65.0 & 68.4 & 85.0 & 53.4 & 30.4 & 53.6\\
		PANet \cite{PANet}			& 62.3 & 39.0 & 53.8 & 80.1 & 84.9 & 66.3 & 70.5 & 85.3 & 55.5 & 32.9 & 54.9\\
		\midrule
		SS-Mask-RCNN	& 62.3 & 38.3 & 52.5 & 80.1 & 84.9 & 65.7 & 69.3 & 85.7 & 55.6 & 36.4 & 54.4\\
		SS-PANet 		& 63.6 & 39.5 & 54.6 & 80.9 & 85.4 & 67.7 & 71.2 & 86.2 & 57.0 & 37.1 & 55.9\\
		\midrule
		SS-Mask-RCNN+   & 65.1 & 41.7 & 54.0 & \bf{81.3} & 86.3 & 71.4 & 71.6 & \bf{88.3} & 58.9 & 40.3 & 57.3\\
		
		SS-PANet+	    & \bf{66.1} & \bf{42.8} & \bf{55.2} & 81.2 & \bf{86.6} & \bf{72.8} & \bf{72.3} & 88.0 & \bf{59.5} & \bf{41.6} & \bf{57.9}\\
		
		\bottomrule  	
	\end{tabular}	
\end{table*}

\begin{table*}[!h]  
	\centering
	\setlength{\abovecaptionskip}{0pt}
	\setlength{\belowcaptionskip}{10pt}
	\caption{Class-wise Object Detection Results on NWPU VHR-10 Instance Segmentation Test Set}
	\label{table-X}	
	\setlength{\tabcolsep}{1.8mm}
	\begin{tabular}{c|c c c c c c c c c c c}  			
		\toprule   			
		Model		& $AP$ & Airplane & Ship & Storagek & Baseball  & Tennis & Basketball & Ground track  & Harbor & Bridge & Vehicle \\		
		&  	  & 		 &  	& tank 	   & diamond   & court  &court 		 &field 
		& 		& 		 & \\	
		\midrule
		Mask-RCNN \cite{MaskRCNN}	 & 60.6 & 70.9 & 56.0 & 76.1 & 79.4 & 65.6 & 63.8 & 71.3 & 41.5 & 25.8 & 55.9\\
		
		PANet \cite{PANet}	 		& 61.7 & 71.3 & 57.8 & 76.3 & 79.6 & 66.7 & 65.4 & 72.8 & 42.8 & 27.7 & 56.6\\
		
		\midrule
		SS-Mask-RCNN	& 63.6 & 71.5 & 61.9 & 76.7 & 81.1 & 66.4 & 66.1 & 75.6 & 45.6 & 34.3 & 57.0\\
		
		SS-PANet 		& 64.5 & 72.5 & 62.2 & 77.0 & 81.6 & 67.0 & 67.1 & 76.2 & 47.5 & 35.9 & 57.8\\
		
		\midrule
		SS-Mask-RCNN+ 	& 65.2 & 72.2 & 63.4 & 77.9 & 81.9 & 67.9 & 67.5 & \bf{76.7} & 50.3 & 36.6 & 57.9\\
		
		SS-PANet+ 		& \bf{65.7} & \bf{73.3}& \bf{63.7}  & \bf{78.1} & \bf{82.2} &\bf{68.4} & \bf{68.2} & 76.6 & \bf{51.6} & \bf{37.0} & \bf{58.5}\\
		
		\bottomrule  	
	\end{tabular}	
\end{table*}

	\subsection{Results on iSAID}
	To quantitatively evaluate the proposed method, we integrate the SEA module and SCMB into two representative networks (Mask-RCNN, PANet) and name them as SS-Mask-RCNN and SS-PANet. We report the overall comparison performance on iSAID validation set in Table \ref{table-IV} to show the performance of our proposed method. As shown in these two tables, our SS-Mask-RCNN/SS-PANet performs better than the baseline Mask-RCNN/PANet by 1.8\%/1.4\% in  $AP^{m}$ and 0.7\%/0.7\% in $AP^{b}$. This not only indicates the superiority of our proposed method but also shows that our SEA module and SCMB are robust to different baselines. Considering the large scale-variation of instances in RSIs, multi-scale training is a common and effective strategy to improve performance \cite{RSIObjectDetection2} \cite{ICN}. We randomly resize the short side of the input image with (1000,800,600,400) in the training phase and name this model as SS-Mask-RCNN+/SS-PANet+. As shown in Table \ref{table-IV}, SS-PANet+ achieves the best performance as 40.8\%/46.9\%. In addition, it can be seen that the comparison results of $AP^{m}_{l}$ and $AP^{b}_{l}$ in Table \ref{table-IV} are unstable. We calculate the distribution of instances' areas in the image patches of the iSAID validation set and find there are only 9 instances belong to the large scale, which is less than 1\% (9 vs 238,138) of the number of instances in the whole validation set. Therefore, a tiny deviation in prediction may lead to a large difference in performance.
\begin{figure*}[!t]
	\subfloat[Ground Truth]{
		\begin{minipage}[t]{0.16\linewidth}
			\centering
			\includegraphics[width=1.1in, height=1.1in ]{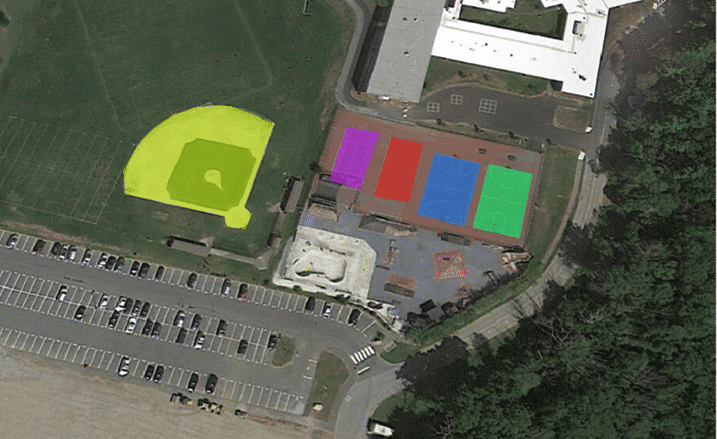}\\
			\vspace{0.2cm}
			\centering
			\includegraphics[width=1.1in, height=1.1in]{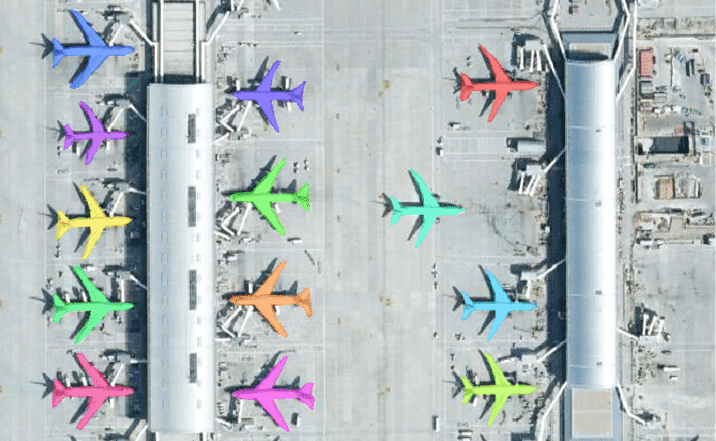}\\
		\end{minipage}%
	}%
	\subfloat[PANet]{
		\begin{minipage}[t]{0.16\linewidth}
			\centering
			\includegraphics[width=1.1in, height=1.1in]{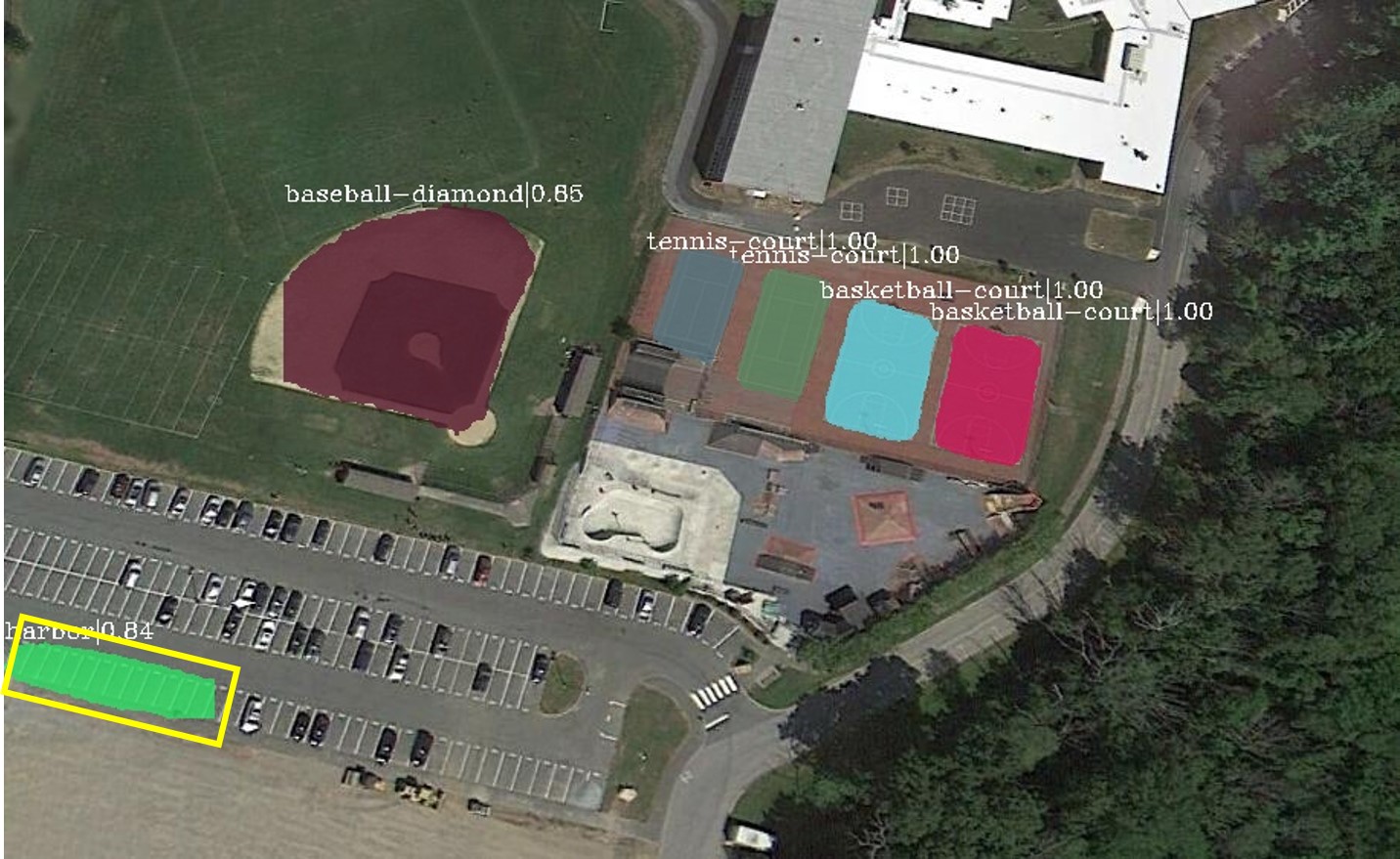}\\
			\vspace{0.2cm}
			\includegraphics[width=1.1in, height=1.1in]{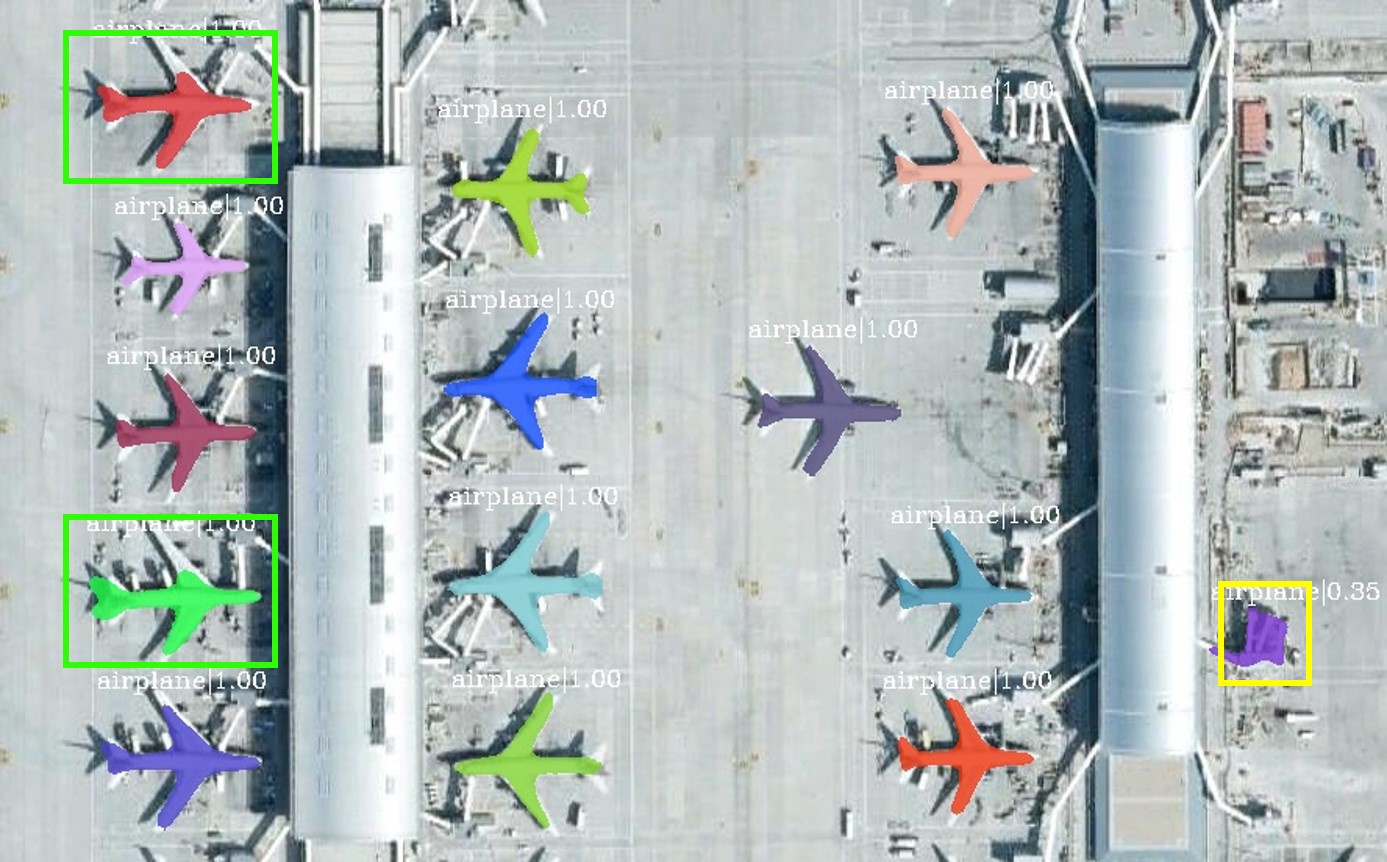}\\
		\end{minipage}%
	}%
	\subfloat[Ours]{
		\begin{minipage}[t]{0.16\linewidth}
			\centering
			\includegraphics[width=1.1in, height=1.1in]{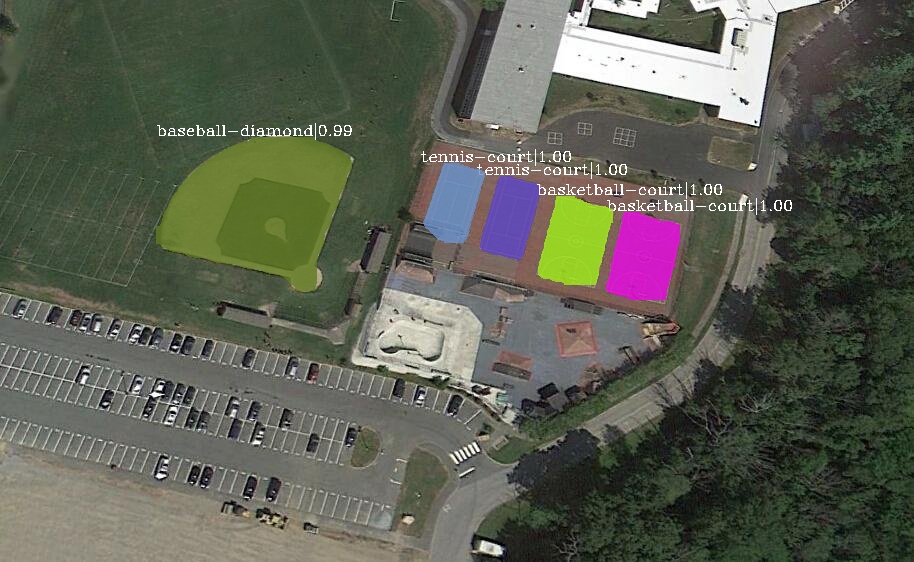}\\
			\vspace{0.2cm}
			\includegraphics[width=1.1in, height=1.1in]{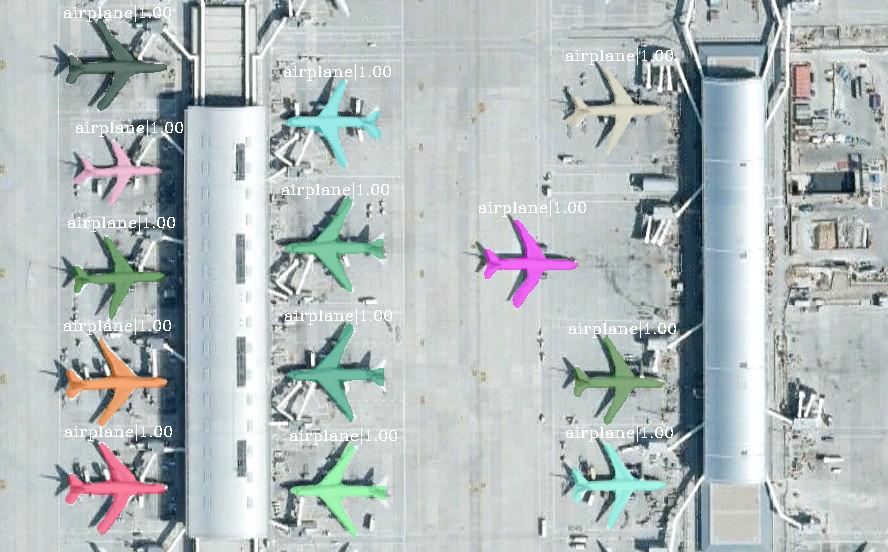}\\
		\end{minipage}%
	}%
	\centering
	\subfloat[Ground Truth]{
		\begin{minipage}[t]{0.16\linewidth}
			\centering
			\includegraphics[width=1.1in, height=1.1in ]{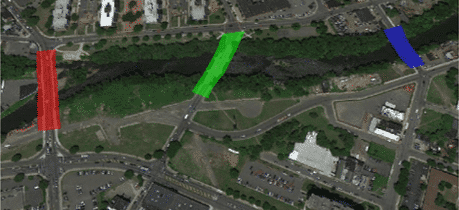}\\
			\vspace{0.2cm}
			\centering
			\includegraphics[width=1.1in, height=1.1in]{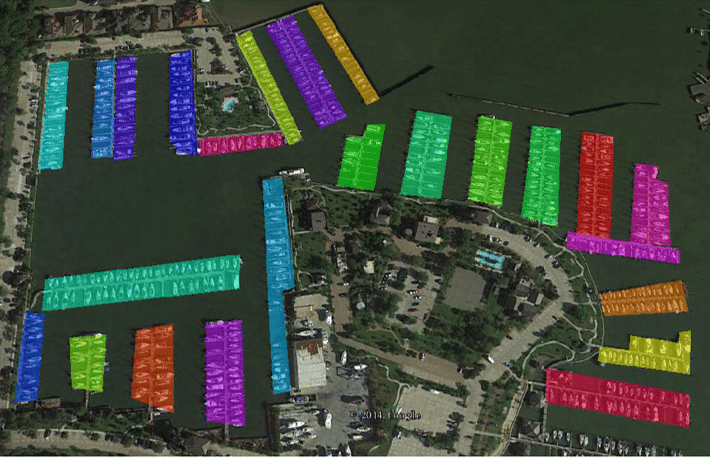}\\
		\end{minipage}%
	}%
	\subfloat[PANet]{
		\begin{minipage}[t]{0.16\linewidth}
			\centering
			\includegraphics[width=1.1in, height=1.1in]{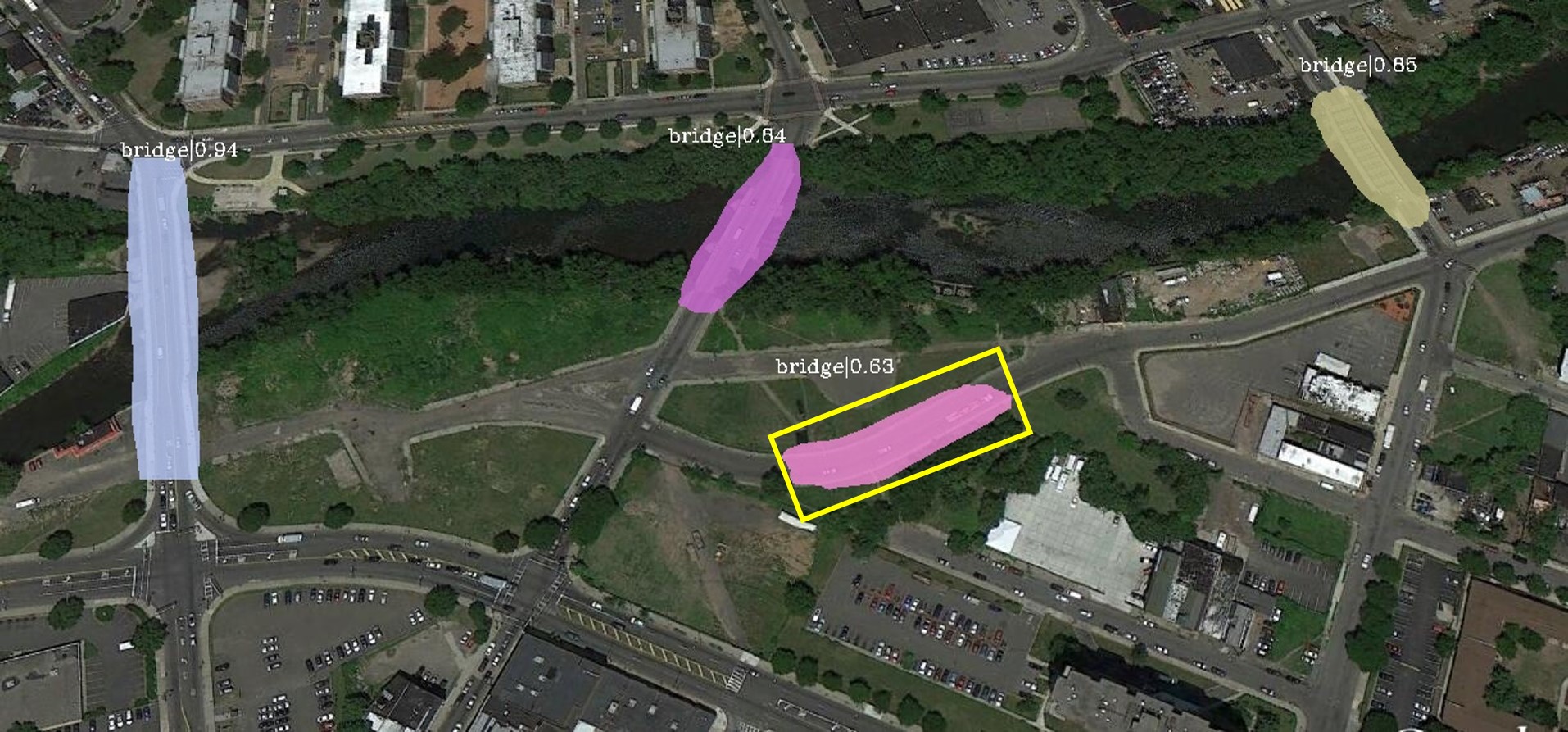}\\
			\vspace{0.2cm}
			\includegraphics[width=1.1in, height=1.1in]{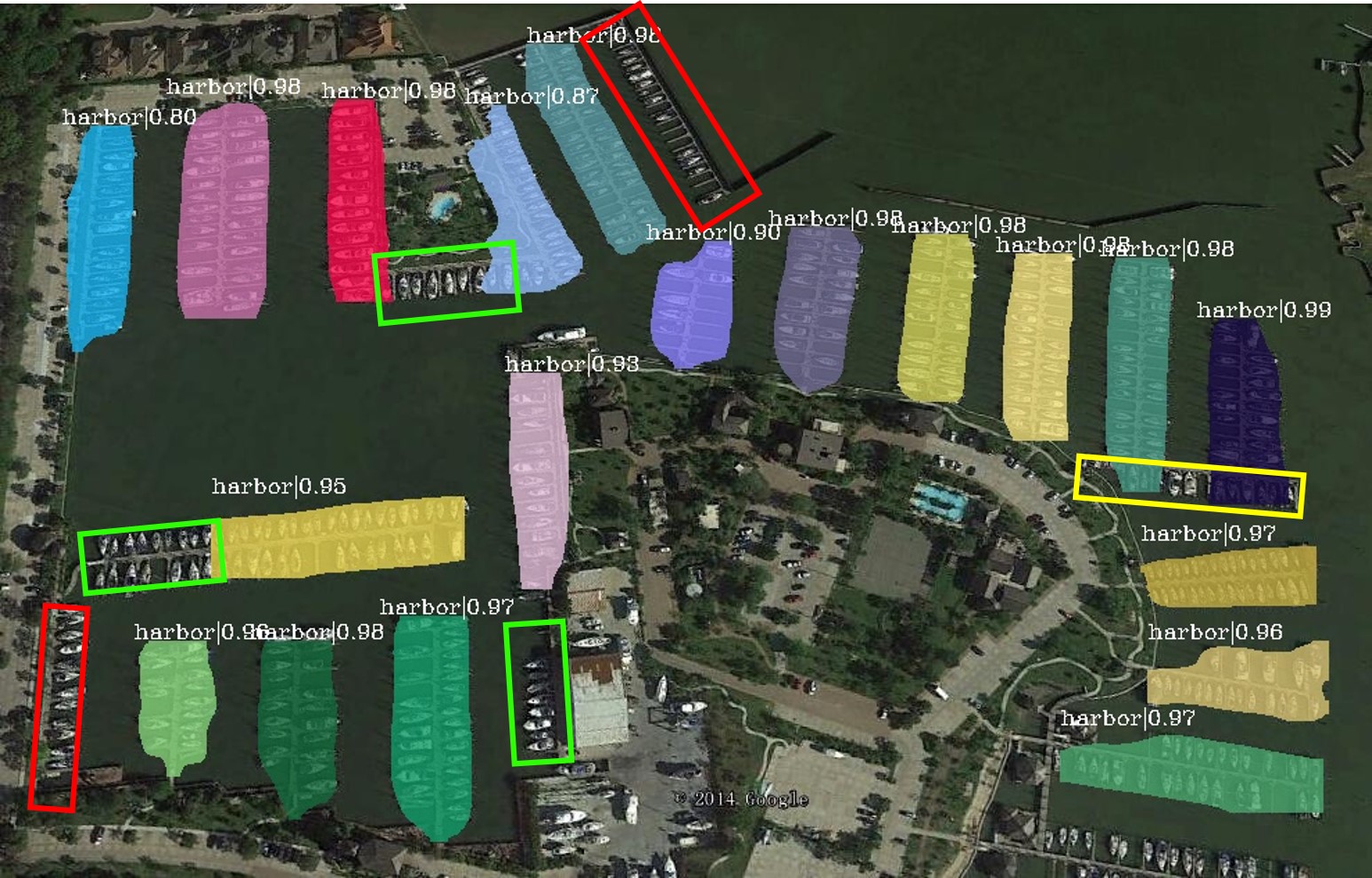}\\
		\end{minipage}%
	}%
	\subfloat[Ours]{
		\begin{minipage}[t]{0.16\linewidth}
			\centering
			\includegraphics[width=1.1in, height=1.1in]{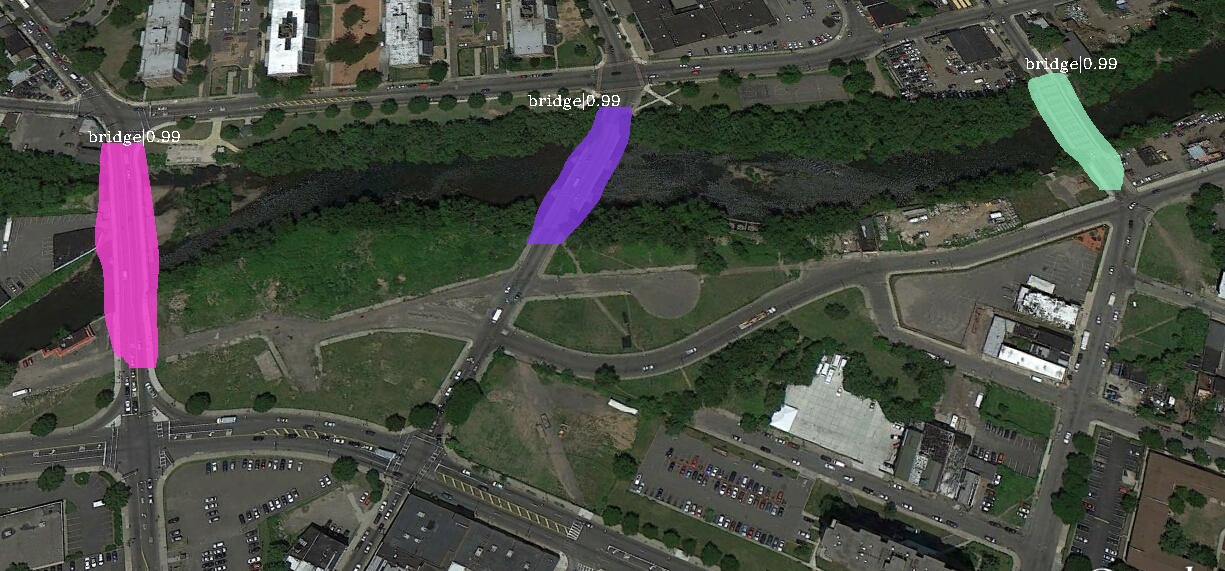}\\
			\vspace{0.2cm}
			\includegraphics[width=1.1in, height=1.1in]{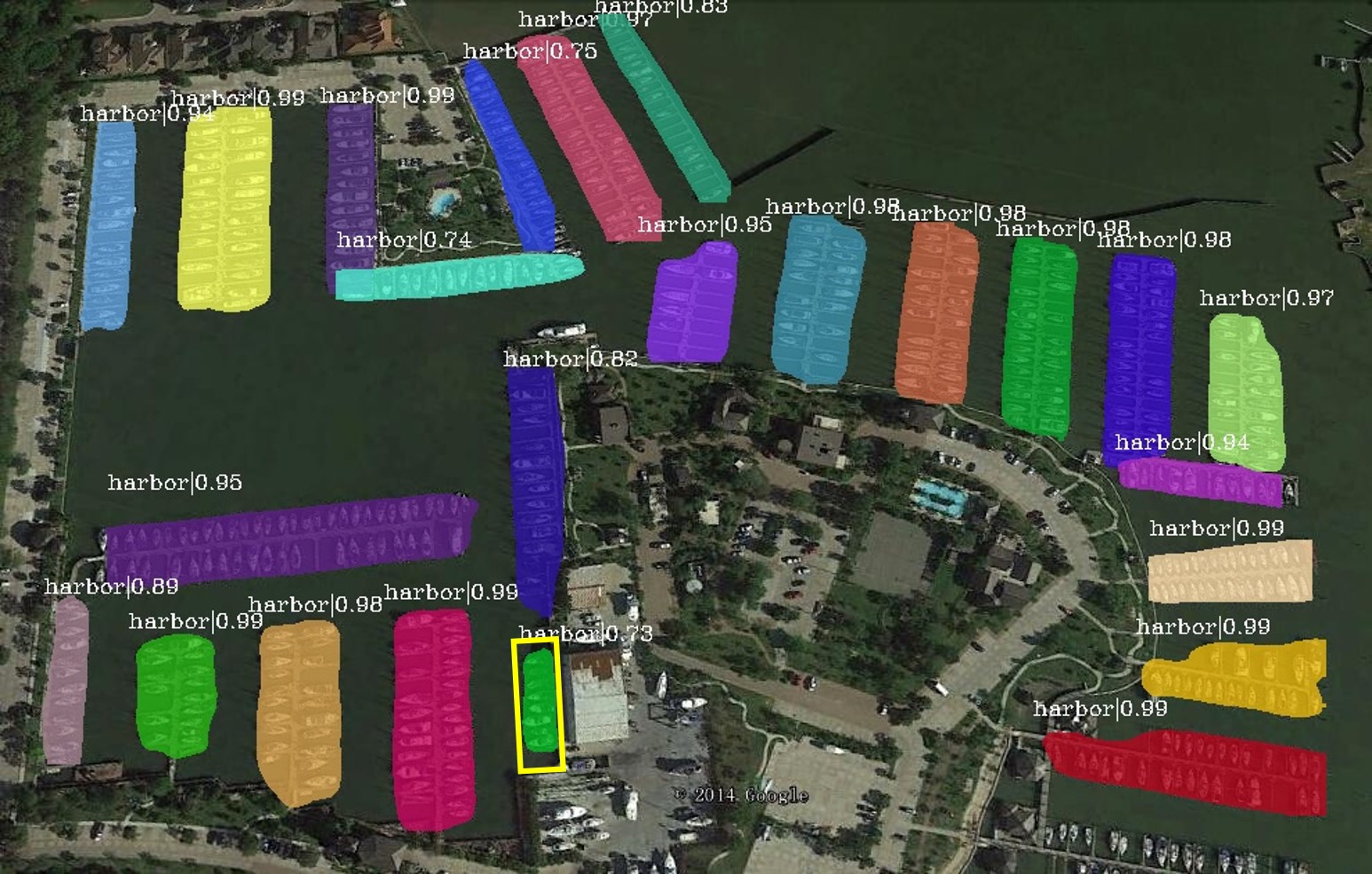}\\
		\end{minipage}%
	}%
	\caption{Comparison results on NWPU VHR-10 Instance Segmentation dataset. We can see that our SS-PANet can better avoid the false detection (in the first row) and reliefs under-segmentation results (in the second row). The false prediction results and the miss prediction results are indicated by yellow and red rectangles, respectively. The under-segmentation results are surrounded by green rectangles. The bounding boxes are removed for simplicity.}
	\label{fig. different NWPU}
\end{figure*}
\begin{figure*}[!t]
	\centering
	\subfloat[Airplane]{
		\begin{minipage}[t]{0.24\linewidth}
			\centering
			\includegraphics[width=1.7in, height=1.7in]{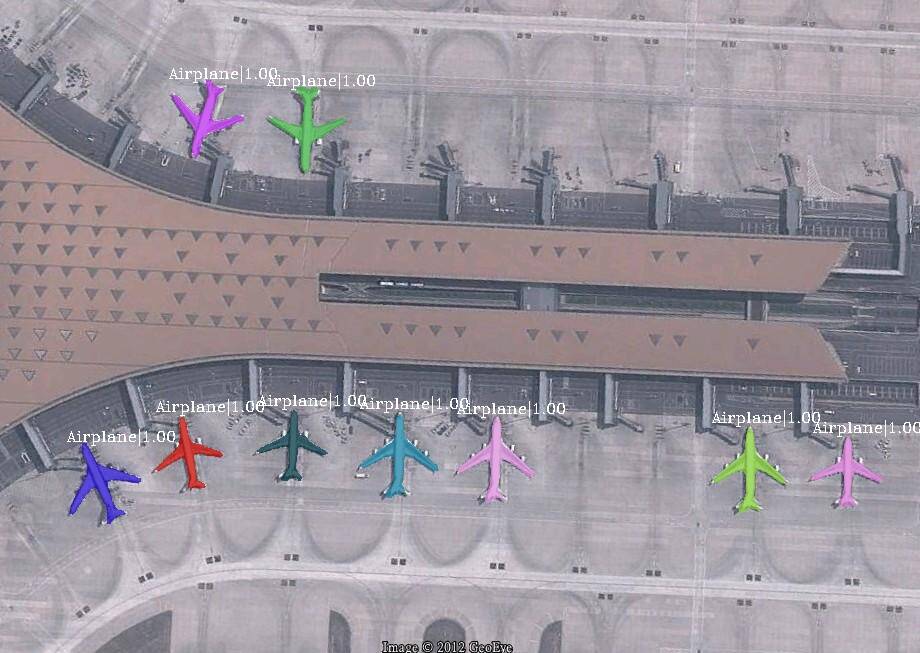}
		\end{minipage}%
	}%
	\subfloat[Ship]{
		\begin{minipage}[t]{0.24\linewidth}
			\centering
			\includegraphics[width=1.7in, height=1.7in]{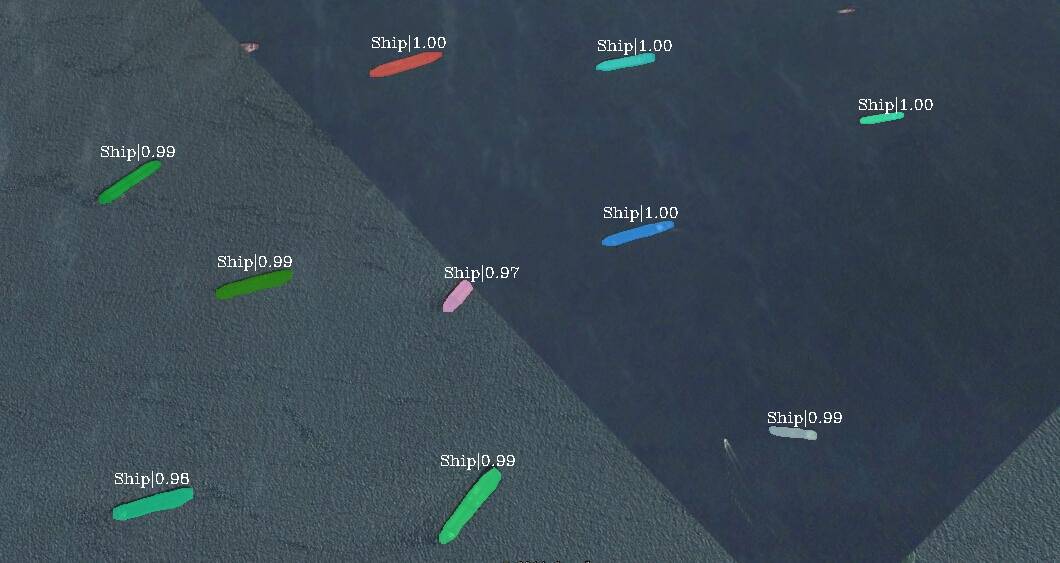}
		\end{minipage}%
	}%
	\subfloat[Baseball diamond, Tennis court and Basketball court]{
		\begin{minipage}[t]{0.24\linewidth}
			\centering
			\includegraphics[width=1.7in, height=1.7in]{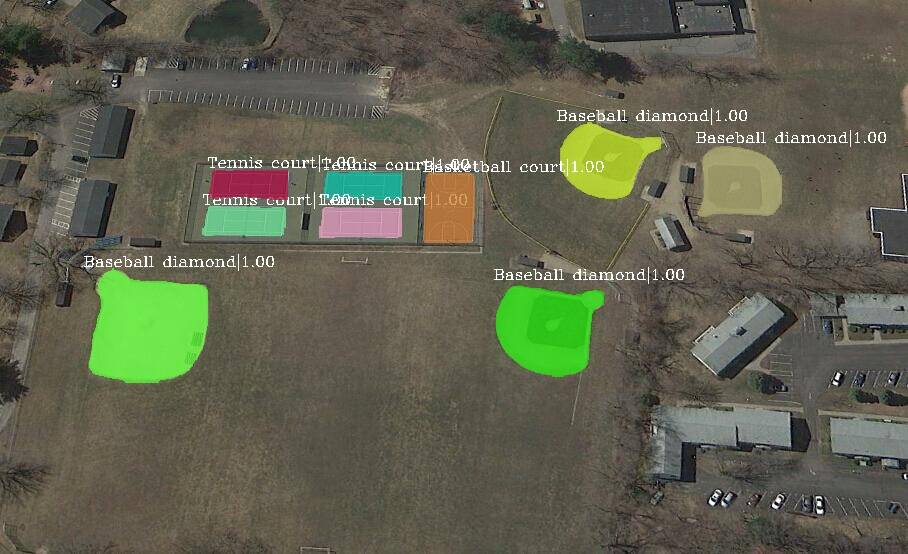}
		\end{minipage}%
	}%
	\subfloat[Bridge]{
		\begin{minipage}[t]{0.24\linewidth}
			\centering
			\includegraphics[width=1.7in, height=1.7in]{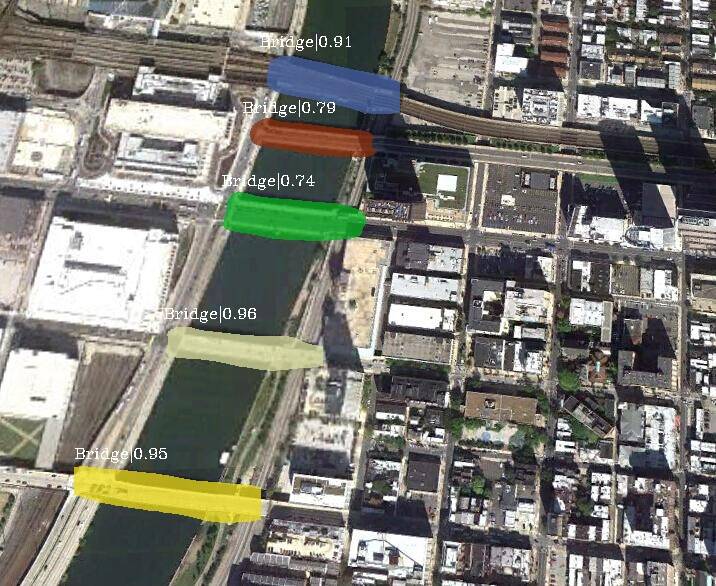}
		\end{minipage}%
	}%
	\hfill
	\centering
	\subfloat[Harbor]{
		\begin{minipage}[t]{0.24\linewidth}
			\centering
			\includegraphics[width=1.7in, height=1.7in]{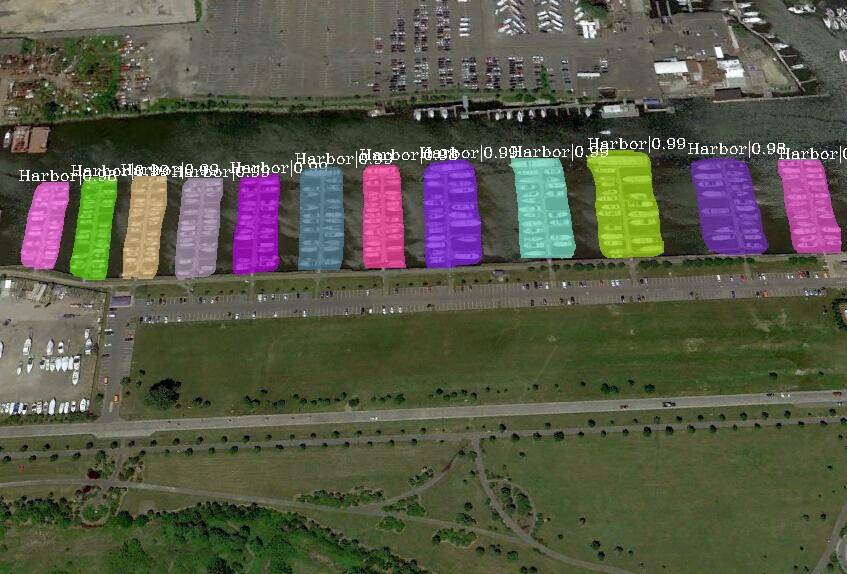}
		\end{minipage}%
	}%
	\subfloat[Storage tank]{
		\begin{minipage}[t]{0.24\linewidth}
			\centering
			\includegraphics[width=1.7in, height=1.7in]{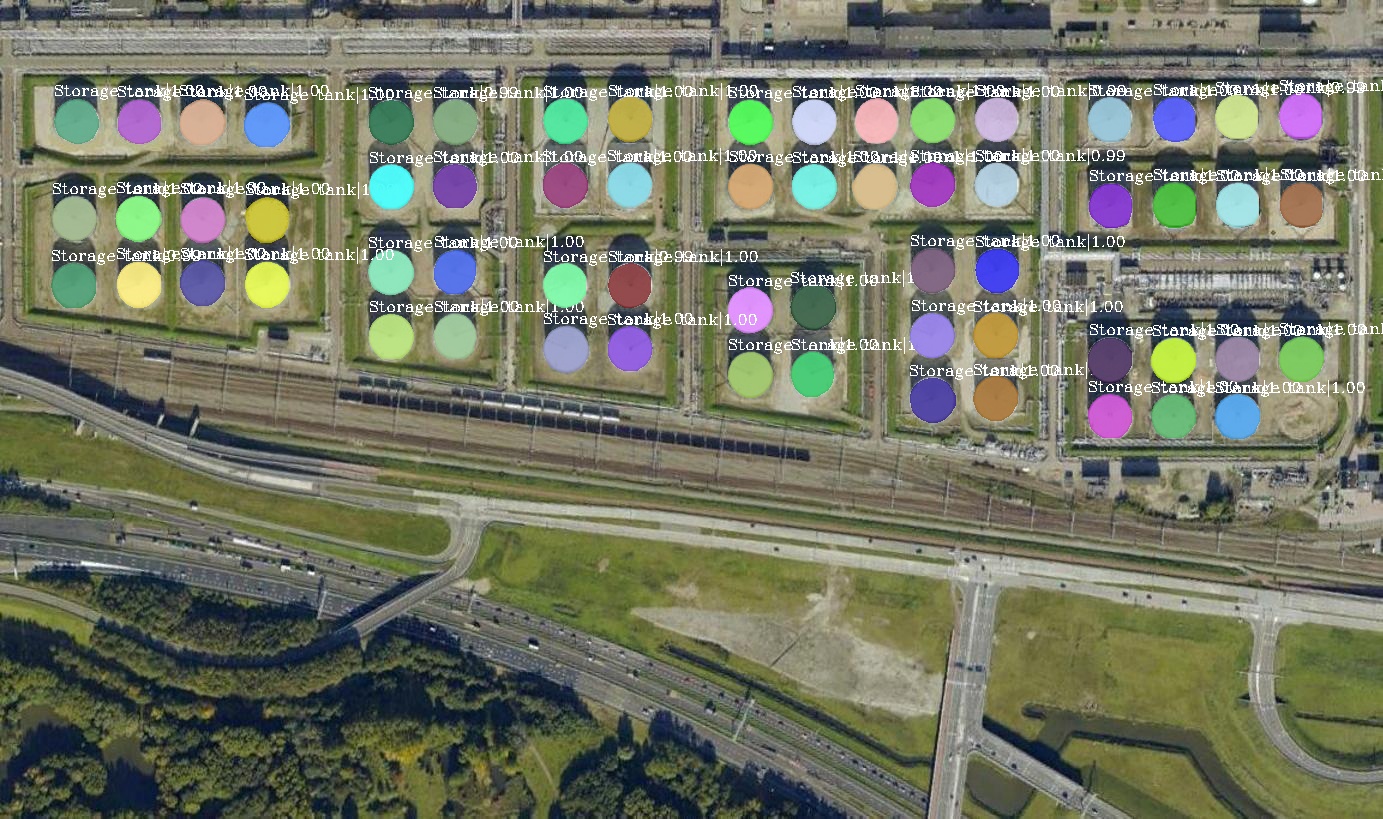}
		\end{minipage}%
	}%
	\subfloat[Vehicle]{
		\begin{minipage}[t]{0.24\linewidth}
			\centering
			\includegraphics[width=1.7in, height=1.7in]{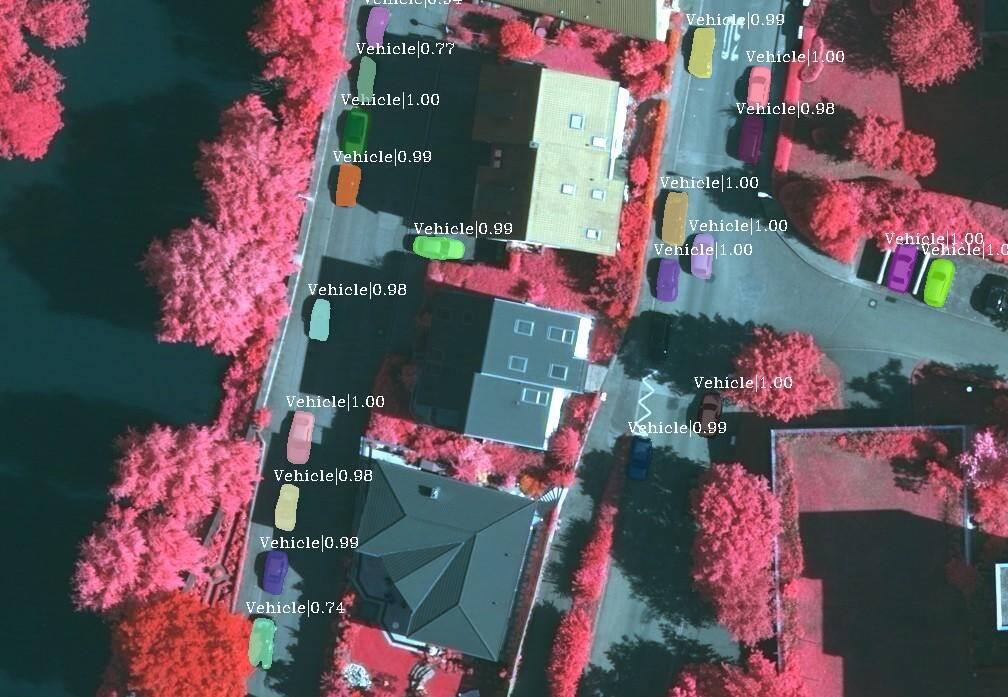}
		\end{minipage}%
	}%
	\subfloat[Ground track field and Tennis court]{
		\begin{minipage}[t]{0.24\linewidth}
			\centering
			\includegraphics[width=1.7in, height=1.7in]{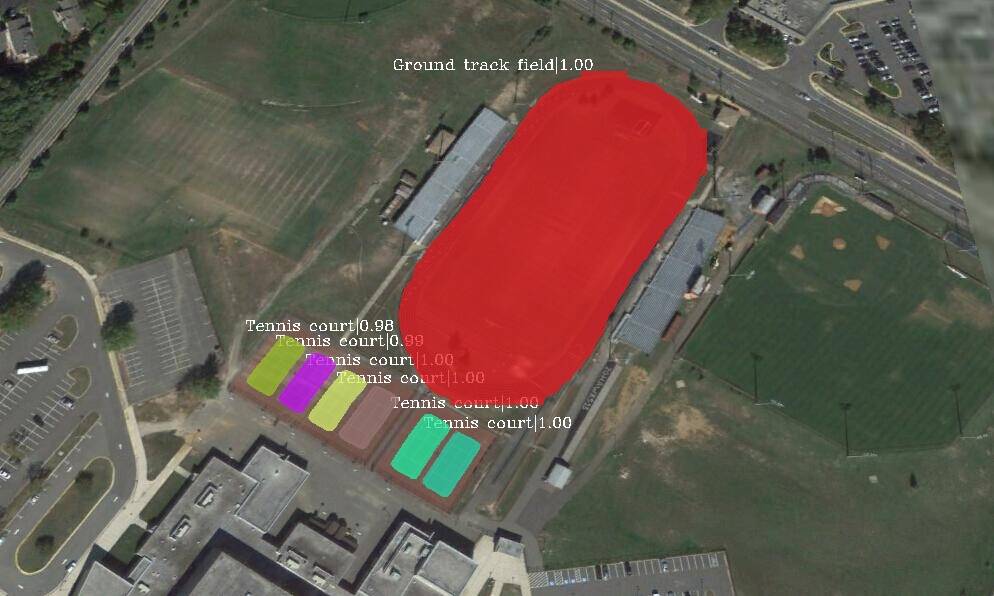}
		\end{minipage}%
	}%
	
	\caption{Performance of our proposed method on NWPU VHR10 Instance Segmentation dataset. Each sub-pictures demonstrates the class-wise results. The bounding boxes are removed for simplicity.}
	\label{fig. result NWPU}
\end{figure*}
	
	To further study the results of different categories, we also reveal the class-wise $AP^{m}$ and $AP^{b}$ in Tables \ref{table-V} and \ref{table-VI}. In Table \ref{table-V}, the SS-PANet+ achieves the best $AP^{m}$ in 14 categories except helicopter and increases more than 4\% for plane and baseball diamond compared with PANet. Similar patterns can be found in Table \ref{table-VI}. Despite achieving impressive results, the proposed method obtains a low $AP$ for the bridge in both detection and segmentation. This is mainly because the aspect ratio of the bridge is large, and hence, the anchors with default aspect ratios of 2:1, 1:1 and 1:2 can not better fit them, which in turn affects the segmentation results. In addition, due to the small size of the small vehicle and the limited number of helicopter samples, their performance is also poor. Besides, it is worth noting that there is still a large margin between the detection results and the segmentation results. Specifically, the segmentation results of some categories decrease more than 15\% compared to the detection results. For the ground track field, they often contain the soccer-ball field leading to misclassification for pixels locating in overlapping regions. As for harbor, plane, and helicopter, the drop is mainly due to the complex contours. Visualization results for all categories are shown in Fig. \ref{fig. result iSAID}.
		
	Table VII shows the performance of our approach on the iSAID test set, where the compared methods are based on the official evaluation in \cite{iSAID}. We use the symbol ‘+’ to denote the models using multi-scale training strategies. From Table VII, we can see that the proposed SS-PANet with the single-scale training strategy achieves comparison results with PANet+\cite{iSAID}. Besides, our proposed SS-PANet is 1.8\% higher than D2Det \cite{D2Det} in the term of $AP^{m}$. Compared to the multi-stage approaches \cite{HTC, CascadeR-CNN}, SS-PANet gets comparable performance in $AP^{m}$, but there is still a large gap in $AP^{b}$. This is mainly because the multi-stage approaches can get better detection performance through multiple regressions and classifications. When applying the multi-scale training strategy in SS-PANet, we obtain the best performance as 40.6\%/46.6\%. In addition, we note that the improvement of our approach in segmentation is significantly higher than the detection results (1.1 vs 0.3). We believe that this difference mainly comes from the difference in multi-scale training strategy. In official implementation \cite{iSAID}, the scale augmentations of the shorter side are at five scales (1200,1000,800,600,400), while we only choose scale augmentations at (1000,800,600,400), which limits the detection results. Despite the detection performance is limited, we still achieve great segmentation performance, which reflects the effectiveness of the proposed method in segmentation.
	
	\subsection{Results on NWPU VHR-10 Instance Segmentation}
	In Table \ref{table-VII}, we report the overall performance of our proposed method.  With the SEA module and SCMB, the $AP^{m}$ and $AP^{b}$ improve by 1.4\%/1.3\% and 3.0\%/2.8\% compared with the Mask-RCNN/PANet. Since we use the modified COCO evaluation metric \cite{iSAID} and the area of all instances in the NWPU VHR-10 Instance Segmentation dataset is less than $512 \times 512$, the $AP^{m}_{l}$ and $AP^{b}_{l}$ in Table \ref{table-VII} are empty. In addition, we find that the $AP_{m}^{m}$ is slightly better than the $AP_{m}^{b}$ of SS-PANet+. This is mainly due to the characteristics of the NWPU VHR-10 Instance Segmentation dataset. The instances of most categories in this dataset (such as ground track field, baseball diamond, basketball court, etc.) are in medium-scale and have regular contours, which is simple for segmentation. However, since the instances are always arbitrarily oriented in RSIs, the horizontal bounding boxes may cause inaccurate detection results, especially under high IoU thresholds. Fig. \ref{fig. different NWPU} demonstrates the comparison results, where the proposed method effectively handles the impact of complex backgrounds, such as the misclassification between roads and bridges, and the false detection of the parking lot as the harbor. Besides, our SS-PANet can better deal with the under-segmentation of harbor and airplane.
	
	Tables \ref{table-IX} and \ref{table-X} show the segmentation and detection results in all categories and our proposed method demonstrates the superior performance compared with the baseline. Especially, for the bridge, we obtain more than 5\% performance gains in detection results, because the proposed SEA module prevents the road from misclassifying as the bridge (as shown in Fig. \ref{fig. different NWPU}). Besides, the segmentation results of the ground track field, basketball court, and baseball diamond are significantly better than their detection results, which validates our conjecture in the previous paragraph. However, due to the large aspect ratio of the bridge and the complex contours of the airplane, their segmentation results are still poor. Fig. \ref{fig. result NWPU} visualizes the results of each category.

	\section{Conclusion}
	In this paper, we focus on the multi-category instance segmentation in remote sensing images and propose an end-to-end instance segmentation framework. Taking into account the complex background in RSIs, we design the Semantic Attention (SEA) module with extra segmentation supervision to improve the activation of instances under noise interference. Meanwhile, we introduce the Scale Complementary Mask Branch (SCMB) which integrates information from different scales to tackle the under-segmentation results. Experiments demonstrate that our method achieves better performance competed with the state-of-the-art methods.
	
	Although the proposed method achieves satisfactory improvements, there is still a large margin between the segmentation and detection results. This is mainly because the bird-views of RSIs lead to the arbitrary orientations of objects, and the horizontal bounding boxes in the detection results can not closely surround the instances, which may affect the segmentation result in the bounding box. Therefore, in future work, we will consider the rotation information of RSIs to further improve both detection and segmentation results.

	
	
	\ifCLASSOPTIONcaptionsoff
	\fi
	\bibliographystyle{IEEEtran}
	\bibliography{IEEEabrv,mybibfile}
	%
	

\end{document}